\documentclass{article}

\PassOptionsToPackage{numbers, compress}{natbib}

\usepackage[final]{neurips_2024}

\usepackage[utf8]{inputenc} % allow utf-8 input
\usepackage[T1]{fontenc}    % use 8-bit T1 fonts
\usepackage{hyperref}       % hyperlinks
\usepackage{url}            % simple URL typesetting
\usepackage{booktabs}       % professional-quality tables
\usepackage{amsfonts}       % blackboard math symbols
\usepackage{nicefrac}       % compact symbols for 1/2, etc.
\usepackage{microtype}      % microtypography
\usepackage{xcolor}         % colors
\usepackage{comment}

\usepackage{orcidlink}
\usepackage{comment}
\usepackage{wasysym}
\usepackage{listings}
\usepackage{xcolor}
\usepackage{float}
\usepackage{graphicx}
\usepackage{epstopdf}
\usepackage{booktabs}
\usepackage{wrapfig}

\usepackage{amsmath} 
\usepackage{wasysym}
\usepackage{algorithm}
\usepackage{algpseudocode}
\definecolor{codegreen}{rgb}{0,0.6,0}
\definecolor{codegray}{rgb}{0.5,0.5,0.5}
\definecolor{codepurple}{rgb}{0.58,0,0.82}
\definecolor{backcolour}{rgb}{0.95,0.95,0.92}
\lstdefinestyle{mystyle}{
    backgroundcolor=\color{backcolour},   
    commentstyle=\color{codegreen},
    keywordstyle=\color{magenta},
    numberstyle=\tiny\color{codegray},
    stringstyle=\color{codepurple},
    basicstyle=\ttfamily\scriptsize,
    breakatwhitespace=false,         
    breaklines=true,                 
    captionpos=b,                    
    keepspaces=true,                 
    showspaces=false,                
    showstringspaces=false,
    showtabs=false,                  
    tabsize=2
}
\let\cite\citep

\newcommand{\model}{{{ICAL}}}

\newcommand{\vlm}{VLM}

\newcommand{\exampleset}{M}

\definecolor{mediumgray}{gray}{0.6}

\title{ICAL: Continual Learning of Multimodal Agents by Transforming Trajectories into Actionable Insights}
\title{VLM Agents Generate Their Own Memories: Distilling Experience into Embodied Programs}
\title{\Large VLM Agents Generate Their Own Memories: \\ \Large Distilling Experience into Embodied Programs of Thought}

\hypersetup{
    colorlinks=true,
    linkcolor=black,     % Color for internal links
    citecolor=black,     % Color for citation links
    filecolor=blue,     % Color for file links
    urlcolor=blue       % Color for external links
}

\author{
  Gabriel Sarch$^1$ \quad Lawrence Jang$^1$ \quad Michael J. Tarr$^1$ \AND William W. Cohen$^{1,2}$ \quad Kenneth Marino$^2$ \quad Katerina Fragkiadaki$^1$ \\
  \vspace{2mm} \\ % Adjust this value to change the spacing
  $^1$Carnegie Mellon University \quad $^2$Google DeepMind \\
  \vspace{2mm}\\ % Adjust this value if needed
  \href{https://ical-learning.github.io/}{https://ical-learning.github.io}
}

\begin{document}

\maketitle

\begin{abstract}
Large-scale generative language and vision-language models (LLMs and VLMs) excel in few-shot learning but require high-quality demonstrations. We propose In-Context Abstraction Learning (ICAL), enabling VLM agents to transform suboptimal trajectories into high-quality training data through self-reflection and human feedback. Given imperfect task demonstrations, a VLM abstracts trajectories into generalized strategies and action annotations by correcting inefficiencies and annotating cognitive abstractions: causal relationships, object state changes, temporal subgoals, and task-relevant visual elements. These annotations are iteratively refined through human feedback during execution in similar environments. The resulting examples significantly improve decision-making when used for retrieval-augmented generation or fine-tuning. As the agent's example library grows, it becomes more efficient at abstracting new examples, requiring less human feedback and fewer environment interactions. ICAL achieves state-of-the-art results across multiple benchmarks. In TEACh dialogue-based instruction following, combining fine-tuning and retrieval on ICAL examples outperforms raw human demonstrations and expert examples by 17.5\% in goal-condition success. In VisualWebArena, retrieval-augmented GPT-4V with ICAL improves task success 1.6x, while fine-tuned Qwen2-VL achieves 2.8x improvement over the base model. In Ego4D action forecasting, we surpass few-shot GPT-4V and remain competitive with supervised models. Our approach scales 2x better than raw demonstrations and significantly reduces manual prompt engineering requirements.
\end{abstract}

% \begin{figure}[t]
%     \centering
%     \includegraphics[width=\linewidth]{Figures/Figure1_vertical_cropped.jpg}
%     \caption{ICAL (In-Context Abstraction Learning) is a method for efficient agent learning
%     from both noisy visual demonstrations and human feedback using large language / vision models.
%     \textit{Left:} The agent can take in a video demonstration, and generate a refined example with language annotations to be used later by the \vlm{} via in-context learning. \textit{Right:}  Humans  provide feedback, correct errors and supply additional knowledge. 
%   }
%     \label{fig:teaser}
% \end{figure}

\section{Introduction}

Humans exhibit remarkable few-shot learning capabilities, rapidly generalizing from a single task demonstration to related conditions by integrating the observed behavior with their internal world model. They discern what is relevant and irrelevant for success and anticipate potential failures. Through repeated practice and feedback, they quickly find the right abstraction that helps to imitate and adapt the task to various situations. This process facilitates continuous refinement and transfer of knowledge across a diverse range of tasks and contexts.

Recent research has explored the use of large language models (LLMs) and visual-language models (VLMs)
\footnote{Throughout the remainder of the paper, we refer to multimodal large language models capable of processing both text and images (e.g., GPT-4V) as `\vlm{}s'.} 
to extract high-level insights from trajectories and experiences. These insights are generated through the model's introspection and are used to enhance performance by appending them to prompts, leveraging their strong in-context learning abilities~\citep{kojima2022large, shinn2023reflexion, majumder2023clin, mu2202improving}. Existing methods often linguistically focus on task reward signals~\citep{shinn2023reflexion, majumder2023clin, wang2023voyager, wang2023jarvis1}, store human corrections following failures~\citep{zha2023distilling, dai2023think, sarch2023helper},
use domain experts to hand-write or hand-pick examples without introspection~\citep{sarch2023helper,song2023llmplanner}, 
or utilize language to shape policies~\citep{harrison2017guiding, tasrin2021influencing} and rewards~\citep{mu2022improving, bahdanau2018learning, goyal2021pixl2r, fu2019language, goyal2019using, mirchandani2021ella, tasrin2021influencing, katara2023gen2sim, ma2023eureka}. Critically, these methods typically are text-based and do not incorporate any visuals cues or demonstrations, or use introspection only in case of failures, which is only one of several ways that humans and machines can consolidate experiences and extract insights.

\textbf{In this work, we teach \vlm{}s novel tasks by learning in-context programs of thought given sub-optimal demonstrations and human natural language feedback}. 
%\textbf{In this work, we guide \vlm{}s to map general, potentially sub-optimal, demonstrations to experience abstractions to be used as prompt examples for in-context task planning}.
We present In-Context Abstraction Learning (\model{}), a method that prompts \vlm{}s to create multimodal programs of thought for unfamiliar domains. Unlike previous works that only store and retrieve successful action plans or trajectories~\citep{sarch2023helper, wang2023voyager, liang2022code}, our approach emphasizes learning thoughts that encapsulate the dynamics and critical knowledge of tasks. Specifically, \model{} tackles four types of cognitive abstractions: \textbf{task and causal relationships}, which identify the fundamental principles or actions needed to achieve a goal and how elements are interconnected through cause and effect~\citep{tenenbaum2011grow}; \textbf{changes in object states}, which describe the various forms or conditions an object will take~\citep{barrett2009see}; \textbf{temporal abstractions}, which break down tasks into subgoals~\citep{botvinick2009hierarchically}; and \textbf{task construals}, which highlight critical visual details within a task~\citep{ho2022people}.
When provided with suboptimal trajectories, \model{} prompts a \vlm{} to transform these trajectories into optimized examples while also creating pertinent language and visual abstractions. These abstractions are then refined through executing the trajectory in the environment, guided by natural language feedback from humans. Each step of thought generation leverages previously learned examples from memory, enabling the model to improve not only its execution but its abstraction capabilities as well. Collectively, the learned abstractions summarize crucial information about action sequences, state transitions, rules, and focus areas, articulated through free-form natural language and visual representations.

We present a comprehensive evaluation of our agent, equipped with the learned examples, across three benchmarks: TEACh~\citep{TEACH} for dialogue-based instruction in household settings, VisualWebArena~\citep{koh2024visualwebarena} for multimodal autonomous web tasks, and Ego4D for video action anticipation~\citep{grauman2022ego4d}. In TEACh, our agent sets a new state-of-the-art, outperforming \vlm{} agents reliant on raw demonstrations or extensive domain-expert hand-written examples, demonstrating the effectiveness of \model{} learned abstractions for in-context learning. Specifically, our approach achieves a 12.6\% improvement in goal condition success compared to the previous SOTA, HELPER~\citep{sarch2023helper}, using retrieval-augmented generation. We show that this approach scales 2x faster than raw human demonstrations when evaluated on unseen tasks, and achieves a 14.7\% performance increase over zero-shot baselines after only ten examples. Moreover, our agent becomes increasingly efficient over time by leveraging stored examples during the learning process, requiring 38.8\% fewer environment steps and 71.6\% less human feedback per example in the latter half of demonstrations processed.
In TEACh, integrating our learned examples with LoRA-based fine-tuning~\citep{hu2021lora} of GPT3.5 further improves goal-condition performance by 4.9\%. 
In the VisualWebArena, our agent surpasses the state-of-the-art, GPT4V + Set of Marks~\citep{koh2024visualwebarena}, improving from 14.3\% to 22.7\% using GPT4V and from 18.9\% to 23.4\% using GPT4o. Finetuning Qwen2-VL on the learned examples improves performance from 2.9\% to 8.2\%.
% In the VisualWebArena, \model{} improves over the the state-of-the-art an absolute 8.4\% (relative 58.7\%) improvement in average success rate over GPT4V and shows a 23.8\% relative improvement in average success rate over GPT4o. 
In the Ego4D setting, ICAL outperforms few-shot GPT4V using chain of thought, reducing the noun and action edit distance by 6.4 and 1.7, respectively, and competes closely with fully supervised methods, despite using 639x less in-domain training data. Our approach significantly reduces reliance on human-expert-crafted examples and consistently outperforms in-context learning from action plans or trajectories that lack such optimized programs of thought~\citep{sarch2023helper, wang2023voyager, liang2022code}.

\section{Related Work}

\paragraph{\vlm{} Agents}
LLMs and \vlm{}s trained from large scale vision-language data have been adapted for task planning and decision making tasks through in-context prompt optimization or finetuning. 
\vlm{}s have been used to plan over high-level actions or code~\citep{wang2023describe,wang2023voyager, sarch2023helper, liang2022code, singh2022progprompt}, incorporate error feedback~\citep{liu2023reflect, liang2024learning, zha2023distilling}, and understanding game instruction manuals~\citep{wu2023read}. 
%They have improved text-based agents in simulated worlds~\citep{yao2022react,shinn2023reflexion,wu2023plan,richards2023auto}. 
Some studies use \vlm{}s for learning from human feedback through retrievable knowledge~\citep{zha2023distilling}, question asking~\citep{ren2023robots,dai2023think}, or converting language to actions or rewards~\citep{liu2023interactive,liu2023model,klissarov2023motif,cheng2023llf,cui2023no}. Other work combines tree search with VLMs to explore multiple trajectory paths~\citep{koh2024tree,putta2024agent}.
Our work utilizes noisy visual trajectories, and uses VLM self-refinement with human-in-the-loop learning to obtain high-quality trajectory data of actions and reasoning chains.

\paragraph{Instructable Interactive Agents}
Benchmarks for embodied instruction following include question answering~\citep{gordon2018iqa,das2018embodied,zhu2023excalibur,datta2022episodic,das2020probing,gao2022dialfred}, navigation~\citep{ku2020room,krantz2023navigating,chen2020soundspaces}, interactive dialogue, and instruction following~\citep{yenamandra2023homerobot,shridhar2020alfred,TEACH,gao2023alexa}. Virtual agent benchmarks focus on web tasks where agents navigate static~\citep{lù2024weblinx,deng2023mind2web} and dynamic web environments~\citep{zhou2023webarena,koh2024visualwebarena,yao2023webshop, koh2024tree}, covering personal shopping, travel assistance, software engineering, and operating system tasks~\citep{liu2023agentbench,jimenez2023swebench,shi2017world,liu2018reinforcement}. 
This includes visual grounding and multi-turn planning, with prior studies using finetuning or few-shot prompts. 
In agent-based domains, retrieval-augmented prompting and prompt optimization have improved task planning in instructional contexts~\citep{song2023llmplanner} and open-world gaming~\citep{wang2023jarvis1,wang2023voyager,majumder2023clin,nottingham2024sso}. 
Unlike studies that rely solely on static external memory or text-based prompting, our research demonstrates that multi-modal, generalizable programs of thought learned from a few noisy trajectories and human feedback via in-context learning or finetuning can significantly improve instruction-following performance.

\section{In-Context Abstraction Learning (ICAL)}
In-Context Abstraction Learning (ICAL) aims at automating the acquisition of generalizable examples and knowledge for in-context agents. ICAL operates by receiving a language instruction $I$ with a noisy trajectory of observations and actions, denoted $\xi_{noisy} = \{o_0, a_0, \ldots, o_T, a_T\}$ in a new task domain $D$. A new domain \( D \) represents changes in task variables not captured in VLM pretraining, such as a different environment (e.g., kitchen \#1 vs. kitchen \#2), task (e.g., "add the cheapest red bike to my wish list"), or user preference (e.g., "I prefer the red cup for coffee"). The core aim of \model{} is to abstract each noisy trajectory into a single example $e$, which then forms part of a memory set \exampleset{}. Each example $e \in \exampleset$ represents an optimized trajectory $\xi_{optimized}$ with generalizable language programs of thought $L$. The objective is to ensure that \exampleset{} collectively encapsulates examples that, when used in a VLMs context window, %an in-context \vlm{},
increase the likelihood of successful task execution in the new domain, while also containing knowledge that is transferable across similar tasks and contexts. This can be encapsulated as:
% \begin{equation} 
%     \max_{\exampleset} \mathbb{E}[R | \exampleset, I, o_t, D],
% \end{equation}
% \begin{equation} 
% \max_{\exampleset} \mathbb{E}\left[\sum_{t=0}^T r_t(s_t, a_t) \Big| \exampleset, I, o_t, D \right], 
% \end{equation}
% \begin{equation} 
% \max_{\exampleset} \mathbb{E}\left[\sum_{t=0}^T r_t(o_t, a_t) \Big| \exampleset, I, D \right], 
% \end{equation}
% \begin{equation} 
% \max_{\exampleset} \mathbb{E}\left[\sum_{t=0}^T r_t(o_t, a_t) \Big| a_t \sim \pi(a | o_t), \exampleset, I, D \right], 
% \end{equation}
% \begin{equation} 
% \max_{\exampleset} \mathbb{E}\left[\sum_{t=0}^T r_t(o_t, a_t) \Big| a_t \sim \pi(a | o_t, M, I), D \right], 
% \end{equation}

% \begin{equation}
% \max_{\mathcal{E}} \mathbb{E}_{\tau}\left[ R(\tau) \,\Big|\, \tau \sim P(\tau | \pi, \mathcal{E}, I, D) \right],
% \end{equation}

% Where:
% \begin{itemize}
%     \item $R(\tau) = \sum_{t=0}^T r_t(o_t, a_t)$ is the cumulative reward over the trajectory $\tau$,
%     \item $\pi(a_t | h_t, \mathcal{E}, I)$ is the policy conditioned on the history $h_t = \{o_0, a_0, \ldots, o_{t-1}, a_{t-1}, o_t\}$, the example set $\mathcal{E}$, and the instruction $I$,
%     \item $P(\tau | \pi, \mathcal{E}, I, D)$ is the probability distribution over trajectories induced by the policy $\pi$ under the given conditions.
% \end{itemize}

\begin{equation}
\max_{M} \mathbb{E}\left[ \sum_{t=0}^T r_t(o_t, a_t) \,\Big|\, a_t \sim \pi(a \mid h_t, M, I), D \right]
\end{equation}

Where: 
\begin{itemize} 
\item $M$ is the memory (e.g., example set) maintained by the in-context agent, 
\item $r_t(o_t,a_t)$ is the reward at time $t$ given observation $o_t$ and action $a_t$, 
\item $\pi(a \mid h_t, M, I)$ is the policy from which actions $a_t$ are sampled, 
\item $h_t = {(o_0,a_0), (o_1,a_1), \ldots, (o_{t-1},a_{t-1}), o_t}$ is the history of observations and actions up to time $t$, 
\item $I$ is the natural language instruction, 
\item $D$ is the task domain (e.g., environment, user preference), which conditions the overall distribution and reward structure. 
\end{itemize}

% \textbf{Objective:}
% \begin{equation}
% \max_{\mathcal{E}} \mathbb{E}_{\tau}\left[ R(\tau) \,\Big|\, \tau \sim P(\tau | \pi, \mathcal{E}, I, D) \right],
% \end{equation}

% \textbf{Where:}
% \begin{itemize}
%     \item $R(\tau) = \sum_{t=0}^T r_t(o_t, a_t)$ is the cumulative reward,
%     \item $\pi(a_t | h_t, \mathcal{E}, I)$ is the policy conditioned on history $h_t$, examples $\mathcal{E}$, and instruction $I$,
%     \item $P(\tau | \pi, \mathcal{E}, I, D)$ is the trajectory distribution induced by $\pi$.
% \end{itemize}

% where $R$ is the return or the cumulative reward acquired by performing actions based on the instruction $I$, observation $o_t$, and in-context example memory set \exampleset{}. 
Rather than optimizing prompt examples through trial and error—which would lead to a challenging search problem that myopically focuses on improving rewards for the current scene—we leverage VLMs' knowledge for thought generation, which we elicit through prompting.

\begin{figure*}[t!]
    \centering
    \includegraphics[width=1.0\textwidth]{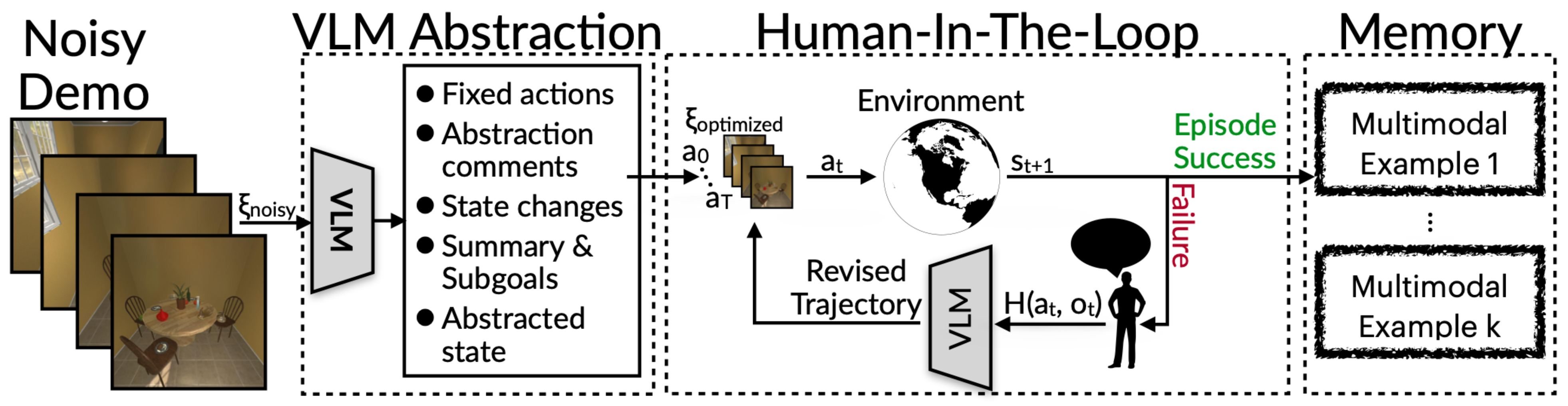}\\[2ex]% Adds more space between the figures
    \includegraphics[width=1.0\textwidth]{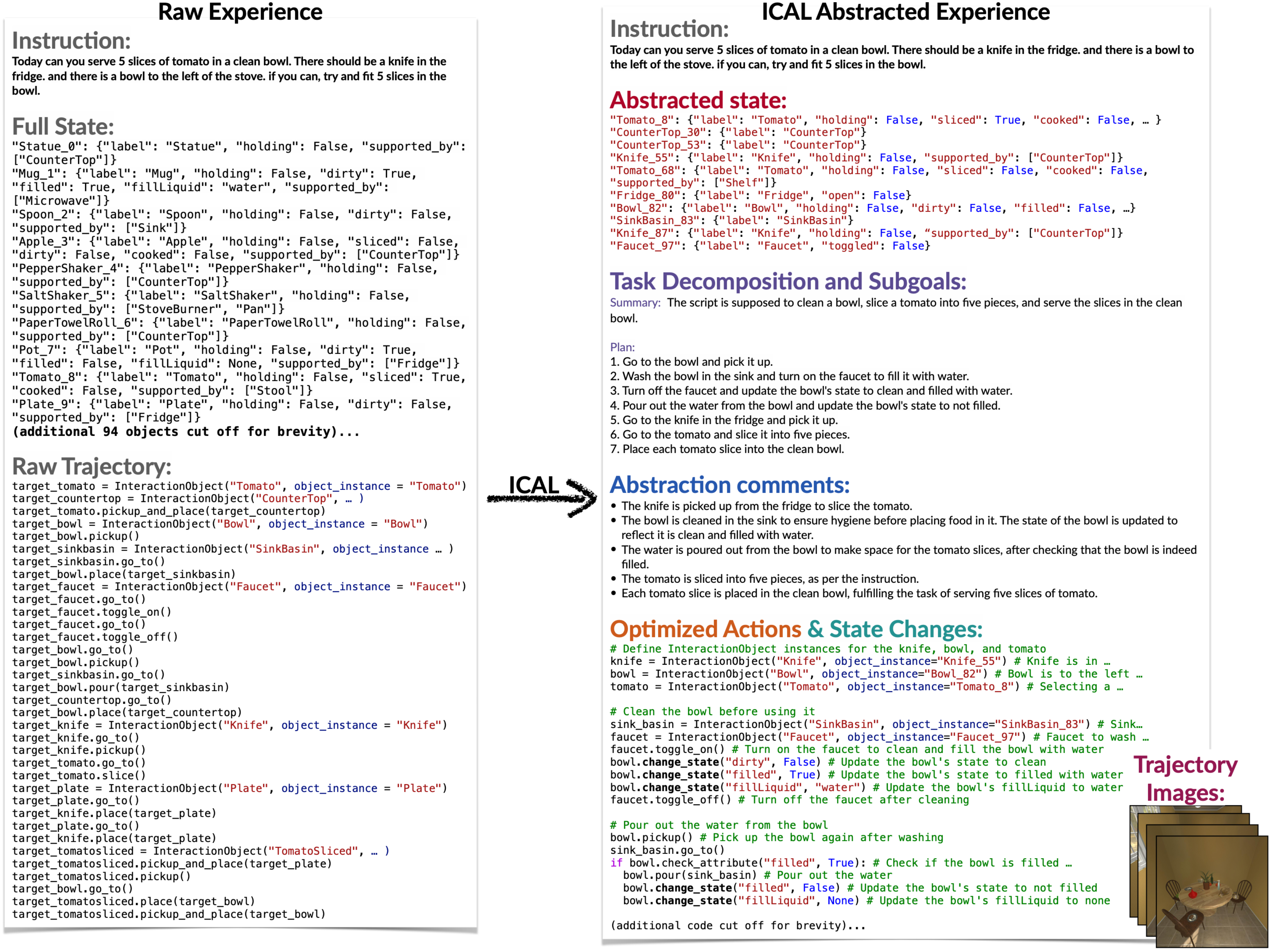}
    \caption{
    \model{} transforms raw experience into useful programs of thought for in-context learning. \textit{Top:} Given a noisy trajectory, 
    It prompts a \vlm{} to optimize actions and add language annotations. The optimized trajectory is executed, incorporating human feedback on failures. Successful examples are stored for future \vlm{} in-context action generation. \textit{Bottom:} An example of the raw, noisy trajectory (left), and the final abstracted example after \model{} (right). 
    }
    \label{fig:methods_overview}
\end{figure*}

\subsection{Overview}
Figure~\ref{fig:methods_overview} shows an overview of ICAL. Each iteration starts with a noisy trajectory, which may come from a human or suboptimal policy. ICAL transform it into a high-quality example in two phases: (1) abstraction phase ($F_{abstract}$), where the VLM identifies and corrects action errors while generating preliminary program-of-thought language annotations (Section~\ref{sec:introspection}). (2) The human-in-the-loop phase, denoted $F_{hitl}$, during which the sequence is executed within the environment and the actions \textit{and} thoughts are further refined by human feedback conveyed in natural language (Section~\ref{sec:hitl}). Upon the successful execution of the trajectory, it is archived within a continually growing repository of examples. These examples serve as contextual references for the agent both during its learning phase and during inference for unseen instructions and environments.

\subsection{\vlm{}-driven Programs of Thought Generation} \label{sec:introspection}

We address the challenge of learning from a diverse set of noisy trajectories $\xi_{noisy} = \{o_0, a_0, \ldots, o_T, a_T\}$, which may be sub-optimal due to several factors: demonstrations by human non-experts, errors in inferring actions from visual passive demonstrations, and agent generated paths that include exploration or failures. 
Please see Section~\ref{sec:envs} for details on noisy trajectory collection.

Abstracting a noisy trajectory, \(\xi_{noisy}\), into a high-quality training example involves transforming it into a more optimized sequence, \(\xi_{optimized}\), and formulating relevant language thoughts, \(L\), as shown in Figure~\ref{fig:methods_overview}. The abstraction function, \(F_{abstract}\), modifies \(\xi_{noisy}\) by correcting actions and generating language thoughts that encapsulate general knowledge and task-specific insights. It is defined as:
\begin{equation}
F_{abstract}: (\xi_{noisy}, I, \{e^1, \ldots, e^k\}) \rightarrow (\xi_{optimized}, L)
\end{equation} 
where \(\xi_{noisy}\) is the initial noisy trajectory, \(I\) is the task instruction, and \(\{e^1, \ldots, e^k\}\) are the top-k previous successful in-context examples. The output consists of the optimized trajectory \(\xi_{optimized}\) and language thoughts \(L\).

Corrections during programs of thought generation include action adjustments and generating annotations (\(L\)) for subgoals, causal relationships, state changes, and reasoning steps. These annotations are produced by prompting the \vlm{} to output a specified type of thought. We prompt the \vlm{} abstraction function, $F_{abstract}$ (GPT4V in this work), to produce the language thoughts detailed below. For the complete prompts, please refer to the Appendix. 

\textbf{1. Task and Causal Abstractions:}
Task and causal abstractions pinpoint the essential principles or actions required to achieve a goal and explain how elements are interconnected through cause and effect. Task and causal abstractions have been shown to be helpful in improving LLM generalization~\citep{majumder2023clin}, and play a strong role in human communication and learning~\citep{tenenbaum2011grow, goodman2016pragmatic}. We prompt the \vlm{} to add annotations of task and causal abstractions in the form of natural language comments. For example, it might add a note explaining unnecessary actions, 
% based on the current state, 
such as "Since the box is already open, there is no need to close it after placing the watches inside, ensuring the task is completed efficiently."
%and as directed."

\textbf{2. State Changes:} Understanding how one's actions will affect the form and conditions of elements in a scene is crucial for decision-making~\citep{barrett2009see}. The \vlm{} is prompted to identify and predict state changes that occur during the demonstration.  For instance, an annotation might note the bowl becoming clean, clearly indicating an expected state transition.

\textbf{3. Task Decomposition and Subgoals:} 
Breaking down a complex task into intermediate steps and subgoals is crucial for managing extended and variable sequences of lower-level actions.
These temporal abstractions are important for human reasoning~\citep{botvinick2009hierarchically} and have been shown to improve LLM outputs~\citep{wei2022chain}. We prompt the \vlm{} to add 1) a step-by-step plan detailing the demonstration, and 2) a natural language summary of the actions.

\textbf{4. State Abstraction:} 
Useful representations do not simply mirror every aspect of the world; instead, they selectively capture a manageable subset of details relevant to a specific purpose~\citep{ho2022people}. We focus on identifying and including only those state variables that are relevant to the task at hand. This is achieved by (1) selecting parts of the state that were directly interacted with by the agent during the demonstration, and (2) prompting the \vlm{} to suggest additional state variables not explicitly included in the demonstrations but potentially relevant to understanding the task. 

\subsection{Programs of Thought Verification with a Human-in-the-loop} \label{sec:hitl}

In this phase, ICAL verifies the generated thoughts with a human-in-the-loop. This involves executing the optimized trajectory, \(\xi_{optimized}\) within the actual task environment, under the watchful guidance of a live human observer. The procedure is:

\textbf{1. Execution of optimized trajectory:} The agent attempts to perform the task by following the optimized sequence of actions \(\xi_{optimized}\) from the programs of thought generation phase.

\textbf{2. Monitoring and Intervention:} As the agent executes \(\xi_{optimized}\), a human observer monitors the process. If an action \(a_t\) fails, denoted by \(F(a_t) = 1\), the observer intervenes by providing natural language feedback \(H(a_t, o_t)\). This feedback is context-specific, addressing the observed failure directly (e.g., explaining that the Toaster is currently full and can only toast one slice of bread). We provide additional details on the human-in-the-loop in the Appendix Section~\ref{app:hitl}.

\textbf{3. Feedback Integration and Trajectory Revision:} Upon receiving feedback \(H(a_t, o_t)\), the \vlm{} is provided with this input alongside the current state of \(\xi_{optimized}\) and any existing language thoughts \(L\). The \vlm{} is prompted to revise \(\xi_{optimized}\) to address the failure, to update existing thoughts \(L\) based on the feedback, and to add new thoughts that capture insights from the feedback.

This process can be represented by an update function:
\begin{equation}
\Xi_{update}(\xi_{optimized}, H(a_t, o_t), L, I, \{e^1, ..., e^k\}) \rightarrow \xi'_{optimized}, L'
\end{equation}
where \(\Xi_{update}\) denotes the update function that takes the current trajectory \(\xi_{optimized}\), human feedback \(H(a_t, o_t)\), and current thoughts \(L\), and outputs the revised trajectory \(\xi'_{optimized}\) and updated thoughts \(L'\). For the complete prompts, please refer to the Appendix.

\textbf{4. Environment Reset and Retrial:} Following a failure and subsequent feedback, the environment is reset to a suitable state for retrying the task. The agent then attempts the task again, utilizing the newly revised trajectory \(\xi'_{optimized}\).

\textbf{5. Success Criteria and Feedback Limit:} This interactive phase continues until the human observer deems the task execution successful, or until a predefined maximum number of feedback iterations, $N_{feedbacks}$, has been reached.

\textbf{6. Saving example:} If successful, we store the revised trajectory $\xi_{optimized}$ and thoughts $L$ to the memory set \exampleset{}. If unsuccessful after $N_{feedbacks}$ iterations, we do not store the example and move to the next demonstration. We experiment with relabeling partially successful demonstrations in Section~\ref{sec:relabel} of the appendix.

Our human-in-the-loop process has similarities to DAgger~\citep{ross2011reduction}, with important differences. We compare our approach to DAgger in Appendix Section~\ref{sec:dagger_appendix}.

\subsection{Retrieval Augmented Generation and Supervised Fine-tuning for Deployment} \label{sec:retrieve}

\begin{wrapfigure}{r}{0.45\textwidth}
\vspace{-5mm}
    \centering
    \includegraphics[width=0.45\textwidth]{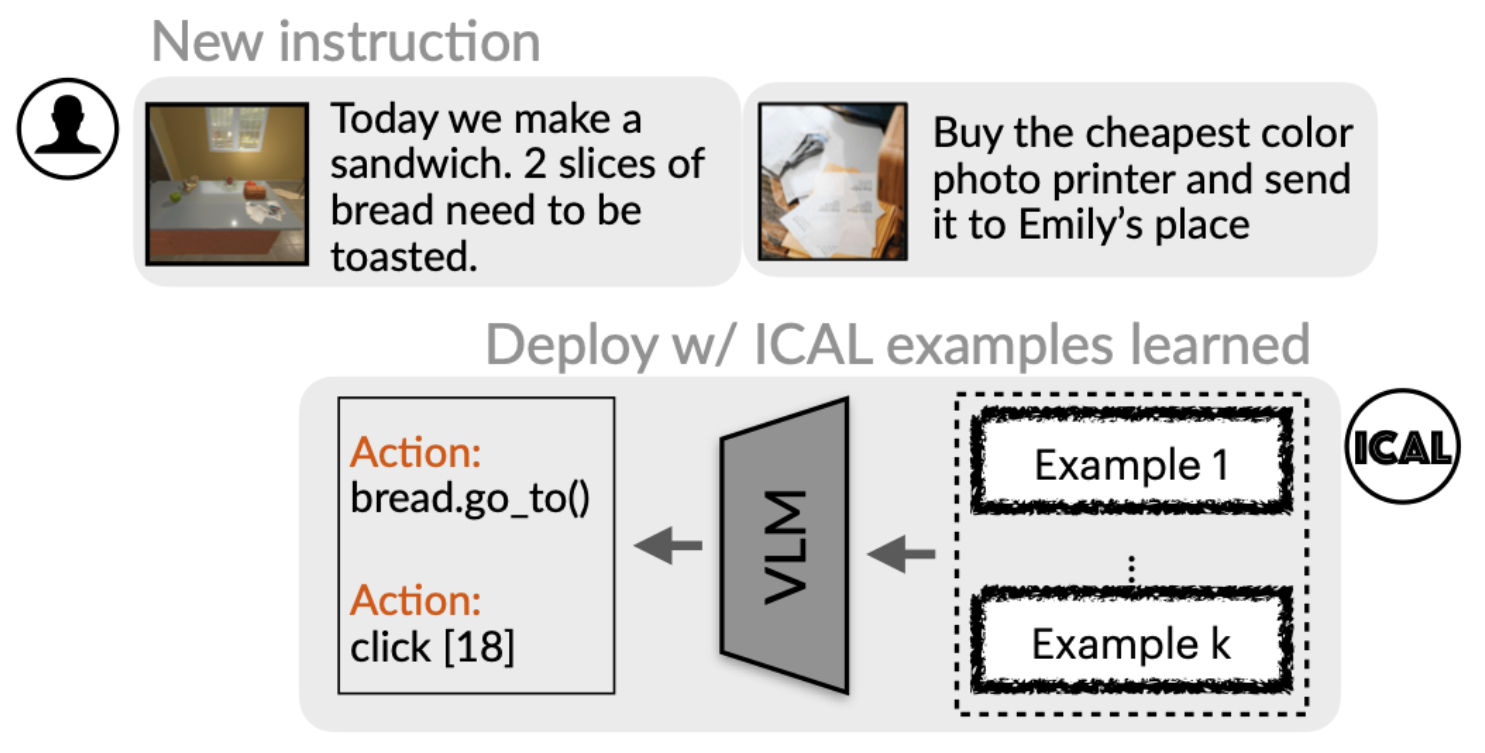}
    \caption{
    After the \model{} examples have been learned, 
    \model{} is deployed for new tasks and environments using retrieval-augmented generation.
    }
    \label{fig:methods_deploy}
    % \vspace{-3mm}
\end{wrapfigure}

\textbf{Retrieval Augmented Generation.}
Given the learned example set \exampleset{} and a new instruction $I$, we prompt the VLM  to carry out the instruction by  producing action sequences \(\{a_0, ..., a_T\} \in A\) from an action API that describes the skills set $A$ (e.g., \texttt{go\_to(X)}, \texttt{pickup(X)}), by retrieving the top $K$ examples from \exampleset{} to include in the prompt based on their textual and visual similarity with the current scene. The aggregated similarity score \(s\) for each example \(e\) reads: %, we use the equation:
\begin{equation}
s = \lambda_{I} \cdot s^I + \lambda_{\text{textual}} \cdot s^{\text{textual}} + \lambda_{\text{visual}} \cdot s^{\text{visual}},
\end{equation}
where \(s^I\), \(s^{\text{textual}}\), and \(s^{\text{visual}}\) are the similarity scores for the input text instruction, textual state, and visual state, respectively, computed via cosine similarity using embeddings from OpenAI's text-embedding-ada-002 model and CLIP ViT-B/32 model. The coefficients \(\lambda_{I}\), \(\lambda_{\text{textual}}\), and \(\lambda_{\text{visual}}\) are weighting hyperparameters chosen in each domain by a held out validation set.

The VLM prompt contains the new instruction $I$, the current webpage image for web agents or 12 video frames for ego4D annotated with set-of-marks~\citep{yang2023setofmark}, a textual state description \(x_t\) describing the objects and their attributes for embodied agents and HTML elements for web agents, the action API $A$, and the retrieved set of in-context examples ${e^1,...,e^k} \in \exampleset$. 
An illustration of this process is shown in Figure~\ref{fig:methods_deploy}. The deployment prompt is provided in the Appendix. 

\textbf{Supervised Fine-tuning.}
Once \exampleset{} is constructed, we also use it to perform supervised fine-tuning (SFT). Each example \( e \in \exampleset \) is formatted as an input-output pair, where the input consists of the instruction \(I\), and relevant environment state and observations, while the output is the programs of thought \(L\) and final optimized action sequence \(\xi_{optimized}\). Through SFT on \exampleset{}, the agent internalizes corrected action patterns and associated reasoning, resulting in improved zero-shot performance when deployed on new tasks or environments.

\textbf{Implementation details} We use \texttt{GPT-4-1106-preview}~\citep{gpt4technical} for text generation, unless otherwise stated, and \texttt{text-embedding-ada-002}~\citep{ada003technical} for text embeddings. We use \texttt{gpt-4-1106-vision-preview}~\citep{gpt4technical} for the text and image generation model. We use $k=5$ for example retrieval. We use a temperature of 0 for TEACh and Ego4D, and 0.2 for VisualWebArena. For fine-tuning Qwen2-VL-7B on the ICAL VisualWebArena dataset, we use a 90/10 train-validation split. The model is fine-tuned for 10 epochs with a learning rate of \(5 \times 10^{-6}\) and a weight decay of 0.001.

\section{Experiments}
We test ICAL for task planning in TEACh~\citep{TEACH} and VisualWebArena~\citep{koh2024visualwebarena} and for action forecasting in Ego4D~\citep{grauman2022ego4d} benchmarks. 
\subsection{Environments} \label{sec:envs}

\textbf{TEACh~\citep{TEACH}}
The TEACh dataset comprises over 3,000 dialogue-based instructions for household tasks in AI2-THOR~\citep{ai2thor}. We use the Trajectory from Dialogue (TfD) tasks where agents convert dialogue instructions into action sequences, such as \textsc{Make Coffee}. It includes training and validation splits (seen and unseen), the latter featuring new environments and instructions. Agents receive egocentric image inputs $o_t$ and perform actions like \texttt{pickup(X)} and \texttt{turn\_left()}. Task success is contingent on fulfilling all instruction conditions. Utilizing HELPER's~\citep{sarch2023helper} perception, navigation, and manipulation modules, the system relies on RGB images, depth maps, object masks, and egomotion for 3D mapping and object recognition. 
We remove domain-specific checks from HELPER's modules to allow ICAL to learn them independently.
\textbf{Noisy Trajectories.}
We use 250 noisy trajectories from TEACh, omitting action labels but retaining language instructions and corresponding RGB videos. To label actions from RGB video, we trained an inverse dynamics model using a transformer encoder-decoder based on the DETR architecture~\citep{carion2020end} from a seperate 300 TEACh episodes. Model predictions and human errors, like unnecessary movements, cause action noise in these demonstrations. 122 examples were successfully abstracted by \model{}.

\textbf{VisualWebArena~\citep{koh2024visualwebarena}}
VisualWebArena consists of 910 episodes across various web tasks (Classifieds, Shopping, Reddit) requiring visual comprehension and reasoning. Instructions may include text and reference images, like adding an item seen in an image to a wish list. Agents operate with instructions $I$, current webpage images, and an API for actions like \texttt{click(X)}, executing tasks to fulfill instruction conditions. \textbf{Noisy Trajectories.}
From VisualWebArena, 30 human demonstrations and 62 model trajectories from few-shot GPT4V were abstracted using \model{}. The process led to an example set of 92 for evaluation.

\textbf{Ego4D~\citep{grauman2022ego4d}}
This task involves anticipating actions from Ego4D RGB egocentric videos in daily scenarios. Models select from 115 verbs and 478 nouns for predicting actions. We evaluate using 200 unseen videos from ego4D validation, applying edit distance as a performance metric. Input to models includes sequences of video frames annotated with set-of-marks~\citep{yang2023setofmark} tracking~\citep{cheng2023tracking} and label masks. The supervised baseline~\citep{grauman2022ego4d} (243 video hrs of Ego4D V2) uses a SlowFast backbone with a Transformer aggregator. \textbf{Noisy Trajectories.}
Due to the passive nature of this task, \model{} proceeds without human-in-the-loop verification during \model{} (only Section~\ref{sec:introspection}, VLM-driven Abstraction Generation). \model{} successfully abstracted 92/100 demonstrations taken from the Ego4D validation set (8 failed due to GPT filters) for evaluation.

\subsection{\model{} beats expert-handwritten \& unchanged human demonstrations in household instruction following}

Table~\ref{tab:teach} presents our findings on the TEACh unseen validation set, assessing performance on new instructions, houses, and objects. \model{} and baselines use HELPER's navigation and manipulation modules~\citep{sarch2023helper}. We compare with these baselines: \textit{1. Hand-written examples} from HELPER, the SOTA on the TEACh benchmark, with 19 expert-written examples for retrieval-augmented prompting. \textit{2. Zero-shot chain of thought}, prompting the LLM to output step-by-step. \textit{3. Raw Visual Demos}, retrieving unchanged demonstrations labeled with the inverse dynamics model. \textit{4. Raw Kinesthetic Demos}, retrieving unchanged demonstrations with true actions. Our metrics are: \textit{1. Task success rate (SR)}, the \% of tasks completed successfully. \textit{2. Goal condition success rate (GC)}, the \% partial fulfillment rate across sessions.

\begin{wrapfigure}{r}{0.6\textwidth}
    \centering   \includegraphics[width=0.6\textwidth]{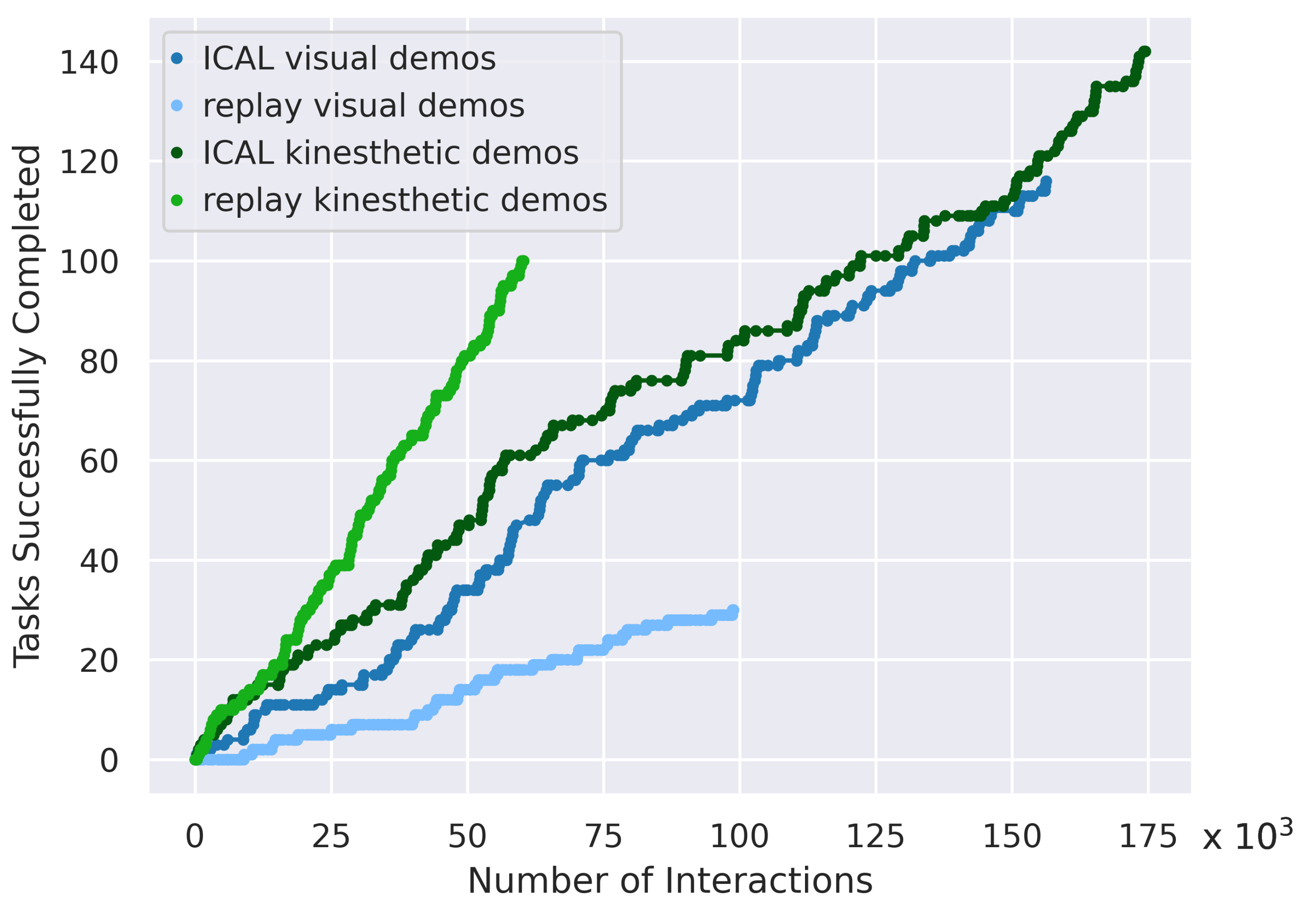}
    \vspace{-5mm}
    \caption{\textbf{\model{} enables greater success on training tasks.} Tasks successfully completed by \model{} over number of interactions when using the \model{} method with kinesthetic or visual demonstrations, and when replaying the kinesthetic or visual demonstrations directly. 
    }
    \label{fig:example_plot_gt}
\end{wrapfigure}

As shown in Figure~\ref{fig:example_plot_gt}, \model{} revises noisy trajectories, enabling more successful tasks completed on training tasks than mimicking raw human trajectories, with increases of 42 and 86 successful tasks for kinesthetic and visual demonstrations, respectively. This shows how \model{} not only adds useful programs of thought but also corrects errors in the passive video demos, improving success in the original demo environment. Please see the Appendix Section~\ref{app:num_examples} for additional analysis.

As shown in Table~\ref{tab:teach}, on unseen tasks, \model{} outperforms unprocessed demonstrations as in-context examples, achieving a 17.9\% absolute improvement in SR over raw demos with predicted actions and 8.6\% over those annotated with true actions. This underscores the effectiveness of our method in improving the quality of examples for in-context learning, unlike previous works that primarily save and retrieve successful action plans or trajectories without optimized actions and thoughts~\citep{sarch2023helper, wang2023voyager, liang2022code}.

Additionally, we surpass the handwritten examples of the previous SOTA HELPER~\citep{sarch2023helper} 
% by 12.6\% in GC and 0.6\% in SR (2.2\% (relative 26.5\%) using estimated perception)
by 12.6\% in GC and 0.6\% in SR, and by 2.2\% (relative 26.5\%) using estimated perception
, demonstrating our method's efficacy with less expert intervention, leveraging only visual demos and non-expert feedback. Unlike HELPER, which requires domain experts to write 48-107 lines of text for each example, ICAL does not rely on such extensive input from experts.
Instead, it allows non-experts to provide up to five natural language feedback corrections to the agent, significantly reducing the required effort and expertise per example.

Fine-tuning on ICAL examples also leads to significant improvements. We fine-tune GPT3.5-turbo-1106 model on the learned \model{} examples using the AzureAI interface (see the Appendix Section~\ref{app:finetune} for details). The training data include the 122 successfully abstracted examples by \model{}, which we randomly split into 99 training samples and 23 validation samples. This leads to an improvement of 11.4\% task success and 15.7\% goal-condition success for the GPT3.5 model. Combining the finetuned model with retrieval-augmented generation using the \model{} examples led to an additional improvement of 0.7\% task success and 4.9\% goal-condition success over using retrieval-augmented generation without finetuning: our top-performing agent. This demonstrates that consolidating the \model{} learned programs of thought with weight fine-tuning helps performance.   

\subsection{\model{} obtains state-of-the-art performance on visual web tasks}
We evaluate our agent with learned \model{} examples on the VisualWebArena evaluation set. We partition this into episodes `seen' by our model during learning, and those `unseen' during learning. 
%We used GPT4o for the evaluation.
% Due to changing versions of \texttt{gpt-4-1106-vision-preview}, we re-evaluated (February 2024) the best VisualWebArena baseline using their provided code repository. We present our findings based on a reduced evaluation set, which consists of a random 488 episodes, representing half of the evaluation.

% Results on VisualWebArena are shown in Table~\ref{tab:vwa}. \model{} outperforms the previous state-of-the-art~\citep{koh2024visualwebarena}, utilizing GPT4o with few-shot hand-designed examples and set of marks~\citep{yang2023setofmark} image prompting, by a relative 23.8\% in average success rate. 

Table~\ref{tab:vwa} presents the results on VisualWebArena. Our model, \model{}, using retrieval-augmented generation with GPT4V outperforms the previous state-of-the-art~\cite{koh2024visualwebarena}, which uses GPT4V with few-shot, hand-designed examples and set-of-marks image prompting~\cite{yang2023setofmark}. \model{} achieves an absolute 8.4\% (relative 58.7\%) improvement in average success rate over GPT4V and shows a 23.8\% relative improvement in average success rate over GPT4o. Fine-tuning Qwen2-VL-7B on ICAL examples yields a 5.3\% absolute (182.7\% relative) improvement in average success rate, achieving state-of-the-art performance on VWA for its model size.
%When fine-tuning Qwen2-VL-7b on the ICAL examples, we observe an absolute 5.3\% (relative 182.76\%) improvement in average success rate over the base Qwen2-VL-7b model, achieving state-of-the-art on VWA for its model size.

\subsection{\model{} outperforms few-shot VLMs on egocentric video action forecasting}
We test ICAL on video action forecasting without using human-in-the-loop verification due to the passive nature of the task. As shown in Table~\ref{tab:ego4d}, \model{} demonstrates superior few-shot performance on Ego4D action anticipation compared to hand-written few-shot GPT4V that uses chain of thought~\citep{wei2022chain}, improving by 6.4 noun and 1.7 action edit distance. \model{} also remains competitive with the fully supervised baseline~\citep{grauman2022ego4d} in noun and action prediction despite using 639x less in-domain training data. We find GPT4V video processing to have the least improvements for verb action prediction, possibly due to its limited video understanding capabilities.

\begin{table}[h!]
\footnotesize
\setlength{\tabcolsep}{2pt} % Adjusting the column separation
\begin{minipage}{.48\linewidth}
\begin{center}
\footnotesize
\setlength{\tabcolsep}{2pt} % Adjusting the column separation
\caption{\textbf{Evaluation on TEACh unseen validation set.} All evaluations are done using GPT3.5-turbo-1106 unless otherwise noted. Visual Demos = demonstrations labeled with inverse dynamics model. Kinesthetic Demos = demos labeled with GT actions. GC = goal-condition success}\label{tab:teach}\vspace{-5mm}
\begin{tabular}{@{}lcc@{}}
 \\
 & \multicolumn{1}{c}{Success} & \multicolumn{1}{c}{GC} \\
\addlinespace[0.15cm] % Adjust the value as needed
\hline
\addlinespace[0.15cm] % Adjust the value as needed
\textit{Ground truth segm, depth, attributes} & & \\
 HELPER hand-written~\citep{sarch2023helper} & 34.5 & 36.7 \\
 Zero-Shot CoT~\citep{kojima2022large} & 11.8 & 24.6 \\
 Raw Visual Demos & 17.2 & 26.6 \\
 Raw Kinesthetic Demos & 26.5 & 29.5 \\
 \model{} retrieval (ours) & \textbf{35.1} & \textbf{49.3} \\
 \hspace{0.5mm} w/o programs of thought phase & 29.4 & 44.9 \\
 \hspace{0.5mm} w/o human-in-the-loop & 29.9 & 41.0 \\
 \hspace{0.5mm} w/ retrieval re-ranking & \textbf{35.3} & \textbf{51.7} \\
 \hspace{0.5mm} w/ GPT4 & 41.7 & 63.6 \\
 \addlinespace[0.15cm] 
 \addlinespace[0.15cm] 
 \hspace{0.5mm} \model{} SFT & 23.2 & 40.3 \\
 \hspace{0.5mm} \model{} SFT + retrieval & \textbf{35.8} & \textbf{54.2} \\
 \addlinespace[0.15cm] % Adjust the value as needed
 \hline
 \addlinespace[0.15cm]
 \textit{Estimated perception} & & \\
HELPER hand-written~\citep{sarch2023helper} & 8.3 & 14.1 \\
\model{} (ours) & \textbf{10.5} & \textbf{15.4} \\
\addlinespace[0.15cm]
 \hline
 \addlinespace[0.15cm] % Adjust the value as needed
\end{tabular}
\label{fig:continual}
\end{center}
\end{minipage}%
\quad % Spacing between tables
\begin{minipage}{.48\linewidth}
\begin{center}
\footnotesize
\setlength{\tabcolsep}{2pt} % Adjusting the column separation
\caption{
\textbf{Results on VisualWebArena.} \model{} outperforms the prior best, GPT4o/V + Set of Marks. All VLM baselines are given Image + SoM + Captions representation (see \citet{koh2024visualwebarena}). Ablation studies were conducted with GPT4V on a subset of 257 episodes. $^*$Run on half of eval set.
}\label{tab:vwa}\vspace{-5mm}
\begin{tabular}{@{}llccc@{}}
 \\
 & \multicolumn{1}{c}{Seen} & \multicolumn{1}{c}{Unseen} & \multicolumn{1}{c}{Average} \\
\addlinespace[0.15cm] % Adjust the value as needed
\hline
\addlinespace[0.15cm] % Adjust the value as needed
 % GPT4V+SoM~\citep{koh2024visualwebarena} &  16.3 & 14.1 & 14.3 \\
 % ICAL (ours) & \textbf{38.8} & \textbf{20.9} & \textbf{22.7} \\
 % \hline
 \textit{Open-source VLMs}\\
 CogVLM~\cite{wang2023cogvlm,koh2024visualwebarena} & -- & -- & 0.33 \\
 IDEFICS-80B-Instruct~\cite{laurencon2023obelics,koh2024visualwebarena} & -- & -- & 0.99 \\
 Qwen2-VL-7B~\citep{Qwen2VL} & -- & -- & 2.9 \\
 ICAL Qwen2-VL-7B SFT (ours) & 16.7 & 7.4 & \textbf{8.2} \\
 \addlinespace[0.15cm] % Adjust the value as needed
 \textit{Proprietary VLMs}\\
 Gemini-Pro-1.5 &  -- & -- & 11.9 \\
 GPT4o~\citep{koh2024visualwebarena} &  -- & -- & 18.9 \\
 ICAL GPT4o (ours) & 32.3 & 22.3 & \textbf{23.4} \\
% \addlinespace[0.15cm] % Adjust the value as needed
% \hline
% \addlinespace[0.15cm] % Adjust the value as needed
GPT4V$^*$~\citep{koh2024visualwebarena} &  16.3 & 14.1 & 14.3 \\
ICAL GPT4V$^*$ (ours) & 38.8 & 20.9 & \textbf{22.7} \\
\addlinespace[0.15cm] % Adjust the value as needed
\textit{Ablations} \\
GPT4V~\citep{koh2024visualwebarena} & 11.5 & 12.9 & 12.7 \\
\model{} (ours) & 28.0 & 21.6 & 22.2 \\ 
\hspace{0.5mm} w/o image & 28.0 & 17.3 & 19.0 \\ 
\hspace{0.5mm} w/ full text trajectory & 57.7 & 21.6 & 25.5 \\
\addlinespace[0.15cm] % Adjust the value as needed
 \hline
\end{tabular}
\end{center}

% \begin{center}
% \caption{
% \textbf{Evaluation on the Ego4D unseen validation subset.}  ICAL outperforms few-shot GPT4V and matches supervised baselines using 639x less in-domain data.
% }\label{tab:ego4d}\vspace{-5mm}
% \begin{tabular}{@{}lcccccc@{}}
%  \\
%  & \multicolumn{3}{c}{ED@(Z=20)} \\
%  % & \multicolumn{3}{c}{Unseen} \\
%  \cmidrule(lr){2-4}
%  & \multicolumn{1}{c}{Verb} & \multicolumn{1}{c}{Noun} & \multicolumn{1}{c}{Action} \\
% \addlinespace[0.15cm] % Adjust the value as needed
% \hline
% \addlinespace[0.15cm] % Adjust the value as needed
%  \textcolor{mediumgray}{Supervised~\citep{grauman2022ego4d}} & \textcolor{mediumgray}{0.7251} & \textcolor{mediumgray}{0.7393} & \textcolor{mediumgray}{0.9235} \\
%  \hspace{0.5mm} \textcolor{mediumgray}{(639x more data)} \\
%  Zero-shot CoT~\citep{kojima2022large} & 0.8796 & 0.7930 & 0.9639 \\
%  Few-shot CoT & 0.7877 & 0.7575 & 0.9414 \\
%  \model{} (ours) & \textbf{0.7802} & \textbf{0.6934} & \textbf{0.9242} \\ 
%  \addlinespace[0.15cm] % Adjust the value as needed
%  \hline
%  \addlinespace[0.15cm] % Adjust the value as needed
% \end{tabular}
% \end{center}
\end{minipage}
\end{table}

\begin{table}
\begin{center}
\caption{
\textbf{Evaluation on the Ego4D unseen validation subset.}  ICAL outperforms few-shot GPT4V and matches supervised baselines using 639x less in-domain data.
}\label{tab:ego4d} %\vspace{-5mm}
\begin{tabular}{@{}lcccccc@{}}
 \\
 & \multicolumn{3}{c}{ED@(Z=20)} \\
 % & \multicolumn{3}{c}{Unseen} \\
 \cmidrule(lr){2-4}
 & \multicolumn{1}{c}{Verb} & \multicolumn{1}{c}{Noun} & \multicolumn{1}{c}{Action} \\
\addlinespace[0.15cm] % Adjust the value as needed
\hline
\addlinespace[0.15cm] % Adjust the value as needed
 \textcolor{mediumgray}{Supervised~\citep{grauman2022ego4d}} & \textcolor{mediumgray}{0.7251} & \textcolor{mediumgray}{0.7393} & \textcolor{mediumgray}{0.9235} \\
 \hspace{0.5mm} \textcolor{mediumgray}{(639x more data)} \\
 Zero-shot CoT~\citep{kojima2022large} & 0.8796 & 0.7930 & 0.9639 \\
 Few-shot CoT & 0.7877 & 0.7575 & 0.9414 \\
 \model{} (ours) & \textbf{0.7802} & \textbf{0.6934} & \textbf{0.9242} \\ 
 \addlinespace[0.15cm] % Adjust the value as needed
 \hline
 \addlinespace[0.15cm] % Adjust the value as needed
\end{tabular}
\end{center}
\end{table}

\clearpage

\subsection{\model{} shows better scaling than human demonstrations}

\begin{wrapfigure}{r}{0.5\textwidth}
    \centering
    \vspace{-.5cm}
    \includegraphics[width=0.5\textwidth]{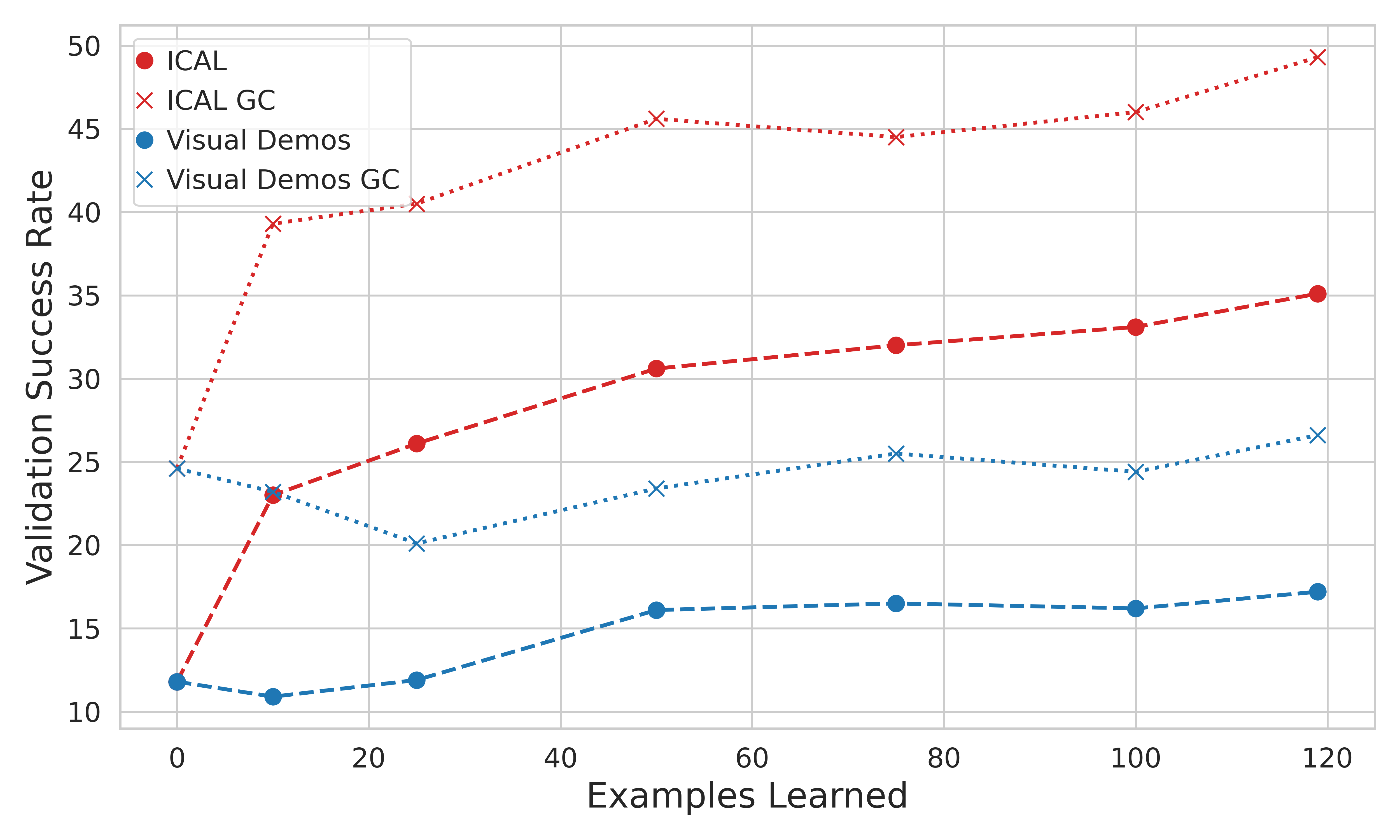}
    \vspace{-.7cm}
    \caption{\textbf{ TEACh validation unseen success rate} for ICAL with increasing number of exemplars. \model{} continually learns without forgetting, scaling 2x better than the unchanged visual human demos used to seed ICAL learning. $\CIRCLE$ denotes task success, while \textbf{x} denotes goal-condition success.
    }
    \label{fig:continual}
    % \vspace{-2mm}
\end{wrapfigure}

ICAL demonstrates twice the scaling performance of human demonstration data on TEACh validation unseen tasks, as shown in Figure~\ref{fig:continual}. Compared to the unchanged visual human demos used to seed ICAL learning, using ICAL data for in-context learning obtains a 1.9x performance gain at 50 examples and 2.1x performance at 120 examples. Importantly, throughout learning, \model{} does not need to worry about forgetting previously learned knowledge since the agent is expanding a memory of examples with a frozen VLM. Also noteworthy, our method benefits from even a small amount of examples learned, with an improvement of an absolute 14.7\% success rate over zero-shot chain-of-thought~\citep{kojima2022large} prompting and 6.8\% over the unchanged demonstrations (with 10x less data) with just 10 ICAL demonstrations, further showing the efficiency of our method.

\subsection{Meta-learning: ICAL example retrieval improves learning efficiency}\label{sec:efficiency}

% \begin{wrapfigure}{r}{0.55\textwidth}
%     \centering
%     \includegraphics[width=0.2\textwidth]{Figures/stats_continual_learning_overall_0_2.png}
%     \includegraphics[width=0.25\textwidth]{Figures/stats_continual_learning_tasks_0_2.png}
%     % \hspace{0.02\textwidth}
%     % \includegraphics[width=0.3\textwidth]{figures/stats_continual_learning_tasks_0_2.png}
%     \caption{\small First half (blue) versus second half (orange) of ICAL learning across tasks \textbf{(left)} and for each task type separately \textbf{(right)} in TEACh. The second half of ICAL learning requires significantly fewer environment steps (436±88 vs. 267±43, p=0.0143) and human feedbacks per episode (0.74±0.17 vs. 0.21±0.08, p=0.0089). This indicates that retrieving ICAL examples during learning is beneficial, reducing both human effort and environment interaction over time.}
%     \label{fig:continual}
%     \vspace{-2mm}
% \end{wrapfigure}

\begin{figure}[ht!]
\centering
\includegraphics[width=0.4\textwidth]{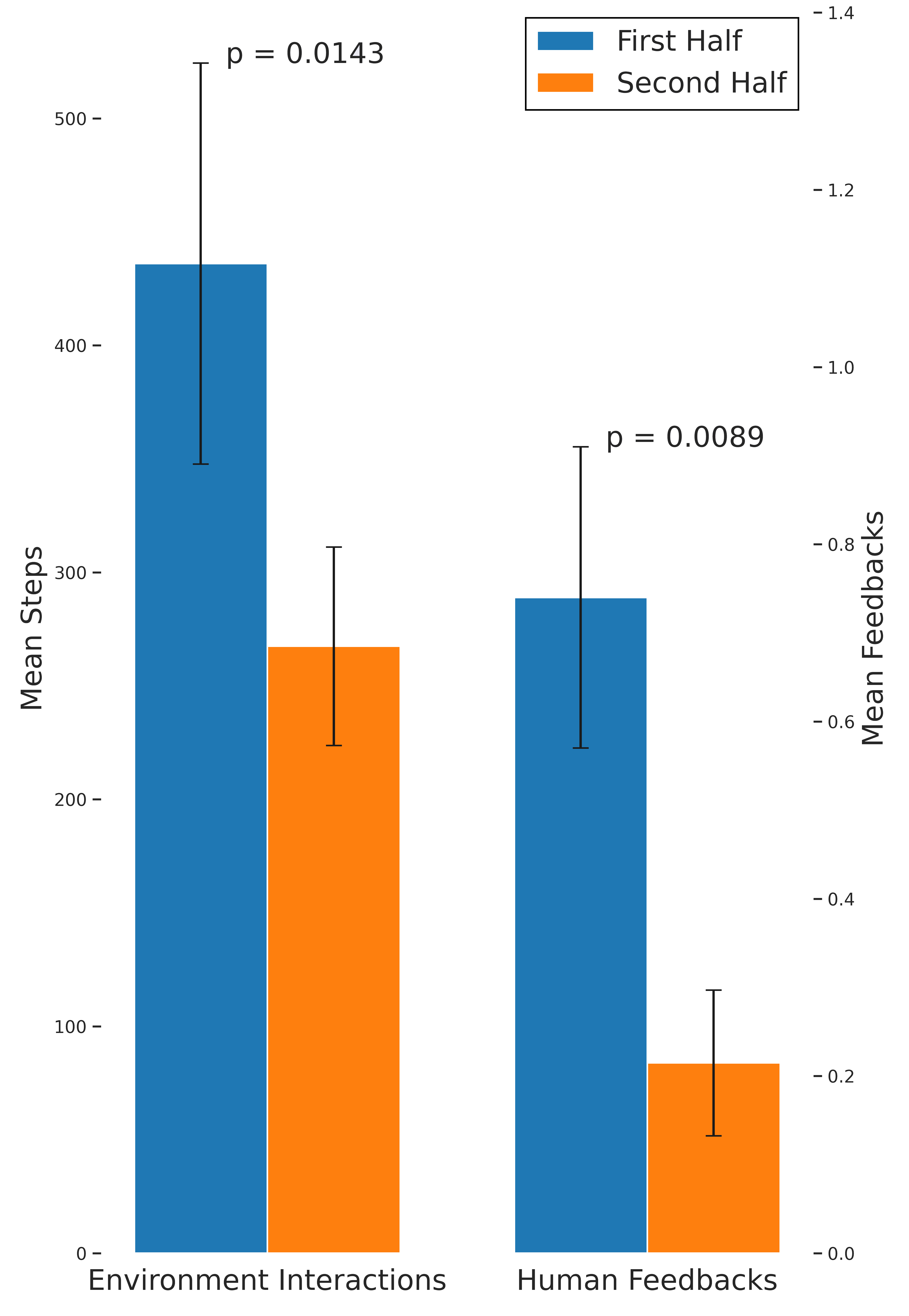}
% \hspace{0.02\textwidth}
% \includegraphics[width=0.55\textwidth]{Figures/stats_continual_learning_tasks_0_2.png}
\caption{\textbf{ICAL improves learning efficiency as more examples are added to memory.} \small First half (blue) versus second half (orange) of ICAL learning across tasks \textbf{(left)} and for each task type separately \textbf{(right)} in TEACh. The second half of ICAL learning requires significantly fewer environment steps (436±88 vs. 267±43, p=0.0143) and human feedbacks per episode (0.74±0.17 vs. 0.21±0.08, p=0.0089). This indicates that retrieving ICAL examples during learning is beneficial, reducing both human effort and environment interaction over time.}
\label{figR1}
\end{figure}

Efficient learning systems benefit greatly from leveraging past knowledge, allowing them to reduce the need for human intervention and environment interactions as they continue to process new data. Our agent becomes increasingly efficient over time, requiring less human feedback and fewer environment interactions as it processes more examples. By retrieving past successful examples during the VLM-driven programs of thought generation and human-in-the-loop phases, it uses previously stored knowledge to help optimize new examples. As shown in Figure~\ref{figR1}, for the second half of examples processed in TEACh, the model requires significantly fewer environment steps (436±88 vs. 267±43, p=0.0143) and human feedbacks (0.74±0.17 vs. 0.21±0.08, p=0.0089) per example. This demonstrates that retrieving previously optimized examples during learning of new examples reduces both human effort and environment interaction over time. Put another way, using previously stored ICAL examples not only improves test performance but also accelerates learning for future examples.

% \subsection{Fine-tuning helps}

% We fine-tune the GPT3.5-turbo-1106 model on the learned \model{} examples in TEACh in the AzureAI interface (see the Appendix Section~\ref{app:finetune} for details). The training data include the 122 successfully abstracted examples by \model{}, which we randomly split into 99 training samples and 23 validation samples. This leads to an improvement of 11.4\% task success and 15.7\% goal-condition success for the GPT3.5 model. Combining the finetuned model with retrieval-augmented generation using the \model{} examples led to an additional improvement of 0.7\% task success and 4.9\% goal-condition success over using retrieval-augmented generation without finetuning: our top-performing agent. This demonstrates that consolidating the \model{} learned programs of thought with weight fine-tuning helps performance.   

\subsection{Ablations show each component of \model{} is important}
We ablate the  components of \model{} in TEACh in Table~\ref{tab:teach}. We conclude:

\noindent 1. The VLM-driven programs of thought generation phase significantly helps for refining the trajectories and adding generalizable knowledge. We observe a decrease in 5.7\% success rate and 4.4\% in goal condition success rate when removing the  VLM-driven programs of thought generation phase. 

\vspace{-2mm}
\noindent 2. The human-in-the-loop phase is important for fixing errors and incorporating feedback from the user. We observe a decrease in 5.2\% success rate and 8.3\% in goal condition success rate when removing the human-in-the-loop phase. 

\vspace{-2mm}
\noindent 3. Our examples demonstrate scalability with larger LLMs. GPT-4 showed a 6.6\% absolute increase in task success and a 14.3\% absolute rise in goal condition success compared to GPT-3.5.

\vspace{-2mm}
\noindent 4. ICAL can be combined with advanced prompting and sampling methods. We test this using re-ranking~\citep{wang2022self}, where the model generates three diverse outputs from different retrieved examples (e.g., top 1-5, 6-10, ...), self-evaluates, and selects the highest scoring output. Improvements are modest but notable: 0.2\% in task success and 2.5\% in goal condition success.

\clearpage

% \begin{wrapfigure}{r}{0.45\textwidth}
% \end{wrapfigure}
\vspace{6mm}

\section{Conclusion}
We presented ICAL, a method that improves the quality of agent trajectory examples by refining actions and thoughts via VLM self-refinement and human-in-the-loop feedback. When the resulting data is used as in-context examples or for supervised fine-tuning, performance of the VLM agent improves significantly over using raw human demonstrations. It also reduces the need for expert examples, and enables more efficient learning as more examples are learned. Tested in TEACh, VisualWebArena, and Ego4D, ICAL achieves state-of-the-art performance, demonstrating adaptability to new tasks and environments. There are several limitations and future research directions for ICAL. While ICAL can handle noisy demos and trajectories, ICAL may not be able to handle extremely misleading demonstrations or feedback, and relies on a fixed action API which may restrict adaptability in dynamic environments. Additionally, GPT4V's visual grounding deficiencies~\citep{zheng2024gpt, wu2023textit, OpenEQA2023, chen2024spatialvlm} cannot always be overcome by in-context learning, and more research is needed to address this.

\clearpage

\paragraph{Acknowledgements}
This material is based upon work supported by National Science Foundation grants GRF DGE1745016 \& DGE2140739 (GS), ONR award N00014-23-1-2415, AFOSR Grant FA9550-23-1-0257, and DARPA No. HR00112490375 from the U.S. DARPA Friction for Accountability in Conversational Transactions (FACT) program. Any opinions, findings and conclusions or recommendations expressed in this material are those of the authors and do not necessarily reflect the views of the United States Army, the National Science Foundation, or the United States Air Force. 

This research project has benefitted from the Microsoft Accelerate Foundation Models Research (AFMR) grant program through which leading foundation models hosted by Microsoft Azure along with access to Azure credits were provided to conduct the research.

\bibliographystyle{plainnat}
\bibliography{main}

\begin{thebibliography}{96}
\providecommand{\natexlab}[1]{#1}
\providecommand{\url}[1]{\texttt{#1}}
\expandafter\ifx\csname urlstyle\endcsname\relax
  \providecommand{\doi}[1]{doi: #1}\else
  \providecommand{\doi}{doi: \begingroup \urlstyle{rm}\Url}\fi

\bibitem[gpt(2023)]{gpt4technical}
Openai. gpt-4 technical report.
\newblock \emph{arXiv preprint arxiv:2303.08774}, 2023.

\bibitem[Awadalla et~al.(2023)Awadalla, Gao, Gardner, Hessel, Hanafy, Zhu, Marathe, Bitton, Gadre, Sagawa, Jitsev, Kornblith, Koh, Ilharco, Wortsman, and Schmidt]{awadalla2023openflamingo}
Anas Awadalla, Irena Gao, Josh Gardner, Jack Hessel, Yusuf Hanafy, Wanrong Zhu, Kalyani Marathe, Yonatan Bitton, Samir Gadre, Shiori Sagawa, Jenia Jitsev, Simon Kornblith, Pang~Wei Koh, Gabriel Ilharco, Mitchell Wortsman, and Ludwig Schmidt.
\newblock Openflamingo: An open-source framework for training large autoregressive vision-language models.
\newblock \emph{arXiv preprint arXiv:2308.01390}, 2023.

\bibitem[Bahdanau et~al.(2018)Bahdanau, Hill, Leike, Hughes, Hosseini, Kohli, and Grefenstette]{bahdanau2018learning}
Dzmitry Bahdanau, Felix Hill, Jan Leike, Edward Hughes, Arian Hosseini, Pushmeet Kohli, and Edward Grefenstette.
\newblock Learning to understand goal specifications by modelling reward.
\newblock \emph{arXiv preprint arXiv:1806.01946}, 2018.

\bibitem[Barrett and Bar(2009)]{barrett2009see}
Lisa~Feldman Barrett and Moshe Bar.
\newblock See it with feeling: affective predictions during object perception.
\newblock \emph{Philosophical Transactions of the Royal Society B: Biological Sciences}, 364\penalty0 (1521):\penalty0 1325--1334, 2009.

\bibitem[Bhat et~al.(2023)Bhat, Birkl, Wofk, Wonka, and M{\"u}ller]{bhat2023zoedepth}
Shariq~Farooq Bhat, Reiner Birkl, Diana Wofk, Peter Wonka, and Matthias M{\"u}ller.
\newblock Zoedepth: Zero-shot transfer by combining relative and metric depth.
\newblock \emph{arXiv preprint arXiv:2302.12288}, 2023.

\bibitem[Botvinick et~al.(2009)Botvinick, Niv, and Barto]{botvinick2009hierarchically}
Matthew~M Botvinick, Yael Niv, and Andew~G Barto.
\newblock Hierarchically organized behavior and its neural foundations: A reinforcement learning perspective.
\newblock \emph{cognition}, 113\penalty0 (3):\penalty0 262--280, 2009.

\bibitem[Carion et~al.(2020)Carion, Massa, Synnaeve, Usunier, Kirillov, and Zagoruyko]{carion2020end}
Nicolas Carion, Francisco Massa, Gabriel Synnaeve, Nicolas Usunier, Alexander Kirillov, and Sergey Zagoruyko.
\newblock End-to-end object detection with transformers.
\newblock In \emph{European conference on computer vision}, pages 213--229. Springer, 2020.

\bibitem[Chaplot et~al.(2019)Chaplot, Gandhi, Gupta, Gupta, and Salakhutdinov]{chaplot2020learning}
Devendra~Singh Chaplot, Dhiraj Gandhi, Saurabh Gupta, Abhinav Gupta, and Ruslan Salakhutdinov.
\newblock Learning to explore using active neural slam.
\newblock In \emph{International Conference on Learning Representations}, 2019.

\bibitem[Chen et~al.(2024)Chen, Xu, Kirmani, Ichter, Driess, Florence, Sadigh, Guibas, and Xia]{chen2024spatialvlm}
Boyuan Chen, Zhuo Xu, Sean Kirmani, Brian Ichter, Danny Driess, Pete Florence, Dorsa Sadigh, Leonidas Guibas, and Fei Xia.
\newblock Spatialvlm: Endowing vision-language models with spatial reasoning capabilities.
\newblock \emph{arXiv preprint arXiv:2401.12168}, 2024.
\newblock URL \url{https://arxiv.org/abs/2401.12168}.

\bibitem[Chen et~al.(2020)Chen, Jain, Schissler, Gari, Al-Halah, Ithapu, Robinson, and Grauman]{chen2020soundspaces}
Changan Chen, Unnat Jain, Carl Schissler, Sebastia Vicenc~Amengual Gari, Ziad Al-Halah, Vamsi~Krishna Ithapu, Philip Robinson, and Kristen Grauman.
\newblock Soundspaces: Audio-visual navigation in 3d environments.
\newblock In \emph{Computer Vision--ECCV 2020: 16th European Conference, Glasgow, UK, August 23--28, 2020, Proceedings, Part VI 16}, pages 17--36. Springer, 2020.

\bibitem[Cheng et~al.(2023{\natexlab{a}})Cheng, Kolobov, Misra, Nie, and Swaminathan]{cheng2023llf}
Ching-An Cheng, Andrey Kolobov, Dipendra Misra, Allen Nie, and Adith Swaminathan.
\newblock Llf-bench: Benchmark for interactive learning from language feedback.
\newblock \emph{arXiv preprint arXiv:2312.06853}, 2023{\natexlab{a}}.

\bibitem[Cheng et~al.(2023{\natexlab{b}})Cheng, Oh, Price, Schwing, and Lee]{cheng2023tracking}
Ho~Kei Cheng, Seoung~Wug Oh, Brian Price, Alexander Schwing, and Joon-Young Lee.
\newblock Tracking anything with decoupled video segmentation.
\newblock In \emph{ICCV}, 2023{\natexlab{b}}.

\bibitem[Cherti et~al.(2022)Cherti, Beaumont, Wightman, Wortsman, Ilharco, Gordon, Schuhmann, Schmidt, and Jitsev]{cherti2022reproducible}
Mehdi Cherti, Romain Beaumont, Ross Wightman, Mitchell Wortsman, Gabriel Ilharco, Cade Gordon, Christoph Schuhmann, Ludwig Schmidt, and Jenia Jitsev.
\newblock Reproducible scaling laws for contrastive language-image learning.
\newblock \emph{arXiv preprint arXiv:2212.07143}, 2022.

\bibitem[Cui et~al.(2023)Cui, Karamcheti, Palleti, Shivakumar, Liang, and Sadigh]{cui2023no}
Yuchen Cui, Siddharth Karamcheti, Raj Palleti, Nidhya Shivakumar, Percy Liang, and Dorsa Sadigh.
\newblock No, to the right: Online language corrections for robotic manipulation via shared autonomy.
\newblock In \emph{Proceedings of the 2023 ACM/IEEE International Conference on Human-Robot Interaction}, pages 93--101, 2023.

\bibitem[Dai et~al.(2023)Dai, Peng, Li, and Chai]{dai2023think}
Yinpei Dai, Run Peng, Sikai Li, and Joyce Chai.
\newblock Think, act, and ask: Open-world interactive personalized robot navigation.
\newblock \emph{arXiv preprint arXiv:2310.07968}, 2023.

\bibitem[Das et~al.(2018)Das, Datta, Gkioxari, Lee, Parikh, and Batra]{das2018embodied}
Abhishek Das, Samyak Datta, Georgia Gkioxari, Stefan Lee, Devi Parikh, and Dhruv Batra.
\newblock Embodied question answering.
\newblock In \emph{Proceedings of the IEEE Conference on Computer Vision and Pattern Recognition}, pages 1--10, 2018.

\bibitem[Das et~al.(2020)Das, Carnevale, Merzic, Rimell, Schneider, Abramson, Hung, Ahuja, Clark, Wayne, et~al.]{das2020probing}
Abhishek Das, Federico Carnevale, Hamza Merzic, Laura Rimell, Rosalia Schneider, Josh Abramson, Alden Hung, Arun Ahuja, Stephen Clark, Gregory Wayne, et~al.
\newblock Probing emergent semantics in predictive agents via question answering.
\newblock In \emph{Proceedings of the 37th International Conference on Machine Learning}, pages 2376--2391, 2020.

\bibitem[Datta et~al.(2022)Datta, Dharur, Cartillier, Desai, Khanna, Batra, and Parikh]{datta2022episodic}
Samyak Datta, Sameer Dharur, Vincent Cartillier, Ruta Desai, Mukul Khanna, Dhruv Batra, and Devi Parikh.
\newblock Episodic memory question answering.
\newblock In \emph{Proceedings of the IEEE/CVF Conference on Computer Vision and Pattern Recognition}, pages 19119--19128, 2022.

\bibitem[Deng et~al.(2023)Deng, Gu, Zheng, Chen, Stevens, Wang, Sun, and Su]{deng2023mind2web}
Xiang Deng, Yu~Gu, Boyuan Zheng, Shijie Chen, Samuel Stevens, Boshi Wang, Huan Sun, and Yu~Su.
\newblock Mind2web: Towards a generalist agent for the web, 2023.

\bibitem[Dong et~al.(2021)Dong, Zeng, Wang, Zhang, and Wei]{dong2021solq}
Bin Dong, Fangao Zeng, Tiancai Wang, Xiangyu Zhang, and Yichen Wei.
\newblock Solq: Segmenting objects by learning queries.
\newblock \emph{Advances in Neural Information Processing Systems}, 34:\penalty0 21898--21909, 2021.

\bibitem[Fu et~al.(2019)Fu, Korattikara, Levine, and Guadarrama]{fu2019language}
Justin Fu, Anoop Korattikara, Sergey Levine, and Sergio Guadarrama.
\newblock From language to goals: Inverse reinforcement learning for vision-based instruction following.
\newblock \emph{arXiv preprint arXiv:1902.07742}, 2019.

\bibitem[Gao et~al.(2023)Gao, Thattai, Gao, Shakiah, Pansare, Sharma, Sukhatme, Shi, Yang, Zheng, et~al.]{gao2023alexa}
Qiaozi Gao, Govind Thattai, Xiaofeng Gao, Suhaila Shakiah, Shreyas Pansare, Vasu Sharma, Gaurav Sukhatme, Hangjie Shi, Bofei Yang, Desheng Zheng, et~al.
\newblock Alexa arena: A user-centric interactive platform for embodied ai.
\newblock \emph{arXiv preprint arXiv:2303.01586}, 2023.

\bibitem[Gao et~al.(2022)Gao, Gao, Gong, Lin, Thattai, and Sukhatme]{gao2022dialfred}
Xiaofeng Gao, Qiaozi Gao, Ran Gong, Kaixiang Lin, Govind Thattai, and Gaurav~S Sukhatme.
\newblock Dial{FRED}: Dialogue-enabled agents for embodied instruction following.
\newblock \emph{IEEE Robotics and Automation Letters}, 7\penalty0 (4):\penalty0 10049--10056, 2022.

\bibitem[Goodman and Frank(2016)]{goodman2016pragmatic}
Noah~D Goodman and Michael~C Frank.
\newblock Pragmatic language interpretation as probabilistic inference.
\newblock \emph{Trends in cognitive sciences}, 20\penalty0 (11):\penalty0 818--829, 2016.

\bibitem[Gordon et~al.(2018)Gordon, Kembhavi, Rastegari, Redmon, Fox, and Farhadi]{gordon2018iqa}
Daniel Gordon, Aniruddha Kembhavi, Mohammad Rastegari, Joseph Redmon, Dieter Fox, and Ali Farhadi.
\newblock Iqa: Visual question answering in interactive environments.
\newblock In \emph{Proceedings of the IEEE conference on computer vision and pattern recognition}, pages 4089--4098, 2018.

\bibitem[Goyal et~al.(2019)Goyal, Niekum, and Mooney]{goyal2019using}
Prasoon Goyal, Scott Niekum, and Raymond~J Mooney.
\newblock Using natural language for reward shaping in reinforcement learning.
\newblock \emph{arXiv preprint arXiv:1903.02020}, 2019.

\bibitem[Goyal et~al.(2021)Goyal, Niekum, and Mooney]{goyal2021pixl2r}
Prasoon Goyal, Scott Niekum, and Raymond Mooney.
\newblock Pixl2r: Guiding reinforcement learning using natural language by mapping pixels to rewards.
\newblock In \emph{Conference on Robot Learning}, pages 485--497. PMLR, 2021.

\bibitem[Grauman et~al.(2022)Grauman, Westbury, Byrne, Chavis, Furnari, Girdhar, Hamburger, Jiang, Liu, Liu, et~al.]{grauman2022ego4d}
Kristen Grauman, Andrew Westbury, Eugene Byrne, Zachary Chavis, Antonino Furnari, Rohit Girdhar, Jackson Hamburger, Hao Jiang, Miao Liu, Xingyu Liu, et~al.
\newblock Ego4d: Around the world in 3,000 hours of egocentric video.
\newblock In \emph{Proceedings of the IEEE/CVF Conference on Computer Vision and Pattern Recognition}, pages 18995--19012, 2022.

\bibitem[Greene et~al.(2022)Greene, Sanders, Weng, and Neelakantan]{ada003technical}
Ryan Greene, Ted Sanders, Lilian Weng, and Arvind Neelakantan.
\newblock Openai. new and improved embedding model.
\newblock 2022.

\bibitem[Harrison et~al.(2017)Harrison, Ehsan, and Riedl]{harrison2017guiding}
Brent Harrison, Upol Ehsan, and Mark~O Riedl.
\newblock Guiding reinforcement learning exploration using natural language.
\newblock \emph{arXiv preprint arXiv:1707.08616}, 2017.

\bibitem[Ho et~al.(2022)Ho, Abel, Correa, Littman, Cohen, and Griffiths]{ho2022people}
Mark~K Ho, David Abel, Carlos~G Correa, Michael~L Littman, Jonathan~D Cohen, and Thomas~L Griffiths.
\newblock People construct simplified mental representations to plan.
\newblock \emph{Nature}, 606\penalty0 (7912):\penalty0 129--136, 2022.

\bibitem[Hu et~al.(2021)Hu, Shen, Wallis, Allen-Zhu, Li, Wang, Wang, and Chen]{hu2021lora}
Edward~J Hu, Yelong Shen, Phillip Wallis, Zeyuan Allen-Zhu, Yuanzhi Li, Shean Wang, Lu~Wang, and Weizhu Chen.
\newblock Lora: Low-rank adaptation of large language models.
\newblock \emph{arXiv preprint arXiv:2106.09685}, 2021.

\bibitem[Jain et~al.(2024)Jain, Katara, Gkanatsios, Harley, Sarch, Aggarwal, Chaudhary, and Fragkiadaki]{jain2024odin}
Ayush Jain, Pushkal Katara, Nikolaos Gkanatsios, Adam~W. Harley, Gabriel Sarch, Kriti Aggarwal, Vishrav Chaudhary, and Katerina Fragkiadaki.
\newblock Odin: A single model for 2d and 3d perception, 2024.

\bibitem[Jimenez et~al.(2023)Jimenez, Yang, Wettig, Yao, Pei, Press, and Narasimhan]{jimenez2023swebench}
Carlos~E. Jimenez, John Yang, Alexander Wettig, Shunyu Yao, Kexin Pei, Ofir Press, and Karthik Narasimhan.
\newblock Swe-bench: Can language models resolve real-world github issues?, 2023.

\bibitem[Katara et~al.(2023)Katara, Xian, and Fragkiadaki]{katara2023gen2sim}
Pushkal Katara, Zhou Xian, and Katerina Fragkiadaki.
\newblock Gen2sim: Scaling up robot learning in simulation with generative models, 2023.

\bibitem[Klissarov et~al.(2023)Klissarov, D'Oro, Sodhani, Raileanu, Bacon, Vincent, Zhang, and Henaff]{klissarov2023motif}
Martin Klissarov, Pierluca D'Oro, Shagun Sodhani, Roberta Raileanu, Pierre-Luc Bacon, Pascal Vincent, Amy Zhang, and Mikael Henaff.
\newblock Motif: Intrinsic motivation from artificial intelligence feedback.
\newblock \emph{arXiv preprint arXiv:2310.00166}, 2023.

\bibitem[Koh et~al.(2024{\natexlab{a}})Koh, Lo, Jang, Duvvur, Lim, Huang, Neubig, Zhou, Salakhutdinov, and Fried]{koh2024visualwebarena}
Jing~Yu Koh, Robert Lo, Lawrence Jang, Vikram Duvvur, Ming~Chong Lim, Po-Yu Huang, Graham Neubig, Shuyan Zhou, Ruslan Salakhutdinov, and Daniel Fried.
\newblock Visualwebarena: Evaluating multimodal agents on realistic visual web tasks.
\newblock \emph{arXiv preprint arXiv:2401.13649}, 2024{\natexlab{a}}.

\bibitem[Koh et~al.(2024{\natexlab{b}})Koh, McAleer, Fried, and Salakhutdinov]{koh2024tree}
Jing~Yu Koh, Stephen McAleer, Daniel Fried, and Ruslan Salakhutdinov.
\newblock Tree search for language model agents.
\newblock \emph{arXiv preprint arXiv:2407.01476}, 2024{\natexlab{b}}.

\bibitem[Kojima et~al.(2022)Kojima, Gu, Reid, Matsuo, and Iwasawa]{kojima2022large}
Takeshi Kojima, Shixiang~Shane Gu, Machel Reid, Yutaka Matsuo, and Yusuke Iwasawa.
\newblock Large language models are zero-shot reasoners.
\newblock \emph{Advances in neural information processing systems}, 35:\penalty0 22199--22213, 2022.

\bibitem[Kolve et~al.(2017)Kolve, Mottaghi, Han, VanderBilt, Weihs, Herrasti, Gordon, Zhu, Gupta, and Farhadi]{ai2thor}
Eric Kolve, Roozbeh Mottaghi, Winson Han, Eli VanderBilt, Luca Weihs, Alvaro Herrasti, Daniel Gordon, Yuke Zhu, Abhinav Gupta, and Ali Farhadi.
\newblock {AI2-THOR: An Interactive 3D Environment for Visual AI}.
\newblock \emph{arXiv}, 2017.

\bibitem[Krantz et~al.(2023)Krantz, Gervet, Yadav, Wang, Paxton, Mottaghi, Batra, Malik, Lee, and Chaplot]{krantz2023navigating}
Jacob Krantz, Theophile Gervet, Karmesh Yadav, Austin Wang, Chris Paxton, Roozbeh Mottaghi, Dhruv Batra, Jitendra Malik, Stefan Lee, and Devendra~Singh Chaplot.
\newblock Navigating to objects specified by images.
\newblock \emph{arXiv preprint arXiv:2304.01192}, 2023.

\bibitem[Ku et~al.(2020)Ku, Anderson, Patel, Ie, and Baldridge]{ku2020room}
Alexander Ku, Peter Anderson, Roma Patel, Eugene Ie, and Jason Baldridge.
\newblock Room-across-room: Multilingual vision-and-language navigation with dense spatiotemporal grounding.
\newblock In \emph{Proceedings of the 2020 Conference on Empirical Methods in Natural Language Processing (EMNLP)}, pages 4392--4412, 2020.

\bibitem[Laurençon et~al.(2023)Laurençon, Saulnier, Tronchon, Bekman, Singh, Lozhkov, Wang, Karamcheti, Rush, Kiela, Cord, and Sanh]{laurencon2023obelics}
Hugo Laurençon, Lucile Saulnier, Léo Tronchon, Stas Bekman, Amanpreet Singh, Anton Lozhkov, Thomas Wang, Siddharth Karamcheti, Alexander~M. Rush, Douwe Kiela, Matthieu Cord, and Victor Sanh.
\newblock Obelics: An open web-scale filtered dataset of interleaved image-text documents, 2023.

\bibitem[Li et~al.(2023)Li, Prabhudesai, Duggal, Brown, and Pathak]{li2023diffusion}
Alexander~C. Li, Mihir Prabhudesai, Shivam Duggal, Ellis Brown, and Deepak Pathak.
\newblock Your diffusion model is secretly a zero-shot classifier.
\newblock In \emph{Proceedings of the IEEE/CVF International Conference on Computer Vision (ICCV)}, pages 2206--2217, October 2023.

\bibitem[Liang et~al.(2022)Liang, Huang, Xia, Xu, Hausman, Ichter, Florence, and Zeng]{liang2022code}
Jacky Liang, Wenlong Huang, Fei Xia, Peng Xu, Karol Hausman, Brian Ichter, Pete Florence, and Andy Zeng.
\newblock Code as policies: Language model programs for embodied control.
\newblock \emph{arXiv preprint arXiv:2209.07753}, 2022.

\bibitem[Liang et~al.(2024)Liang, Xia, Yu, Zeng, Arenas, Attarian, Bauza, Bennice, Bewley, Dostmohamed, et~al.]{liang2024learning}
Jacky Liang, Fei Xia, Wenhao Yu, Andy Zeng, Montserrat~Gonzalez Arenas, Maria Attarian, Maria Bauza, Matthew Bennice, Alex Bewley, Adil Dostmohamed, et~al.
\newblock Learning to learn faster from human feedback with language model predictive control.
\newblock \emph{arXiv preprint arXiv:2402.11450}, 2024.

\bibitem[Lin et~al.(2014)Lin, Maire, Belongie, Hays, Perona, Ramanan, Doll{\'a}r, and Zitnick]{lin2014microsoft}
Tsung-Yi Lin, Michael Maire, Serge Belongie, James Hays, Pietro Perona, Deva Ramanan, Piotr Doll{\'a}r, and C~Lawrence Zitnick.
\newblock Microsoft coco: Common objects in context.
\newblock In \emph{Computer Vision--ECCV 2014: 13th European Conference, Zurich, Switzerland, September 6-12, 2014, Proceedings, Part V 13}, pages 740--755. Springer, 2014.

\bibitem[Liu et~al.(2018)Liu, Guu, Pasupat, Shi, and Liang]{liu2018reinforcement}
Evan~Zheran Liu, Kelvin Guu, Panupong Pasupat, Tianlin Shi, and Percy Liang.
\newblock Reinforcement learning on web interfaces using workflow-guided exploration.
\newblock In \emph{International Conference on Learning Representations ({ICLR})}, 2018.
\newblock URL \url{https://arxiv.org/abs/1802.08802}.

\bibitem[Liu et~al.(2023{\natexlab{a}})Liu, Li, Li, and Lee]{liu2023improvedllava}
Haotian Liu, Chunyuan Li, Yuheng Li, and Yong~Jae Lee.
\newblock Improved baselines with visual instruction tuning, 2023{\natexlab{a}}.

\bibitem[Liu et~al.(2023{\natexlab{b}})Liu, Chen, Zhu, Swaminathan, Kolobov, and Cheng]{liu2023interactive}
Huihan Liu, Alice Chen, Yuke Zhu, Adith Swaminathan, Andrey Kolobov, and Ching-An Cheng.
\newblock Interactive robot learning from verbal correction.
\newblock \emph{arXiv preprint arXiv:2310.17555}, 2023{\natexlab{b}}.

\bibitem[Liu et~al.(2023{\natexlab{c}})Liu, Dass, Mart{\'\i}n-Mart{\'\i}n, and Zhu]{liu2023model}
Huihan Liu, Shivin Dass, Roberto Mart{\'\i}n-Mart{\'\i}n, and Yuke Zhu.
\newblock Model-based runtime monitoring with interactive imitation learning.
\newblock \emph{arXiv preprint arXiv:2310.17552}, 2023{\natexlab{c}}.

\bibitem[Liu et~al.(2023{\natexlab{d}})Liu, Yu, Zhang, Xu, Lei, Lai, Gu, Ding, Men, Yang, Zhang, Deng, Zeng, Du, Zhang, Shen, Zhang, Su, Sun, Huang, Dong, and Tang]{liu2023agentbench}
Xiao Liu, Hao Yu, Hanchen Zhang, Yifan Xu, Xuanyu Lei, Hanyu Lai, Yu~Gu, Hangliang Ding, Kaiwen Men, Kejuan Yang, Shudan Zhang, Xiang Deng, Aohan Zeng, Zhengxiao Du, Chenhui Zhang, Sheng Shen, Tianjun Zhang, Yu~Su, Huan Sun, Minlie Huang, Yuxiao Dong, and Jie Tang.
\newblock Agentbench: Evaluating llms as agents, 2023{\natexlab{d}}.

\bibitem[Liu et~al.(2023{\natexlab{e}})Liu, Bahety, and Song]{liu2023reflect}
Zeyi Liu, Arpit Bahety, and Shuran Song.
\newblock Reflect: Summarizing robot experiences for failure explanation and correction.
\newblock \emph{arXiv preprint arXiv:2306.15724}, 2023{\natexlab{e}}.

\bibitem[Lù et~al.(2024)Lù, Kasner, and Reddy]{lù2024weblinx}
Xing~Han Lù, Zdeněk Kasner, and Siva Reddy.
\newblock Weblinx: Real-world website navigation with multi-turn dialogue, 2024.

\bibitem[Ma et~al.(2023)Ma, Liang, Wang, Huang, Bastani, Jayaraman, Zhu, Fan, and Anandkumar]{ma2023eureka}
Yecheng~Jason Ma, William Liang, Guanzhi Wang, De-An Huang, Osbert Bastani, Dinesh Jayaraman, Yuke Zhu, Linxi Fan, and Anima Anandkumar.
\newblock Eureka: Human-level reward design via coding large language models.
\newblock \emph{arXiv preprint arXiv:2310.12931}, 2023.

\bibitem[Majumdar et~al.(2024)Majumdar, Ajay, Zhang, Putta, Yenamandra, Henaff, Silwal, Mcvay, Maksymets, Arnaud, Yadav, Li, Newman, Sharma, Berges, Zhang, Agrawal, Bisk, Batra, Kalakrishnan, Meier, Paxton, Sax, and Rajeswaran]{OpenEQA2023}
Arjun Majumdar, Anurag Ajay, Xiaohan Zhang, Pranav Putta, Sriram Yenamandra, Mikael Henaff, Sneha Silwal, Paul Mcvay, Oleksandr Maksymets, Sergio Arnaud, Karmesh Yadav, Qiyang Li, Ben Newman, Mohit Sharma, Vincent Berges, Shiqi Zhang, Pulkit Agrawal, Yonatan Bisk, Dhruv Batra, Mrinal Kalakrishnan, Franziska Meier, Chris Paxton, Sasha Sax, and Aravind Rajeswaran.
\newblock Openeqa: Embodied question answering in the era of foundation models.
\newblock In \emph{Conference on Computer Vision and Pattern Recognition (CVPR)}, 2024.

\bibitem[Majumder et~al.(2023)Majumder, Dalvi~Mishra, Jansen, Tafjord, Tandon, Zhang, Callison-Burch, and Clark]{majumder2023clin}
Bodhisattwa~Prasad Majumder, Bhavana Dalvi~Mishra, Peter Jansen, Oyvind Tafjord, Niket Tandon, Li~Zhang, Burch Callison-Burch, and Peter Clark.
\newblock Clin: A continually learning language agent for rapid task adaptation and generalization.
\newblock \emph{arXiv}, 2023.

\bibitem[Min et~al.(2021)Min, Chaplot, Ravikumar, Bisk, and Salakhutdinov]{min2021film}
So~Yeon Min, Devendra~Singh Chaplot, Pradeep Ravikumar, Yonatan Bisk, and Ruslan Salakhutdinov.
\newblock Film: Following instructions in language with modular methods, 2021.

\bibitem[Min et~al.(2022)Min, Zhu, Salakhutdinov, and Bisk]{min2022don}
So~Yeon Min, Hao Zhu, Ruslan Salakhutdinov, and Yonatan Bisk.
\newblock Don’t copy the teacher: Data and model challenges in embodied dialogue.
\newblock In \emph{Proceedings of the 2022 Conference on Empirical Methods in Natural Language Processing}, pages 9361--9368, 2022.

\bibitem[Mirchandani et~al.(2021)Mirchandani, Karamcheti, and Sadigh]{mirchandani2021ella}
Suvir Mirchandani, Siddharth Karamcheti, and Dorsa Sadigh.
\newblock Ella: Exploration through learned language abstraction.
\newblock \emph{Advances in neural information processing systems}, 34:\penalty0 29529--29540, 2021.

\bibitem[Mu et~al.()Mu, Zhong, Raileanu, Jiang, Goodman, Rockt{\"a}schel, and Grefenstette]{mu2202improving}
Jesse Mu, Victor Zhong, Roberta Raileanu, Minqi Jiang, Noah Goodman, Tim Rockt{\"a}schel, and Edward Grefenstette.
\newblock Improving intrinsic exploration with language abstractions (2022).
\newblock \emph{URL https://arxiv. org/abs/2202.08938}.

\bibitem[Mu et~al.(2022)Mu, Zhong, Raileanu, Jiang, Goodman, Rockt{\"a}schel, and Grefenstette]{mu2022improving}
Jesse Mu, Victor Zhong, Roberta Raileanu, Minqi Jiang, Noah Goodman, Tim Rockt{\"a}schel, and Edward Grefenstette.
\newblock Improving intrinsic exploration with language abstractions.
\newblock \emph{Advances in Neural Information Processing Systems}, 35:\penalty0 33947--33960, 2022.

\bibitem[Nottingham et~al.(2024)Nottingham, Majumder, Dalvi~Mishra, Singh, Clark, and Fox]{nottingham2024sso}
Kolby Nottingham, Bodhisattwa~Prasad Majumder, Bhavana Dalvi~Mishra, Sameer Singh, Peter Clark, and Roy Fox.
\newblock Skill set optimization: Reinforcing language model behavior via transferable skills.
\newblock \emph{arXiv}, 2024.
\newblock URL \url{https://arxiv.org/abs/2402.03244}.

\bibitem[Padmakumar et~al.(2021)Padmakumar, Thomason, Shrivastava, Lange, Narayan-Chen, Gella, Piramuthu, Tur, and Hakkani-Tur]{TEACH}
Aishwarya Padmakumar, Jesse Thomason, Ayush Shrivastava, Patrick Lange, Anjali Narayan-Chen, Spandana Gella, Robinson Piramuthu, Gokhan Tur, and Dilek Hakkani-Tur.
\newblock Teach: Task-driven embodied agents that chat, 2021.

\bibitem[Pashevich et~al.(2021)Pashevich, Schmid, and Sun]{pashevich2021episodic}
Alexander Pashevich, Cordelia Schmid, and Chen Sun.
\newblock Episodic transformer for vision-and-language navigation.
\newblock In \emph{Proceedings of the IEEE/CVF International Conference on Computer Vision}, pages 15942--15952, 2021.

\bibitem[Putta et~al.(2024)Putta, Mills, Garg, Motwani, Finn, Garg, and Rafailov]{putta2024agent}
Pranav Putta, Edmund Mills, Naman Garg, Sumeet Motwani, Chelsea Finn, Divyansh Garg, and Rafael Rafailov.
\newblock Agent q: Advanced reasoning and learning for autonomous ai agents.
\newblock \emph{arXiv preprint arXiv:2408.07199}, 2024.

\bibitem[Radford et~al.(2021)Radford, Kim, Hallacy, Ramesh, Goh, Agarwal, Sastry, Askell, Mishkin, Clark, Krueger, and Sutskever]{Radford2021LearningTV}
Alec Radford, Jong~Wook Kim, Chris Hallacy, A.~Ramesh, Gabriel Goh, Sandhini Agarwal, Girish Sastry, Amanda Askell, Pamela Mishkin, Jack Clark, Gretchen Krueger, and Ilya Sutskever.
\newblock Learning transferable visual models from natural language supervision.
\newblock In \emph{ICML}, 2021.

\bibitem[Ren et~al.(2023)Ren, Dixit, Bodrova, Singh, Tu, Brown, Xu, Takayama, Xia, Varley, et~al.]{ren2023robots}
Allen~Z Ren, Anushri Dixit, Alexandra Bodrova, Sumeet Singh, Stephen Tu, Noah Brown, Peng Xu, Leila Takayama, Fei Xia, Jake Varley, et~al.
\newblock Robots that ask for help: Uncertainty alignment for large language model planners.
\newblock In \emph{Conference on Robot Learning}, pages 661--682. PMLR, 2023.

\bibitem[Ross et~al.(2011)Ross, Gordon, and Bagnell]{ross2011reduction}
St{\'e}phane Ross, Geoffrey Gordon, and Drew Bagnell.
\newblock A reduction of imitation learning and structured prediction to no-regret online learning.
\newblock In \emph{Proceedings of the fourteenth international conference on artificial intelligence and statistics}, pages 627--635. JMLR Workshop and Conference Proceedings, 2011.

\bibitem[Sarch et~al.(2023)Sarch, Wu, Tarr, and Fragkiadaki]{sarch2023helper}
Gabriel Sarch, Yue Wu, Michael Tarr, and Katerina Fragkiadaki.
\newblock Open-ended instructable embodied agents with memory-augmented large language models.
\newblock In \emph{Findings of the Association for Computational Linguistics: EMNLP 2023}, 2023.

\bibitem[Shi et~al.(2017)Shi, Karpathy, Fan, Hernandez, and Liang]{shi2017world}
Tianlin Shi, Andrej Karpathy, Linxi Fan, Jonathan Hernandez, and Percy Liang.
\newblock World of bits: An open-domain platform for web-based agents.
\newblock In \emph{ICML}, 2017.

\bibitem[Shinn et~al.(2023)Shinn, Labash, and Gopinath]{shinn2023reflexion}
Noah Shinn, Beck Labash, and Ashwin Gopinath.
\newblock Reflexion: an autonomous agent with dynamic memory and self-reflection.
\newblock \emph{arXiv preprint arXiv:2303.11366}, 2023.

\bibitem[Shridhar et~al.(2020)Shridhar, Thomason, Gordon, Bisk, Han, Mottaghi, Zettlemoyer, and Fox]{shridhar2020alfred}
Mohit Shridhar, Jesse Thomason, Daniel Gordon, Yonatan Bisk, Winson Han, Roozbeh Mottaghi, Luke Zettlemoyer, and Dieter Fox.
\newblock {ALFRED}: A benchmark for interpreting grounded instructions for everyday tasks.
\newblock In \emph{Proceedings of the IEEE/CVF conference on computer vision and pattern recognition}, pages 10740--10749, 2020.

\bibitem[Singh et~al.(2023)Singh, Blukis, Mousavian, Goyal, Xu, Tremblay, Fox, Thomason, and Garg]{singh2022progprompt}
Ishika Singh, Valts Blukis, Arsalan Mousavian, Ankit Goyal, Danfei Xu, Jonathan Tremblay, Dieter Fox, Jesse Thomason, and Animesh Garg.
\newblock Progprompt: Generating situated robot task plans using large language models.
\newblock In \emph{2023 IEEE International Conference on Robotics and Automation (ICRA)}, pages 11523--11530. IEEE, 2023.

\bibitem[Song et~al.(2023)Song, Wu, Washington, Sadler, Chao, and Su]{song2023llmplanner}
Chan~Hee Song, Jiaman Wu, Clayton Washington, Brian~M. Sadler, Wei-Lun Chao, and Yu~Su.
\newblock Llm-planner: Few-shot grounded planning for embodied agents with large language models, 2023.

\bibitem[Tasrin et~al.(2021)Tasrin, Nahian, Perera, and Harrison]{tasrin2021influencing}
Tasmia Tasrin, Md~Sultan~Al Nahian, Habarakadage Perera, and Brent Harrison.
\newblock Influencing reinforcement learning through natural language guidance.
\newblock \emph{arXiv preprint arXiv:2104.01506}, 2021.

\bibitem[Tenenbaum et~al.(2011)Tenenbaum, Kemp, Griffiths, and Goodman]{tenenbaum2011grow}
Joshua~B Tenenbaum, Charles Kemp, Thomas~L Griffiths, and Noah~D Goodman.
\newblock How to grow a mind: Statistics, structure, and abstraction.
\newblock \emph{science}, 331\penalty0 (6022):\penalty0 1279--1285, 2011.

\bibitem[Wang et~al.(2023{\natexlab{a}})Wang, Xie, Jiang, Mandlekar, Xiao, Zhu, Fan, and Anandkumar]{wang2023voyager}
Guanzhi Wang, Yuqi Xie, Yunfan Jiang, Ajay Mandlekar, Chaowei Xiao, Yuke Zhu, Linxi Fan, and Anima Anandkumar.
\newblock Voyager: An open-ended embodied agent with large language models.
\newblock \emph{arXiv preprint arXiv: Arxiv-2305.16291}, 2023{\natexlab{a}}.

\bibitem[Wang et~al.(2024)Wang, Bai, Tan, Wang, Fan, Bai, Chen, Liu, Wang, Ge, Fan, Dang, Du, Ren, Men, Liu, Zhou, Zhou, and Lin]{Qwen2VL}
Peng Wang, Shuai Bai, Sinan Tan, Shijie Wang, Zhihao Fan, Jinze Bai, Keqin Chen, Xuejing Liu, Jialin Wang, Wenbin Ge, Yang Fan, Kai Dang, Mengfei Du, Xuancheng Ren, Rui Men, Dayiheng Liu, Chang Zhou, Jingren Zhou, and Junyang Lin.
\newblock Qwen2-vl: Enhancing vision-language model's perception of the world at any resolution.
\newblock \emph{arXiv preprint arXiv:2409.12191}, 2024.

\bibitem[Wang et~al.(2023{\natexlab{b}})Wang, Lv, Yu, Hong, Qi, Wang, Ji, Yang, Zhao, Song, Xu, Xu, Li, Dong, Ding, and Tang]{wang2023cogvlm}
Weihan Wang, Qingsong Lv, Wenmeng Yu, Wenyi Hong, Ji~Qi, Yan Wang, Junhui Ji, Zhuoyi Yang, Lei Zhao, Xixuan Song, Jiazheng Xu, Bin Xu, Juanzi Li, Yuxiao Dong, Ming Ding, and Jie Tang.
\newblock Cogvlm: Visual expert for pretrained language models, 2023{\natexlab{b}}.

\bibitem[Wang et~al.(2022)Wang, Wei, Schuurmans, Le, Chi, Narang, Chowdhery, and Zhou]{wang2022self}
Xuezhi Wang, Jason Wei, Dale Schuurmans, Quoc Le, Ed~Chi, Sharan Narang, Aakanksha Chowdhery, and Denny Zhou.
\newblock Self-consistency improves chain of thought reasoning in language models.
\newblock \emph{arXiv preprint arXiv:2203.11171}, 2022.

\bibitem[Wang et~al.(2023{\natexlab{c}})Wang, Cai, Liu, Jin, Hou, Zhang, Lin, He, Zheng, Yang, Ma, and Liang]{wang2023jarvis1}
Zihao Wang, Shaofei Cai, Anji Liu, Yonggang Jin, Jinbing Hou, Bowei Zhang, Haowei Lin, Zhaofeng He, Zilong Zheng, Yaodong Yang, Xiaojian Ma, and Yitao Liang.
\newblock Jarvis-1: Open-world multi-task agents with memory-augmented multimodal language models.
\newblock \emph{arXiv preprint arXiv: 2311.05997}, 2023{\natexlab{c}}.

\bibitem[Wang et~al.(2023{\natexlab{d}})Wang, Cai, Liu, Ma, and Liang]{wang2023describe}
Zihao Wang, Shaofei Cai, Anji Liu, Xiaojian Ma, and Yitao Liang.
\newblock Describe, explain, plan and select: Interactive planning with large language models enables open-world multi-task agents.
\newblock \emph{arXiv preprint arXiv:2302.01560}, 2023{\natexlab{d}}.

\bibitem[Wei et~al.(2022)Wei, Wang, Schuurmans, Bosma, Chi, Le, and Zhou]{wei2022chain}
Jason Wei, Xuezhi Wang, Dale Schuurmans, Maarten Bosma, Ed~Chi, Quoc Le, and Denny Zhou.
\newblock Chain of thought prompting elicits reasoning in large language models.
\newblock \emph{arXiv preprint arXiv:2201.11903}, 2022.

\bibitem[Wu and Xie(2023)]{wu2023textit}
Penghao Wu and Saining Xie.
\newblock V*: Guided visual search as a core mechanism in multimodal llms.
\newblock \emph{arXiv preprint arXiv:2312.14135}, 2023.

\bibitem[Wu et~al.(2023)Wu, Fan, Liang, Azaria, Li, and Mitchell]{wu2023read}
Yue Wu, Yewen Fan, Paul~Pu Liang, Amos Azaria, Yuanzhi Li, and Tom~M Mitchell.
\newblock Read and reap the rewards: Learning to play atari with the help of instruction manuals.
\newblock In \emph{NeurIPS}, 2023.

\bibitem[Yang et~al.(2023)Yang, Zhang, Li, Zou, Li, and Gao]{yang2023setofmark}
Jianwei Yang, Hao Zhang, Feng Li, Xueyan Zou, Chunyuan Li, and Jianfeng Gao.
\newblock Set-of-mark prompting unleashes extraordinary visual grounding in gpt-4v, 2023.
\newblock URL \url{https://arxiv.org/abs/2310.11441}.

\bibitem[Yao et~al.(2023)Yao, Chen, Yang, and Narasimhan]{yao2023webshop}
Shunyu Yao, Howard Chen, John Yang, and Karthik Narasimhan.
\newblock Webshop: Towards scalable real-world web interaction with grounded language agents, 2023.

\bibitem[Yenamandra et~al.(2023)Yenamandra, Ramachandran, Yadav, Wang, Khanna, Gervet, Yang, Jain, Clegg, Turner, et~al.]{yenamandra2023homerobot}
Sriram Yenamandra, Arun Ramachandran, Karmesh Yadav, Austin Wang, Mukul Khanna, Theophile Gervet, Tsung-Yen Yang, Vidhi Jain, Alexander~William Clegg, John Turner, et~al.
\newblock Homerobot: Open-vocabulary mobile manipulation.
\newblock \emph{arXiv preprint arXiv:2306.11565}, 2023.

\bibitem[Zeng et~al.(2021)Zeng, Zhang, and Li]{xvlm}
Yan Zeng, Xinsong Zhang, and Hang Li.
\newblock Multi-grained vision language pre-training: Aligning texts with visual concepts.
\newblock \emph{arXiv preprint arXiv:2111.08276}, 2021.

\bibitem[Zha et~al.(2023)Zha, Cui, Lin, Kwon, Arenas, Zeng, Xia, and Sadigh]{zha2023distilling}
Lihan Zha, Yuchen Cui, Li-Heng Lin, Minae Kwon, Montserrat~Gonzalez Arenas, Andy Zeng, Fei Xia, and Dorsa Sadigh.
\newblock Distilling and retrieving generalizable knowledge for robot manipulation via language corrections.
\newblock In \emph{2nd Workshop on Language and Robot Learning: Language as Grounding}, 2023.

\bibitem[Zhang et~al.(2022)Zhang, Yang, Pan, Storks, Devraj, Ma, Yu, Bao, and Chai]{zhang2022danli}
Yichi Zhang, Jianing Yang, Jiayi Pan, Shane Storks, Nikhil Devraj, Ziqiao Ma, Keunwoo Yu, Yuwei Bao, and Joyce Chai.
\newblock Danli: Deliberative agent for following natural language instructions.
\newblock In \emph{Proceedings of the 2022 Conference on Empirical Methods in Natural Language Processing}, pages 1280--1298, 2022.

\bibitem[Zheng et~al.(2024)Zheng, Gou, Kil, Sun, and Su]{zheng2024gpt}
Boyuan Zheng, Boyu Gou, Jihyung Kil, Huan Sun, and Yu~Su.
\newblock Gpt-4v (ision) is a generalist web agent, if grounded.
\newblock \emph{arXiv preprint arXiv:2401.01614}, 2024.

\bibitem[Zheng et~al.(2022)Zheng, Zhou, Gu, Fan, Wang, Li, He, and Wang]{zheng2022jarvis}
Kaizhi Zheng, Kaiwen Zhou, Jing Gu, Yue Fan, Jialu Wang, Zonglin Li, Xuehai He, and Xin~Eric Wang.
\newblock Jarvis: A neuro-symbolic commonsense reasoning framework for conversational embodied agents.
\newblock 2022.

\bibitem[Zhou et~al.(2023)Zhou, Xu, Zhu, Zhou, Lo, Sridhar, Cheng, Ou, Bisk, Fried, Alon, and Neubig]{zhou2023webarena}
Shuyan Zhou, Frank~F. Xu, Hao Zhu, Xuhui Zhou, Robert Lo, Abishek Sridhar, Xianyi Cheng, Tianyue Ou, Yonatan Bisk, Daniel Fried, Uri Alon, and Graham Neubig.
\newblock Webarena: A realistic web environment for building autonomous agents, 2023.

\bibitem[Zhu et~al.(2023)Zhu, Kapoor, Min, Han, Li, Geng, Neubig, Bisk, Kembhavi, and Weihs]{zhu2023excalibur}
Hao Zhu, Raghav Kapoor, So~Yeon Min, Winson Han, Jiatai Li, Kaiwen Geng, Graham Neubig, Yonatan Bisk, Aniruddha Kembhavi, and Luca Weihs.
\newblock Excalibur: Encouraging and evaluating embodied exploration.
\newblock In \emph{Proceedings of the IEEE/CVF Conference on Computer Vision and Pattern Recognition}, pages 14931--14942, 2023.

\end{thebibliography}

\clearpage
\makeatletter
% \clearpage
% \maketitlesupplementary
\renewcommand \thesection{S\@arabic\c@section}
\renewcommand\thetable{S\@arabic\c@table}
\renewcommand \thefigure{S\@arabic\c@figure}
\renewcommand \thelstlisting{S\@arabic\c@lstlisting}
\renewcommand \thealgorithm{S\@arabic\c@algorithm}
\makeatother

\setcounter{section}{0}
\setcounter{figure}{0}  
\setcounter{table}{0} 

\renewcommand{\theHsection}{Supplement.\thesection}
\renewcommand{\theHtable}{Supplement.\thetable}
\renewcommand{\theHfigure}{Supplement.\thefigure}

\section{Overview}
The structure of this Appendix is as follows: 
\begin{itemize}
\item Section~\ref{sec:neg_impact} contains negative potential impacts.
\item Section~\ref{sec:supp_exps} contains additional experiments.
\item Section~\ref{sec:supp_meths} contains additional methods details. 
\item Section~\ref{sec:supp_envs} contains additional details on the evaluation environments.
\end{itemize}

\section{Potential Negative Impact} \label{sec:neg_impact}
The work introduced by ICAL for AI agents carries potential risks including the perpetuation of biases, privacy infringement, user dependency, economic displacement, security vulnerabilities, and the emergence of unintended behaviors due to technical limitations. Mitigating these risks necessitates the development of mechanisms for bias correction, privacy preservation, ethical guidelines, and security protocols. Engaging with a diverse range of stakeholders is imperative to ensure that the deployment of these technologies aligns with societal values and contributes positively to the realm of human-AI collaboration, fostering advancements that are both innovative and responsible.

\section{ICAL relation to dAgger}
\subsubsection{Relation of Abstraction Verification to DAgger}\label{sec:dagger_appendix}

The human-in-the-loop phase of ICAL bears a conceptual resemblance to the Dataset Aggregation (DAgger) algorithm~\citep{ross2011reduction}, as both methods involve iterative refinement of an agent's policy through interaction with expert feedback. However, ICAL extends this framework by incorporating natural language feedback, updating both actions and abstractions, and utilizing retrieval-augmented generation (RAG) with an explicit memory of optimized examples for policy improvement.

In DAgger, the agent collects data by executing its current policy and then queries an expert to obtain the correct action for each encountered state. Specifically, at iteration \( t \), the agent observes a state \( s_t \) and takes an action \( a_t = \pi_t(s_t) \) according to its policy \( \pi_t \). The expert provides the optimal action \( a_t^* \), and the agent aggregates this data into a dataset \( \mathcal{D} \):
\begin{equation}
\mathcal{D} = \mathcal{D} \cup \{(s_t, a_t^*)\}.
\end{equation}
The policy is then updated by minimizing a loss function over \( \mathcal{D} \):
\begin{equation}
\pi_{t+1} = \arg\min_{\pi} \sum_{(s_i, a_i^*) \in \mathcal{D}} L(\pi(s_i), a_i^*).
\end{equation}

Similarly, in ICAL's human-in-the-loop phase \( F_{\text{hitl}} \), the agent refines its behavior based on human feedback. When the agent executes an optimized trajectory \( \xi_{\text{optimized}} \) and encounters a failure at action \( a_t \), a human observer provides natural language feedback \( H(a_t, o_t) \) concerning the action \( a_t \) and the observation \( o_t \). The agent integrates this feedback to update both the trajectory and the associated language abstractions:
\begin{equation}
(\xi'_{\text{optimized}}, L') = \Xi_{\text{update}}(\xi_{\text{optimized}}, H(a_t, o_t), L, I, \{e^1, \ldots, e^k\}),
\end{equation}
where \( L \) represents the current language annotations, \( I \) is the task instruction, and \( \{e^1, \ldots, e^k\} \) are retrieved examples from memory. This updated trajectory \( \xi'_{\text{optimized}} \) and abstractions \( L' \) are then added to an explicit memory \( \mathcal{E} \), enhancing the agent's policy through enriched context:
\begin{equation}
\mathcal{E} = \mathcal{E} \cup \{(\xi'_{\text{optimized}}, L')\}.
\end{equation}
The agent's policy \( \pi_{\text{ICAL}} \) is implicitly updated by conditioning on this memory during action generation:
\begin{equation}
\pi_{\text{ICAL}}(s_t, \mathcal{E}) = \text{VLM}(s_t, \mathcal{E}),
\end{equation}
where \( \text{VLM} \) denotes the Vision-Language Model used for in-context learning.

The similarities between ICAL and DAgger lie in their iterative approach to policy refinement using expert feedback. However, ICAL offers several key benefits:

\textbf{Natural Language Feedback:} Unlike DAgger, which requires the expert to provide explicit action corrections \( a_t^* \), ICAL accepts natural language feedback \( H(a_t, o_t) \). This allows the human to convey richer information, including explanations, suggestions, and contextual details that can address not only the immediate failure but also underlying misconceptions.

\textbf{Revision of Actions and Abstractions:} ICAL updates both the action sequence and the associated language abstractions \( L \). By refining the abstractions, the agent enhances its understanding of task structures, causal relationships, and state changes, which promotes better generalization to new tasks and environments.

\textbf{Policy Improvement via Retrieval-Augmented Generation:} ICAL maintains an explicit memory \( \mathcal{E} \) of optimized examples and abstractions. During deployment, the agent retrieves relevant examples from \( \mathcal{E} \) based on similarity measures and uses them as context for action generation. This retrieval-augmented generation (RAG) approach allows the agent to leverage past experiences effectively, adapting its policy without explicit parameter updates.

In contrast, DAgger relies solely on aggregating state-action pairs and updating the policy through supervised learning, which may not capture higher-level task structures or facilitate transfer to new domains. ICAL's ability to process natural language feedback and to update both actions and abstractions provides a more flexible and powerful framework for policy refinement, aligning more closely with human learning processes. ICAL extends the traditional imitation learning paradigm represented by DAgger, enabling more efficient and generalizable learning from human feedback.

\section{Additional Experiments} \label{sec:supp_exps}

\subsection{Learning efficiency broken down by task type}
\begin{figure}[ht!]
\centering
% \includegraphics[width=0.33\textwidth]{Figures/stats_continual_learning_overall_0_3.jpg}
% \hspace{0.02\textwidth}
\includegraphics[width=0.55\textwidth]{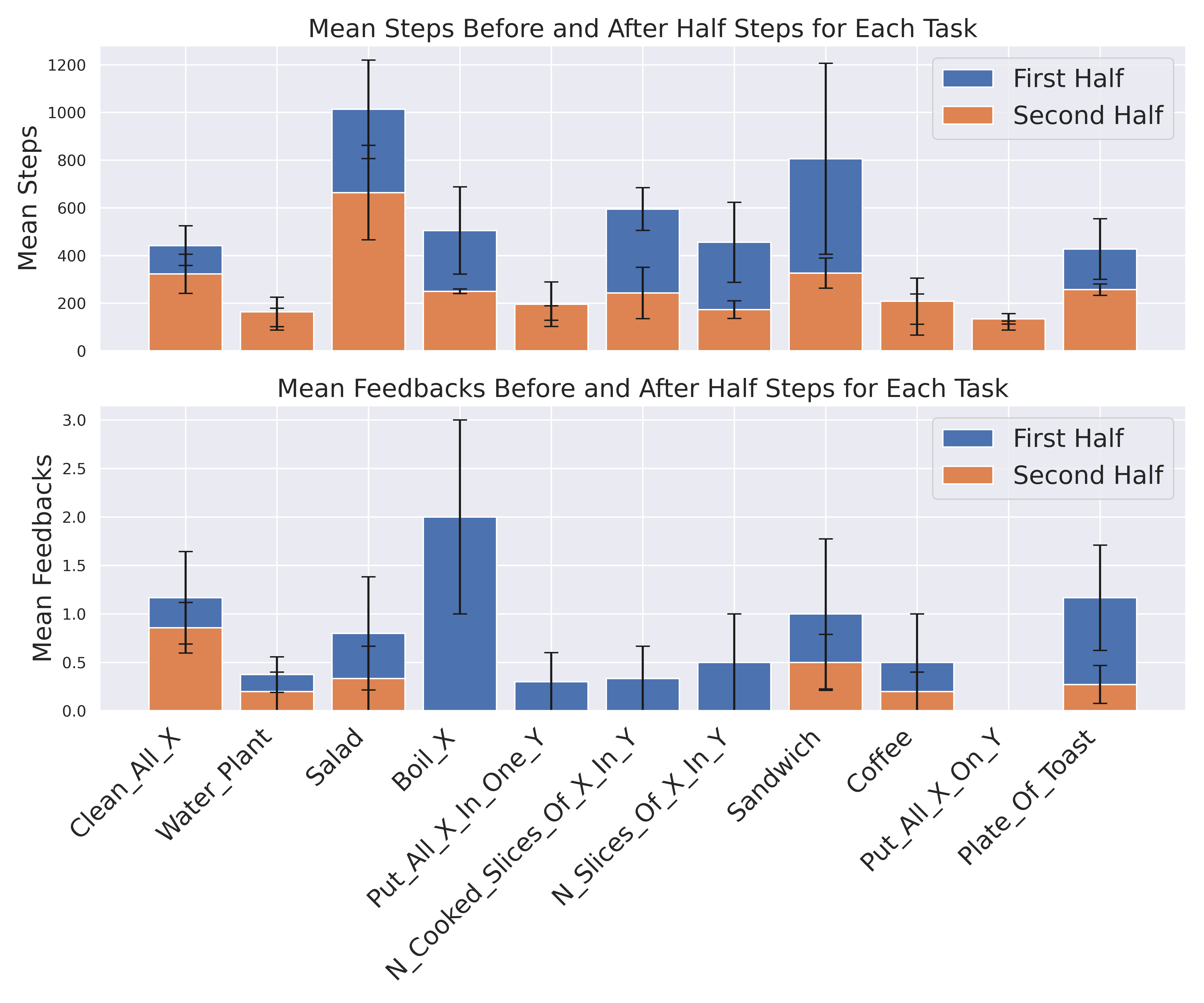}
\caption{\small First half (blue) versus second half (orange) of ICAL learning across tasks \textbf{(left)} and for each task type separately \textbf{(right)} in TEACh.}
\label{figR1_task}
\end{figure}

In Section~\ref{sec:efficiency} of the main paper, we showed how ICAL enables fewer environment interactions and human feedbacks per example. We provide the learning efficiency between the first and second half of demonstrations processed broken down by task type in Figure~\ref{figR1_task}.
% \begin{wrapfigure}{r}{0.45\textwidth}
% \end{wrapfigure}

\subsection{Experimenting with different types of in-context examples in VisualWebArena}
We experiment with an alternate way to provide the \model{} in-context examples to the VLM. Instead of retrieving a single time step, we give the full trajectory of observations, abstractions, and actions in textual format (no images provided). We run this on a reduced subset of 239 VisualWebArena tasks. The results are presented in Table~\ref{tab:vwa_context_type}. We find that providing the full trajectory increases seen success rate, but does not improve unseen success rate. In our final evaluation, we utilize the retrieval of a single time step with image input, since expanding the context length through the full trajectory adds to the cost without significantly improving the success rate on unseen tasks.

\subsection{TEACh results on \model{} learning using trajectories with ground truth action labels and GPT3.5} \label{app:num_examples}

We present tasks successfully completed for each task type in Table~\ref{tab:num_examples}, comparing \model{} that uses visual demonstrations and kinesthetic demonstrations.

We see that using GPT3.5 for \model{} significantly reduces the number of successful tasks by over half compared to using GPT4 (52 versus 122 tasks successfully completed). We show in Section~\ref{sec:relabel} of the main paper how relabeling unsuccessful tasks can help improve performance when using weaker models, such as GPT 3.5. 

\subsection{TEACh validation accuracy by task type}
We present \model{} agent performance after learning on the TEACh validation set for each task type in Table~\ref{tab:teach_task_type}.

\begin{table}[t!]
\begin{center}
\footnotesize
\setlength{\tabcolsep}{2pt} % Adjusting the column separation
\caption{VisualWebArena performance for \model{} performance when using a single time step with image input for in-context examples versus providing all time steps, but without image inputs. 
}\label{tab:vwa_context_type}
\begin{tabular}{@{}llccc@{}}
 \\
 & \multicolumn{1}{c}{Seen} & \multicolumn{1}{c}{Unseen} & \multicolumn{1}{c}{Average} \\
\addlinespace[0.15cm] % Adjust the value as needed
\hline
\addlinespace[0.15cm] % Adjust the value as needed
GPT4V+SoM~\citep{koh2024visualwebarena} & 11.5 & 12.9 & 12.7 \\
\model{} + text + full trajectory & \textbf{57.7} & \textbf{21.6} & \textbf{25.5} \\
\model{} + text + single time step & 28.0 & 17.3 & 19.0 \\
\model{} + text + single time step + image & 28.0 & \textbf{21.6} & 22.2 \\
\addlinespace[0.15cm] % Adjust the value as needed
\hline
\end{tabular}
\end{center}
% \vspace{-10pt}
\end{table}

\begin{table}[t!]
\begin{center}
% \footnotesize
\scriptsize
\setlength{\tabcolsep}{2pt} % Adjusting the column separation
\caption{
Tasks successfully completed after applying the \model{} method (out of 250).
We compare \model{} using either visual demonstrations or kinesthetic demonstrations. 
Kinesth. = Kinesthetic; demos with GT actions. Visual = action labeled from RGB frames with inverse dynamics model.}\label{tab:num_examples}
\begin{tabular}{@{}lccc@{}}
 \\
 & \multicolumn{2}{c}{GPT-4} & \multicolumn{1}{c}{GPT-3.5} \\
 \cmidrule(lr){2-3}\cmidrule(lr){4-4} & \multicolumn{1}{c}{Visual Demos} & \multicolumn{1}{c}{Kinesth. Demos} & \multicolumn{1}{c}{Visual Demos} \\
\addlinespace[0.15cm] % Adjust the value as needed
\hline
\addlinespace[0.15cm] % Adjust the value as needed
Breakfast & 0 & 7 & 0 \\
Boil\_X & 5 & 11& 3 \\
Water\_Plant & 13 & 16 & 9 \\
Salad & 11 & 8 & 2 \\
Sandwich & 9 & 8 & 0 \\ 
Put\_All\_X\_On\_Y & 12 & 15 & 6 \\
Plate\_Of\_Toast & 17 & 14 & 9 \\
N\_Slices\_Of\_X\_In\_Y & 11 & 13 & 6 \\
Clean\_All\_X & 13 & 16 & 3 \\
Put\_All\_X\_In\_One\_Y & 13 & 16 & 10 \\
Coffee & 9 & 15 & 1 \\
N\_Cooked\_Slices\_Of\_X\_In\_Y & 5 & 3 & 3 \\
\textbf{Total} & 122 & 142 & 52 \\
\end{tabular}
\end{center}
\end{table}

\begin{table}[t]
\begin{center}
% \footnotesize
\scriptsize
\setlength{\tabcolsep}{2pt} % Adjusting the column separation
\caption{TEACh validation performance of \model{} after examples have been learned for each task type when evaluated using GPT3.5 or GPT4.}\label{tab:teach_task_type}
\begin{tabular}{@{}lcccc@{}}
 \\
 & \multicolumn{2}{c}{GPT3.5} & \multicolumn{2}{c}{GPT4} \\
 \cmidrule(lr){2-3}\cmidrule(lr){4-5} & \multicolumn{1}{c}{Success (\%)} & \multicolumn{1}{c}{GC (\%)} & \multicolumn{1}{c}{Success (\%)} & \multicolumn{1}{c}{GC (\%)} \\
\addlinespace[0.15cm] % Adjust the value as needed
\hline
\addlinespace[0.15cm] % Adjust the value as needed
Breakfast & \textbf{2.3} & 37.5 & \textbf{2.3} & \textbf{56.6} \\
Boil\_X & \textbf{18.2} & \textbf{18.2} & 13.6 & 13.6 \\
Water\_Plant & \textbf{73.0} & \textbf{73.0} & 61.9 & 61.9 \\
Salad & 31.9 & 61.7 & \textbf{40.4} & \textbf{74.6} \\
Sandwich & 2.1 & 45.4 & \textbf{2.1} & \textbf{56.0} \\ 
Put\_All\_X\_On\_Y & 56.0 & 65.6 & \textbf{66.0} & \textbf{74.0} \\
Plate\_Of\_Toast & 0.0 & 50.0 & \textbf{11.7} & \textbf{60.2} \\
N\_Slices\_Of\_X\_In\_Y & 48.1 & 57.6 & \textbf{63.0} & \textbf{72.2} \\
Clean\_All\_X & 58.6 & 58.8 & \textbf{70.7} & \textbf{75.0} \\
Put\_All\_X\_In\_One\_Y & 62.5 & \textbf{72.0} & \textbf{66.7} & \textbf{72.0} \\
Coffee & 31.6 & 44.6 & \textbf{49.1} & \textbf{59.5} \\
N\_Cooked\_Slices\_Of\_X\_In\_Y & 16.3 & 44.7 & \textbf{30.6} & \textbf{70.0} \\
\textbf{Average} & 35.1 & 49.3 & \textbf{41.7} & \textbf{63.6} \\
\end{tabular}
\end{center}
\end{table}

\subsection{Relabeling unsuccessful examples improves \model{} when using weaker models} \label{sec:relabel}
% \sectionvspace
Instead of removing unsuccessful examples, we can instead relabel the examples by querying an LLM to generate a new task instruction, step-by-step plan, and summary for the partial task completion. In Table~\ref{tab:relabel}, we show performance of \model{} with and without relabeling. Relabeling improves performance when using a weaker model, GPT3.5 during \model{} learning, by an absolute 3.4\% in success and 0.5\% in GC.

\begin{table}[h!]
\footnotesize
\setlength{\tabcolsep}{2pt} % Adjusting the column separation
\caption{Relabeling experiments. Relabeling unsuccessful demonstrations improves performance when using weaker models, such as GPT3.5, during the \model{} learning process.}\label{tab:relabel}
\begin{center}
\begin{tabular}{@{}llcccc@{}}
 \\
 & \multicolumn{1}{c}{Success} & \multicolumn{1}{c}{GC} \\
\addlinespace[0.15cm] % Adjust the value as needed
\hline
\addlinespace[0.15cm] % Adjust the value as needed
\model{} & 22.4 & 36.9 \\
\hspace{0.5mm} + relabeling & \textbf{25.8} & \textbf{37.4} \\
\addlinespace[0.15cm] % Adjust the value as needed
\hline
\end{tabular}
\end{center}
\end{table}

\subsection{Running \model{} from RGB-only input} \label{sec:estimated}
We run \model{} from RGB inputs. We use the perception, navigation, and manipulation modules from HELPER~\citep{sarch2023helper}, which uses SOLQ~\citep{dong2021solq} for object detection and ZeoDepth~\citep{bhat2023zoedepth} for depth estimation. HELPER initializes objects with default attributes based on the domain, and uses domain-specific pre-condition checks and error correction. However, we wish to automate the learning of these modules, and thus remove them. Additionally, HELPER initializes a memory of examples hand-written by a domain expert. We replace these with our \model{} examples. For inferring attributes of the objects in the abstracted state, we apply CogVLM~\citep{wang2023cogvlm}, an open-source visual language model, on the detected object crops, which we found to work best compared to other models for object attribute detection on a separate dataset of cropped object images (see Appendix). 

As shown in Table~\ref{tab:teach_tfd_estimated}, we find that our \model{} agent obtains performance close to that of HELPER, lagging behind 1.7\% success and 3.2\% goal condition success, despite not hand-designing object attributes, pre-condition checks, error correction, and in-context examples. Additionally, when using the hand-written HELPER examples with the \model{} execution modules, we find that the \model{} examples outperform the HELPER examples by 2.2\% in task success and by 1.3\% in goal-condition success, despite the \model{} examples being obtained without hand-writing from a domain-expert. Additionally, when using the perception of ODIN~\citep{jain2024odin}, which utilizes multi-view images and a 3D bottleneck for semantic segmentation, \model{} obtains performance on-par with that of HELPER.

\begin{table}[h!]
\caption{TEACh validation set (seen) from RGB input. Our \model{} agent obtains performance close to that of HELPER, despite not hand-designing object attributes, pre-condition checks, error correction, and in-context examples.
}\label{tab:teach_tfd_estimated}
% \vspace{-0.4cm}
\begin{center}
\begin{tabular}{@{}lcc@{}}
& \multicolumn{1}{c}{Success} & \multicolumn{1}{c}{GC} \\ 
\midrule
E.T.~\citep{pashevich2021episodic} & 1.0 & 1.4 \\
JARVIS~\citep{zheng2022jarvis} & 1.7 & 5.4 \\
FILM~\citep{min2021film} & 5.5 & 5.8 \\
DANLI~\citep{zhang2022danli} & 5.0 & 10.5 \\
HELPER~\citep{sarch2023helper} & 12.2 & 18.6 \\
\model{} w/ HELPER examples & 8.3 & 14.1 \\
\model{} (ours) & 10.5 & 15.4 \\
\addlinespace[0.15cm] % Adjust the value as needed
\hline
\addlinespace[0.15cm] % Adjust the value as needed
HELPER~\citep{sarch2023helper} + ODIN~\citep{jain2024odin} & 13.8 & 26.6 \\
\model{} (ours) + ODIN & 13.8 & 25.5 \\
\addlinespace[0.15cm] % Adjust the value as needed
\hline
\end{tabular}
\end{center}
\end{table}

\subsection{Benchmarking open-source VLMs for attribute detection in TEACh}
\begin{figure*}[t]
    \centering
    \includegraphics[width=0.3\textwidth]{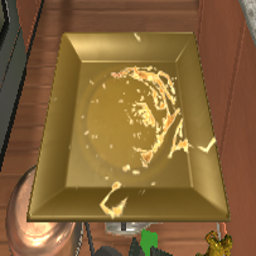}
    \includegraphics[width=0.3\textwidth]{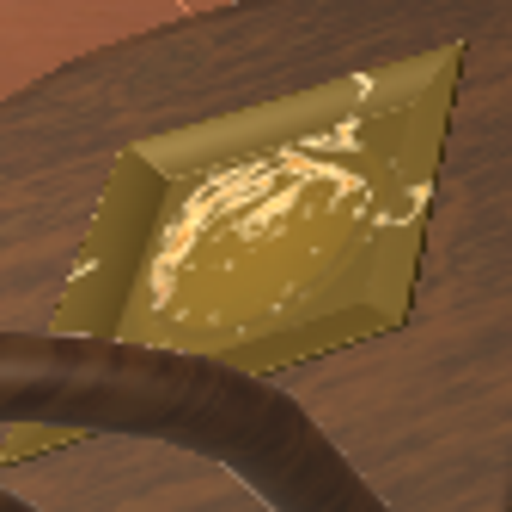}
    \caption{An example image crop of the attribute dataset collected in TEACh of a dirty plate. 
    A clean viewpoint example image is shown on the left, and a random viewpoint example image is shown on the right.
    }
    \label{fig:dirty_plate}
\end{figure*}

In household instruction following, \model{} benefits from accurate object and attribute detection from sensory input for state inference. For benchmarking object attribute detection in TEACh, we build a dataset of 2581 object crops of clean viewpoints of the object by having the agent pick up the object and centering the object in view. We build a second dataset of 661 from random viewpoints of the object in the TEACh training set with different objects varying in their "dirty" and "cooked" attributes. Clean viewpoints are always centered, unoccluded, and posed, while the random viewpoints are often occluded and show the object from different angles. Example crops for the datasets for a 'dirty plate' is shown in Figure~\ref{fig:dirty_plate}. We test the following models: \textsc{OpenCLIP CLIP-ViT-bigG-14-laion2B-39B-b160k}\citep{cherti2022reproducible}, OpenAI CLIP \textit{clip-vit-base-patch32}\citep{Radford2021LearningTV}, \textit{X-VLM}\citep{xvlm}, \textit{Llava 1.5}\citep{liu2023improvedllava}, \textit{cogVLM}\citep{wang2023cogvlm}, \textit{diffusion classifier}\citep{li2023diffusion}, \textit{Open Flamingo}\citep{awadalla2023openflamingo}. We queried CLIP by taking the best match of the image encoding with [a photo of a \{category\} that is \{word1\}, a photo of a \{category\} that is \{word2\}], where word1 and word2 are opposite attributes. We queried diffusion classifier with [a blurry photo of a \{word1\} \{category\}., a blurry photo of a \{word2\} \{category\}.], as per the paper. We queried CogVLM, Llava, and Open Flamingo with the image crop and asked it Is this \{category\} \{word1\} or \{word2\}? Provide a single word answer, either "\{word1\}" or "\{word2\}". We show the results on our evaluation dataset in Table~\ref{tab:attribute_detection}. We find CogVLM outperforms the other open-source VLMs at posed and unposed attribute detection for object crops. We use cogVLM for our estimated perception experiments for detecting object attributes.

\begin{table}[t]
\begin{center}
\footnotesize
\setlength{\tabcolsep}{2pt} % Adjusting the column separation
\caption{Attribute detection accuracy in TEACh for different open-source VLMs. We find CogVLM currently outperforms the other open-source VLMs at posed and unposed attribute detection for object crops.}\label{tab:attribute_detection}
\begin{tabular}{@{}lcc@{}}
 \\
 & \multicolumn{1}{c}{Clean Views} & \multicolumn{1}{c}{Random Views} \\
\addlinespace[0.15cm] % Adjust the value as needed
\hline
\addlinespace[0.15cm] % Adjust the value as needed
OpenCLIP CLIP-ViT-bigG-14-laion2B-39B-b160k~\citep{cherti2022reproducible} & 0.860 & 0.715 \\
OpenAI CLIP clip-vit-base-patch32~\citep{Radford2021LearningTV} & 0.758 & -- \\
X-VLM~\citep{xvlm} & 0.785 & -- \\
Llava 1.5~\citep{liu2023improvedllava} & 0.862 & 0.839 \\
cogVLM~\citep{wang2023cogvlm} & \textbf{0.898} & \textbf{0.857} \\ 
Diffusion Classifier~\citep{li2023diffusion} & 0.665 & -- \\
Open Flamingo~\citep{awadalla2023openflamingo} & 0.530 & -- \\
\end{tabular}
\end{center}
\end{table}

\section{Additional Methods Details} \label{sec:supp_meths}

\subsection{In-Context Abstraction Learning (ICAL) Algorithm}

We present the method for In-Context Abstraction Learning (ICAL) for a single trajectory in Algorithm~\ref{alg:ICAL}. Given a noisy trajectory, the method proceeds by first abstracting the trajectory through a function $F_{abstract}$, which leverages a LLM or VLM to correct errors or inefficiencies in the trajectory and generates language abstractions that capture the essence of the task, including subgoals, causal relationships, and state changes. This phase does not require interaction with the environment or humans.

Initialization sets up for the Human-In-The-Loop (HITL) phase by resetting the feedback count and success flag. The method then enters a feedback loop where the optimized trajectory is executed in the environment. If the task execution is successful, the loop breaks, and the method proceeds to update the example set with the abstracted trajectory and its associated language abstractions. Otherwise, human feedback is solicited at the point of failure to revise the trajectory and language abstractions further, utilizing the VLM again. This feedback loop continues until either the task is successfully executed or a predefined maximum number of feedback iterations is reached.

\subsubsection{Noisy Trajectories} \label{sec:traj}
% \sectionvspace

We collect a noisy sequence of observations and actions, denoted as $\xi_{noisy} = {o_0, a_0, \ldots, o_T, a_T}$, which represents a trajectory for the language-defined task we aim for our agent to learn and adapt to. 

The trajectory sequences can come from a variety of sources, including unlabeled video sequences. We can also accommodate sub-optimal or unsuccessful attempts. In our work, we identify three scenarios in which a given sequence, $\xi_{noisy}$, might be inefficient or incorrect, and using them directly as in-context examples for our LLM/VLM agents could result in poor performance:

\begin{itemize}
    \item \textbf{Human Non-experts:} We gather demonstrations from humans without domain-specific expertise. Specifically, these demonstrations ${o_0, a_0, \ldots, o_T, a_T}$ are collected from human participants who are provided with a textual instruction \(I\) and an RGB image at each time step, and are instructed to choose actions $a \in A$ to complete the instruction. These humans commit errors, choose sub-optimal actions, and may not complete the task to its fullest extent. For instance, an episode within TEACh ~\citep{TEACH} has a participant who picks up a knife and then looks up and down before placing the knife down again, a sub-optimal action sequence not required by the instruction~\citep{min2022don}.

    \item \textbf{Visual Passive Demonstrations:} Here, the agent is given a sequence of observations \(\xi = \{o_0, o_1, \ldots, o_T\}\) that lack corresponding action labels. We collect these visual demonstrations for TEACh only. We use the TEACh human demonstrations as described in the previous text and take the egocentric RGB images without actions as the observation sequence. To infer the actions executed in these demonstrations, an inverse dynamics model is applied to consecutive pair of frames $F_{idm}(o_t, o_{t+1})$, which predicts the action $a_t$ responsible for the state transition $o_t \rightarrow o_{t+1}$. Along with sub-optimal human trajectories, the inverse dynamics may make prediction errors. We trained a transformer encoder-decoder model based on the DETR~\citep{carion2020end} architecture on 300 TEACh episodes (see Section~\ref{sec:IDM} for more details).

    \item \textbf{Agent Trajectories:} In Visual Web Arena~\citep{koh2024visualwebarena}, we obtain additional demonstrations sourced from deploying our in-context VLM on new tasks. Specifically, we first run our \model{} process on 30 human demonstrations collected by non-experts. We run the \model{} process to abstract the 30 human demonstrations and then deploy our \model{} agent using the learned examples as in-context examples.
    %(see Section~\ref{sec:deploy} for deployment details). 
    Using this \model{} agent, we collect an additional 62 new trajectories on Visual Web Arena tasks and continue to run the \model{} learning on these new trajectories collected by the model.
    %We save the trajectory if the model is successful or reaches a maximum of 30 steps. 
\end{itemize}

\subsubsection{Abstraction phase implementation details} \label{sec:teach_abstract}

We present our prompt template for the VLM abstraction generation phase in Listing~\ref{prompt_abstraction_phase}.

\paragraph{TEACh.} 
We iterate through each Python program demonstration labeled with the inverse dynamics model. Given the noisy Python program, instruction, action API, and object state, the abstraction phase proceeds by prompting the LLM to 1) revise the the code for maximal efficiency and fix mistakes in the code (abstracted trajectory), 2) provide a summary of the functionality of the script (Task Decomposition \& Subgoals), 3) provide a step-by-step plan of the steps of the script (Task Decomposition \& Subgoals), 4) Add object attribute state changes to the Python program (State Changes), and 5) add abstraction comments (Task and Causal Abstractions). For state changes, we parameterize the state changes in TEACh, allowing the LLM to add a change\_state() function to the actions to indicate a change in state of the objects the agent is interacting with. Each step uses retrieved examples of successful examples previously saved in memory.

\paragraph{VisualWebArena.}
We perform the abstraction phase for each individual action taken $a_t$ (e.g., click(element), hover(element)) in each noisy trajectory $\xi_{noisy}$ obtained in VisualWebArena. Specifically, for each action in $\xi_{noisy}$, we first prompt the VLM to optionally revise the action (optimized trajectory), and output a summary and step-by-step reasoning for the chosen action (Task Decomposition \& Subgoals), given the instruction, image observation, textual state description, previous actions taken, and proposed trajectory action for the current time step. Next, we prompt the VLM to output a a predicted next state (State Changes), given the instruction, current and next image observation, current and next textual state, and the action taken $a_t$. We next prompt the VLM to output the most relevant state elements for the task instruction (State Abstraction), given the instruction, image observation, and textual state. 
 
Finally, we prompt the VLM to output a set of abstraction comments, given the full sequence of abstracted actions, task decomposition and subgoals, state changes, and abstracted state descriptions. 

\paragraph{Ego4D.} We perform the abstraction phase for each full demonstration, consisting of 20 predicted future times steps of actions. We give the VLM 3 video frames annotated with set-of-marks, the GT actions, and the actions in the video, and prompt the VLM to annotate the four types of abstractions for each example. 

\subsubsection{Human-in-the-loop phase implementation details} \label{app:hitl}

We present our prompt template for the human-in-the-loop phase revisions in Listing~\ref{prompt_hitl}.

\paragraph{TEACh.} The TEACh simulator enables fine-grained analysis of task progress. During the human-in-the-loop phase, we formulate this task progress into natural language feedback for failed actions (e.g., ``The Toaster is full right now.'' or missed task steps (e.g., ``The Pillow needs to be put onto a Sofa''). The natural language feedback, along with the instruction, object state, action API, and failed actions/code, are given to the LLM to revise the program and abstractions. 

\paragraph{VisualWebArena.} For 20 tasks for each website related to the tasks in the demonstrations collected, we develop an interface to provide natural language corrections to the model based on the predicted next action by the model. The humans are tasked to intervene and provide feedback when they deem an action predicted by the model sub-optimal. When an action is proposed, the humans are given the ability to accept the action or reject the action if it is sub-optimal. If sub-optimal, the humans can type in natural language feedback which will be sent to the VLM to revise the action and abstractions. We provide an example of the previous outputs, human feedback, and revised outputs in Listing~\ref{hitl_vwa}.

\paragraph{Ego4D.} Due to the passive nature of the Ego4D task, where there is no agent executing the actions predicted, no human in the loop phase is implemented. Ego4D only uses the abstraction phase. 

\begin{figure}
\begin{algorithm}[H]
\caption{In-Context Abstraction Learning (ICAL) method for a single trajectory 
}
\label{alg:ICAL}
\begin{algorithmic}[1]
\Require Noisy trajectory $\xi_{noisy} = \{o_0, a_0, ..., o_T, a_T\}$, Task instruction $I$, Maximum feedback iterations $N_{feedbacks}$
\Ensure Updated example set \exampleset{}

\Statex % Adds a little extra vertical space

\State \textbf{Abstraction Phase:} \Comment{Abstraction of the trajectory}
% \Indent
\State $(\xi_{abstracted}, L) \gets F_{abstract}(\xi_{noisy}, I, \{e^1, ..., e^k\})$ \Comment{Prompts VLM}
%\Comment{$F_{abstract}$ prompts an LLM/VLM}
% \EndIndent

\Statex % Adds a little extra vertical space

\State \textbf{Initialization:} \Comment{Prepare for the HITL phase}
% \Indent
\State $feedback\_count \gets 0$
\State $success \gets \text{False}$
% \EndIndent

\Statex % Adds a little extra vertical space

\While{$feedback\_count < N_{feedbacks}$ \textbf{and not} $success$} \Comment{Feedback loop}
    \State Execute $\xi_{abstracted}$ in the environment \Comment{Attempt task execution}
    \If{Task is successful}
        \State $success \gets \text{True}$
        \State \textbf{break}
    \Else
        \State Obtain human feedback $H$ \Comment{Get feedback at failure point}
        \State $(\xi_{abstracted}, L) \gets \Xi_{update}(\xi_{abstracted}, H, L, I, \{e^1, ..., e^k\})$ \Comment{Prompts VLM}
        % \Comment{$\Xi_{update}$ prompts an LLM/VLM}
        % \State $(\xi_{abstracted}, L) \gets \Xi_{update}(\xi_{abstracted}, H, L, I,$
        % \Statex \hspace{\algorithmicindent}\hspace{\algorithmicindent}\hspace{\algorithmicindent} $\{e^1, ..., e^k\})$
        % \State $(\xi_{abstracted}, L) \gets \Xi_{update}(\xi_{abstracted}, H(a_t, o_t), L, I, \{e^1, ..., e^k\})$ \algmultilinecomment{$\Xi_{update}$ prompts an LLM/VLM}
        %\algcomment{$\Xi_{update}$ prompts an LLM/VLM without overflowing the text into the margins or onto the next page.}
        \State $feedback\_count \gets feedback\_count + 1$
    \EndIf
\EndWhile

\Statex % Adds a little extra vertical space

\If{$success$}
    \State Update example set $\exampleset \gets \exampleset \cup \{(\xi_{abstracted}, L)\}$ \Comment{Store the abstracted example}
\EndIf

\end{algorithmic}
\end{algorithm}
\end{figure}

\lstset{escapeinside={<@}{@>}, language=}
\onecolumn\begin{lstlisting}[caption={Example of human-in-the-loop revision for visual web arena.},captionpos=t,label={hitl_vwa}] 

<@\textcolor{orange}{Instruction:}@> Show me the most recently posted blue iPhone.

-----------------------------------------------------------------------------------------

<@\textcolor{orange}{Abstractions \& action before human feedback:}@>

Abstraction Comments:
- When searching for the most recently posted item of a specific color, ensure the listings are sorted by the newest first and visually inspect the images for the color before clicking on the correct listing.

Plan: To find the most recently posted blue iPhone, I need to use the search function on the classifieds website. The search input field is identified by [5]. I will type "blue iPhone" into the search box to filter the listings for items related to blue iPhones. This will help me find the most recently listed blue iPhone.

Summary: Typing "blue iPhone" into the search field will filter the listings to show relevant items, allowing progress towards identifying the most recently posted blue iPhone.

Predicted Next State: After typing "blue iPhone" into the search field and initiating the search, the next state will display search results related to blue iPhones, potentially including listings with images and descriptions of the items.

Action: In summary, the next action I will perform is ```type [5] [blue iPhone]```

-----------------------------------------------------------------------------------------

<@\textcolor{orange}{Human feedback:}@> "That search likely will not work as well as clicking on the Cell phones button and searching for the iPhone there on this classifieds site."

-----------------------------------------------------------------------------------------

<@\textcolor{orange}{Abstractions \& action before human feedback:}@>

Abstraction Comments:
- When searching for the most recently posted item of a specific color, ensure the listings are sorted by the newest first and visually inspect the images for the color before clicking on the correct listing.
- When searching for specific items on a classifieds website, it is more effective to narrow down the search by selecting the appropriate category before typing in the search term.

Plan: Let's think step-by-step. The objective is to find the most recently posted blue iPhone. The feedback indicates that I should first narrow down the category by clicking on the 'Cell phones' category, which has the ID [41]. After selecting the category, I can then search for 'blue iPhone' within that category.

Summary: To find the most recently posted blue iPhone, I need to first narrow down the search by selecting the 'Cell phones' category.

Predicted State: After clicking on the 'Cell phones' category, the page will display listings under that category, allowing for a more targeted search for a blue iPhone.

Revised Action: In summary, the next action I will perform is ```click [41]```

\end{lstlisting}

\subsection{Deploying the \model{} agent after the examples have been learned.}
We present our algorithm for deploying our \model{} agent after the examples have been learned on new instructions in Algorithm~\ref{alg:inference}. We additional present our prompt template for the VLM planning after examples have been learned in Listing~\ref{prompt_planner}.

\begin{algorithm}
\caption{Deploying the \model{} agent after the examples have been learned.}
\label{alg:inference}
\begin{algorithmic}[1]
\Require
\Statex Predefined action API based on skill set $A$.
\Statex Set of in-context examples $\exampleset = \{e^1, e^2, \ldots, e^k\}$.
\Statex Language instruction $I$.
\Statex Initial observation $o_0$.
\Statex Initial textual observation $x_0$.
\Statex Maximum steps $T$.
\State Initialize observation $o_t \gets o_0$, $x_t \gets x_0$.
\For{$t = 0$ to $T-1$}
    \State Retrieve top $K$ examples: $\{e^1_t, \ldots, e^K_t\} \gets \text{RetrieveTopK}(x_t, o_t, I, \exampleset)$.
    \State $a_t \gets \text{VLM}(x_t, o_t, I, \{e^1_t, \ldots, e^K_t\}, A)$ to generate an action or Python code.
    \State Execute action $a_t$ to receive new observation $o_{t+1}$, $x_{t+1}$.
    \If{stop criteria met}
        \State \textbf{break}
    \EndIf
\EndFor
\end{algorithmic}
\end{algorithm}

\subsection{Inverse Dynamics Model} \label{sec:IDM}
\begin{figure*}[t]
    \centering
    \includegraphics[width=\textwidth]{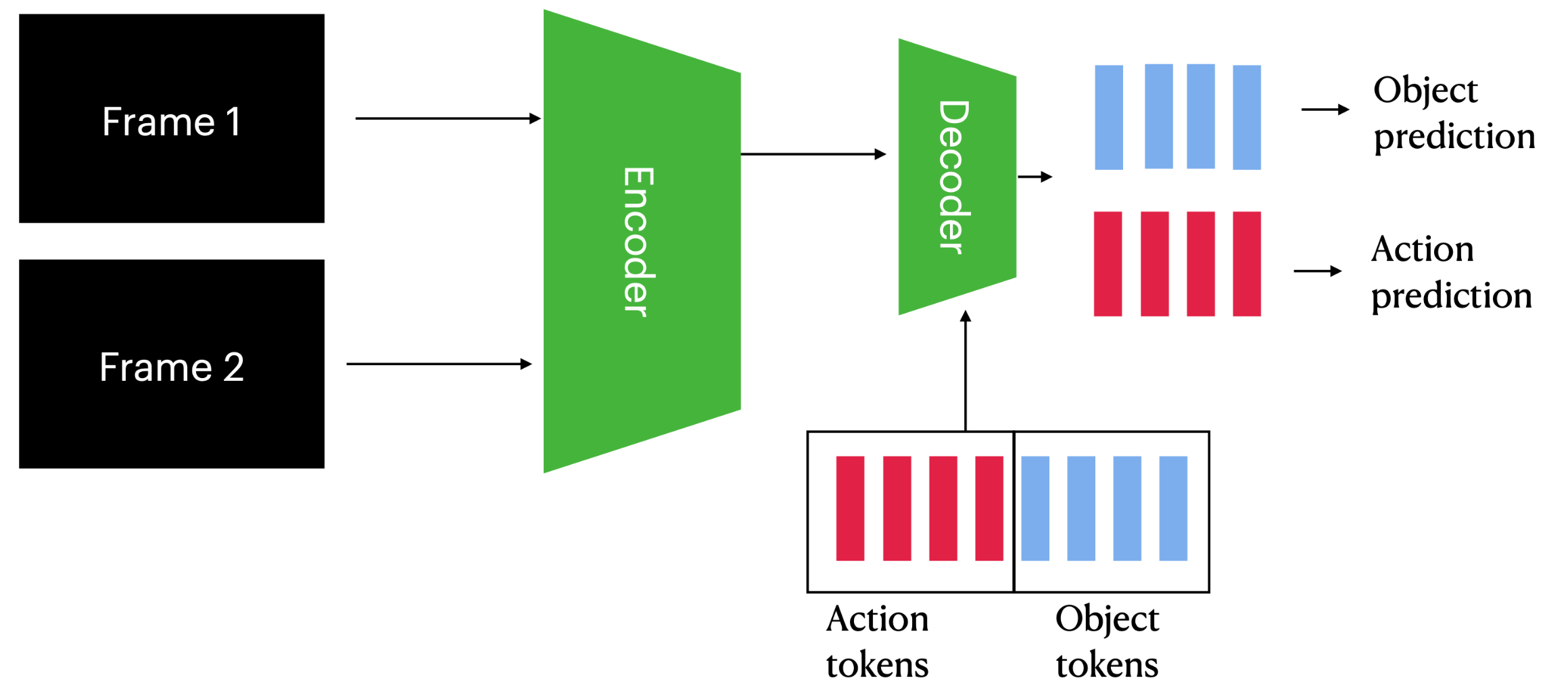}
    \caption{Architecture of inverse dynamics model used for labeling TEACh trajectories.}
    \label{fig:idm}
\end{figure*}

% \katef{let's go through it together}
In this section, we provide implementation details for the inverse dynamics model used in TEACh. A high-level architecture diagram is shown in Figure~\ref{fig:idm}

\textit{Backbone.} Given an input image pair \(x_{\text{images}} \in \mathbb{R}^{2 \times 3 \times H_0 \times W_0}\) (2 frames and 3 color channels), we use a CNN  to produce  lower-resolution activation maps \(f \in \mathbb{R}^{2 \times C \times H \times W}\), where \(H_0\) and \(W_0\) denote the original height and width, respectively, and \(H\) and \(W\) represent the dimensions of the resulting feature map.

\textit{Transformer Encoder.} 
The spatial features are input into a transformer encoder, where they undergo self-attention. We reshape the spatial dimensions into a single dimension, resulting in a feature map with dimensions \(d \times HW\). Each layer of the encoder consists of a multi-head self-attention mechanism and a feed-forward network. Fixed spatial positional encodings and learned frame encodings are added to the inputs at each attention layer. The transformer encoder comprises six self-attention layers, utilizes eight heads, has an embedding size of 384, and contains six encoder layers.

\textit{Transformer Decoder.} 
The decoder incorporates cross-attention mechanisms for both object and action queries with respect to the encoder features. It processes \(N\) queries simultaneously across its layers. The embeddings, comprising action queries and object queries, add learned positional encodings at the input of each attention layer. Each layer of the decoder includes cross-attention from the queries to the encoder features, self-attention among the query features, and a feed-forward network. The transformer decoder consists of six self-attention layers, employs eight heads, has an embedding size of 384, and includes six encoder layers.

After the decoder, we use a feed-forward network to reduce each embedding from dimension \(d\) to a scalar value. These scalar values for actions and objects are concatenated, creating final action logits for each action and each object. The model is trained using cross-entropy loss for both actions and objects, such as 'pickup' and 'apple'. Additionally, we introduce an extra query for 'no object' to accommodate actions that do not involve manipulating an object (e.g., move\_forward()).

\textit{Dataset.} We use a random subset of 649 training episodes from the TEACh training dataset and 181 validation episodes from the TEACh validation seen dataset, which do not overlap with the episodes used for the \model{} example learning. We use Each episode contains a trajectory of observations and actions $\{o_0, a_0, \ldots, o_T, a_T\}$. We use each pair of observations ($o_t$, $o_{t+1}$), and the action $a_t$ responsible for the state transition $o_t \rightarrow o_{t+1}$, as training samples for the network. 
% $F_{idm}(o_t, o_{t+1})$, which predicts 

\textit{Implementation details.} We use a learning rate of $2e-5$, batch size of 64, a step learning rate scheduler with $\gamma$ = $0.1$ and step size = $30$ epochs. We use early stopping based on validation loss and train for 45 epochs. We use cross entropy loss with a manual class weight rescaling based on frequencies in the training set. Training and model implementation is done in PyTorch.

\textit{Applying Inverse Dynamics Model on held-out demonstrations.}
On the held out 250 TEACh episodes used for \model{} example learning, we feed each pair of observations to the trained inverse dynamics model to predict actions. 
We convert the sequence of predicted actions into a Python program based on the \model{} action API for TEACh. This involves converting each action into a Python function and aggregating contiguous navigation actions (move\_forward(), turn\_left()) into a single go\_to() function in the program. An example of the predicted Python program is shown in Listing~\ref{example_before_cleanup}. We also provide the fully revised program after running \model{} learning on the predicted program in Listing~\ref{example_after_cleanup}.
% \begin{minipage}{.48\linewidth}
\lstset{escapeinside=, language=Python}
\onecolumn\begin{lstlisting}[caption={Demonstration program of actions inferred from the inverse dynamics model for an episode of making a Salad.},captionpos=t,label={example_before_cleanup}] 
target_fridge = InteractionObject("Fridge", object_instance = "Fridge")
target_fridge.go_to()
target_fridge.open()
target_lettuce = InteractionObject("Lettuce", object_instance = "Lettuce")
target_countertop = InteractionObject("CounterTop", object_instance = "CounterTop")
target_lettuce.pickup_and_place(target_countertop)
target_knife = InteractionObject("Knife", object_instance = "Knife")
target_knife.pickup()
target_lettuce.go_to()
target_lettuce.slice()
target_bread = InteractionObject("Bread", object_instance = "Bread")
target_bread.go_to()
target_bread.slice()
target_countertop.go_to()
target_knife.place(target_countertop)
target_tomato = InteractionObject("Tomato", object_instance = "Tomato")
target_tomato.pickup_and_place(target_countertop)
target_knife.pickup()
target_bread.go_to()
target_bread.slice()
target_countertop.go_to()
target_knife.place(target_countertop)
target_breadsliced = InteractionObject("BreadSliced", object_instance = None, parent_object = "Bread")
target_plate = InteractionObject("Plate", object_instance = "Plate")
target_breadsliced.pickup_and_place(target_plate)
target_breadsliced.pickup()
target_pot = InteractionObject("Pot", object_instance = "Pot")
target_pot.go_to()
target_breadsliced.place(target_pot)
target_breadsliced.pickup_and_place(target_plate)
target_spoon = InteractionObject("Spoon", object_instance = "Spoon")
target_spoon.pickup()
target_countertop.go_to()
target_spoon.place(target_countertop)
target_lettucesliced = InteractionObject("LettuceSliced", object_instance = None, parent_object = "Lettuce") parent
target_lettucesliced.pickup_and_place(target_plate)
target_lettucesliced.pickup()
target_plate.go_to()
target_lettucesliced.place(target_plate)
target_tomatosliced = InteractionObject("TomatoSliced", object_instance = "TomatoSliced")
target_tomatosliced.pickup_and_place(target_countertop)
\end{lstlisting}
% \end{minipage}
% \begin{minipage}{.48\linewidth}
\lstset{escapeinside=, language=Python}
\onecolumn\begin{lstlisting}[caption={Revised demonstration program (revised from the program in Listing S1) after abstraction phase cleanup and human-in-the-loop for an episode of making a salad.},captionpos=t,label={example_after_cleanup}] 

# Initialize InteractionObject instances for the fridge, lettuce, knife, tomato, and plate
fridge = InteractionObject("Fridge", object_instance="Fridge_71")
lettuce = InteractionObject("Lettuce", object_instance="Lettuce_11")  # Lettuce in the fridge
knife = InteractionObject("Knife", object_instance="Knife_73")  # Knife on the countertop
tomato = InteractionObject("Tomato", object_instance="Tomato_80")  # Tomato on the table
plate = InteractionObject("Plate", object_instance="Plate_66")  # Plate on the countertop
sink_basin = InteractionObject("SinkBasin", object_instance="SinkBasin_74")  # Sink basin
faucet = InteractionObject("Faucet", object_instance="Faucet_87")  # Faucet

# Go to the fridge and open it to retrieve the lettuce
fridge.go_to()
fridge.open()

# Retrieve two lettuces from the fridge, slice them, and create instances for the sliced lettuce
lettuce.go_to()
lettuce.pickup()
knife.go_to()
knife.pickup()
lettuce.slice()
lettuce_sliced_1 = InteractionObject("LettuceSliced", parent_object=lettuce.object_instance)  # Initialize new sliced object from sliced parent
lettuce_sliced_2 = InteractionObject("LettuceSliced", parent_object=lettuce.object_instance)  # Initialize new sliced object from sliced parent

# Go to the tomato on the table, slice it, and create an instance for the sliced tomato
tomato.go_to()
tomato.slice()
tomato_sliced = InteractionObject("TomatoSliced", parent_object=tomato.object_instance)  # Initialize new sliced object from sliced parent

# Check if the plate is dirty and clean it if necessary
plate.go_to()
if plate.check_attribute("dirty", True):
    sink_basin.go_to()
    plate.place(sink_basin)
    faucet.go_to()
    faucet.toggle_on()  # Turn on the faucet to clean the plate
    faucet.toggle_off()  # Turn off the faucet after cleaning
    plate.pickup()  # Pick up the clean plate
    plate.change_state("dirty", False)  # Update the plate's state to clean

# Place two slices of lettuce and one slice of tomato on the clean plate
lettuce_sliced_1.go_to()
lettuce_sliced_1.pickup()
plate.go_to()
lettuce_sliced_1.place(plate)
lettuce_sliced_2.pickup()
lettuce_sliced_2.place(plate)
tomato_sliced.go_to()
tomato_sliced.pickup()
tomato_sliced.place(plate)

# Close the fridge after retrieving the items
fridge.go_to()
fridge.close()
\end{lstlisting}
% \end{minipage}

% \onecolumn\begin{lstlisting}[caption={Full API for the parametrized macro-actions used in TEACh.},captionpos=t,label={TEACH_API}] 
% example text
% \end{lstlisting}

% \onecolumn\begin{lstlisting}[caption={Full API for the parametrized macro-actions used in TEACh.},captionpos=t,label={TEACH_API}] 
% example text
% \end{lstlisting}

\subsection{LLM finetuning details} \label{app:finetune}
We use the examples obtained from the \model{} method applied in TEACh, a total of 122 examples. We split the dataset randomly into 99 training samples and 23 validation samples. Input tokens for training consist of each example (instruction,  object state, and API) with the prompt template used for zero-shot prompting. Output tokens consist of the Python program with abstraction comments for each example The mean input token length per sample is 3145.17, while the mean output token length per sample is 432.82. We use the Azure OpenAI Service for fine-tuning, which uses the next-token prediction objective and LoRA~\citep{hu2021lora} for parameter-efficient finetuning of gpt-35-turbo-1106. 

\subsection{Compute Resources}
We use a single Nvidia RTX A6000 for training the inverse dynamics model and running all our evaluations. We use Azure for finetuning GPT-3.5-1106 as mentioned in Section~\ref{app:finetune}. We use Azure OpenAI API for VLM inference. 

\section{Additional implementation details} \label{sec:supp_envs}

\subsection{TEACh} \label{Simulator}
The TEACh dataset builds on the Ai2thor simulation environment~\citep{ai2thor}. At each time step the agent may choose from the following actions: Forward(), Backward(), Turn Left(), Turn Right(), Look Up(), Look Down(), Strafe Left(), Strafe Right(), Pickup(X), Place(X), Open(X), Close(X), ToggleOn(X), ToggleOff(X), Slice(X), and Pour(X), where X refers an object specified via a relative coordinate $(x, y)$ on the egocentric RGB frame. Navigation actions move the agent in discrete steps. We rotate in the yaw direction by 90 degrees, and rotate in the pitch direction by 30 degrees. The RGB and depth sensors are at a resolution of 480x480, a field of view of 90 degrees, and lie at a height of 0.9015 meters. The agent's coordinates are parameterized by a single $(x,y,z)$ coordinate triplet with $x$ and $z$ corresponding to movement in the horizontal plane and $y$ reserved for the vertical direction. The TEACh benchmark allows a maximum of 1000 steps and 30 API failures per episode.

\subsubsection{Planning at test time}
Given a new environment and instruction, \model{} first maps out the scene to build a navigation map and detect objects and their attributes (see next sections). \model{} then retrieves the top-$k$ examples relevant to the instruction and object state (see Section~\ref{sec:retrieve}). \model{} then obtains the abstracted object state to give to the LLM (see Section~\ref{sec:teach_abstract}). \model{} then prompts the LLM, given the instruction, abstracted object state, and retrieved in-context examples, to output Python code to carry out the new instruction in the environment. 
If code execution failures occur, we re-prompt the LLM with the execution error and ask the LLM to revise the code. 

\subsubsection{\model{} differences with HELPER}
In TEACh, we build on HELPER~\citep{sarch2023helper} for program execution. Here, we give an account of HELPER. HELPER prompts an LLM, namely GPT-4~\citep{gpt4technical}, to generate plans as Python programs. It assumes that the agent has access to a set of action skills $S$ (e.g., \texttt{go\_to(X)}, \texttt{pickup(X)}, etc.). We use a reduced set of these skills (e.g., we remove the cook(), clean(), and toast() primitives as we wish for our model to learn these). 
HELPER generates code that is decomposed into these action skills. Instead of demoposing them into action primitives, we run the Python code generated from the LLM directly (i.e., using the 'exec' function in Python). Each action skill comes with a set of pre-engineered pre-condition checks, which we also remove. HELPER maintains a 3D semantic map for navigation and keeping track of objects (see next sections).

\subsubsection{Obstacle map} 
\model{} maintains a 2D overhead occupancy map of its environment $\in \mathbb{R}^{H \times W}$ that it updates at each time step from the input RGB-D stream. The map is used for exploration and navigation in the environment. At every time step $t$, we unproject the input depth maps using intrinsic and extrinsic information of the camera to obtain a 3D occupancy map registered to the coordinate frame of the agent, similar to earlier navigation agents \citep{chaplot2020learning}. The 2D overhead maps of obstacles and free space are computed by projecting the 3D occupancy along the height direction at multiple height levels and summing. For each input RGB image, we run a SOLQ object segmentor~\citep{dong2021solq} (pretrained on COCO~\citep{lin2014microsoft} then finetuned on TEACh rooms) to localize each of 116 semantic object categories. For failure detection, we use a simple matching approach from \citep{min2021film} to compare RGB pixel values before and after taking an action. When using ground truth perception, we use ground truth semantic segmentation, depth maps, object attributes, and action failure detection.

\subsubsection{Object location and state tracking}
We maintain an object memory as a list of object detection 3D centroids and their predicted semantic labels $\{ [ (X,Y,Z)_i, \ell_i\in\{1...N\} ] , i=1..K  \}  $, where $K$ is the number of objects detected thus far.
The object centroids are expressed with respect to the coordinate system of the agent, and, similar to the semantic maps, updated over time using egomotion. We track previously detected objects by their 3D centroid $C \in \mathbb{R}^{3}$. We estimate the centroid by taking the 3D point corresponding to the median depth within the segmentation mask and bring it to a common coordinate frame. We do a simple form of non-maximum suppression on the object memory, by comparing the euclidean distance of centroids in the memory to new detected centroids of the same category, and keep the one with the highest score if they fall within a distance threshold.

For each object in the object memory, we maintain an object state dictionary with a pre-defined list of attributes. These attributes include: category label, centroid location, holding, detection score, can use, sliced, toasted, clean, cooked. For the attributes, these are initialized by sending the detected object crops in the abstracted state, defined by the detector mask, to the VLM model, and asking it "Is this \{category\} \{word1\} or \{word2\}? Provide only your answer, either "\{word1\}" or "\{word2\}", and taking the answer as the output attribute.

\subsection{VisualWebArena} \label{VWA_env}
The VisualWebArena~\citep{koh2024visualwebarena} builds on Web Arena~\citep{zhou2023webarena} contains 910 evaluation instructions with three interactive websites: Classifieds, Reddit, and Shopping. At each time step, the agent obtains a Set of Marks annotated image and and the webpage content, in a textual format listing the button text with their set of marks ID. The set of marks bounding boxes and textual state are extracted from the HTML code for the current webpage. At each time step, the agent must select an action to carry out the instruction. The instruction includes a natural language description and potentially one or more reference images.
The action space is as follows:
\begin{itemize}
\item click [elem] Click on element elem.
\item hover [elem] Hover on element elem.
\item type [elem] [text] Type text on element elem.
\item press [key comb] Press a key combination.
\item new tab Open a new tab.
\item tab focus [index] Focus on the i-th tab.
\item tab close Close current tab.
\item goto [url] Open url.
\item go back Click the back button.
\item go forward Click the forward button.
\item scroll [up|down] Scroll up or down the page.
\item stop [answer] End the task with an optional output
\end{itemize}

\subsection{Additional details on \model{} agent deployment in VisualWebArena}
At each time step, we retrieve the top-5 examples and prompt the model with the 5 in-context examples. Each in-context example consists of the image input, abstracted textual state, summary, step-by-step reasoning, predicted next state, abstraction comments, and predicted action. We use the Set of Marks (SoM)~\citep{yang2023setofmark} representation for image inputs, implemented in VisualWebArena by \citep{koh2024visualwebarena}. 
%An example of the set of marks image input is show in Figure~\ref{fig:set_of_marks}. 
An example in-context example is shown in Listing~\ref{example_incontext_vwa}.

% \begin{figure*}[t!]
%     \centering
%     \includegraphics[width=\textwidth]{supplemental_figures/screenshots_websites.jpg}
%     \caption{Example Set of Marks images for Classifieds site (left) and Reddit site (right).}
%     \label{fig:set_of_marks}
% \end{figure*}

\lstset{escapeinside={<@}{@>}, language=}
\onecolumn\begin{lstlisting}[caption={In-context example used in VisualWebArena. Note that the webpage screenshot with SoM annotations for the in-context example is also provided to the VLM, but is not displayed.},captionpos=t,label={example_incontext_vwa}] 
Input:

OBJECTIVE: I recall seeing this exact item of pillows in the Household section on the site, add a comment on its listing with the title "Commentary" and text "How funky...".

OBSERVATION:
[4] [A] [Publish Ad]
[] [StaticText] [> Search results: pillows]
[8] [INPUT] []
[] [StaticText] [Min.]
[15] [A] [Household]
[18] [A] []
[] [StaticText] [Listings]
[] [StaticText] [North Potomac  (Maryland)]
[] [StaticText] [15.00 $]
[] [StaticText] [Pottery Barn Matine Drape (1 panel)                      20.00 $                          Household              Lake Barcroft  (Virginia)               2023/11/11                                Beaut]
[] [StaticText] [100.00 $]
[33] [IMG] [Comforter with Pillows, description: a green quilt and pillows on the ground, url: oc-content/uploads/1243/1243_thumbnail.png]
[] [StaticText] [Roanoke  (Virginia)]
[] [StaticText] [Comforter with 2 Shams and 1 Throw Pillow 7'1 x 7'3]

PREVIOUS ACTIONS: 
1. type  [INPUT] [] [pillows]
2. hover  [SELECT] [Select a categoryAntiquesAppliancesArts + craftsAuto partsBeauty + healthBikesBoatsBooksCars + trucksCell phonesCollectiblesComputer partsComputersElectronicsFarm + gardenFurnitureHouseholdJewelryMoto]
3. click  [SELECT] [Select a categoryAntiquesAppliancesArts + craftsAuto partsBeauty + healthBikesBoatsBooksCars + trucksCell phonesCollectiblesComputer partsComputersElectronicsFarm + gardenFurnitureHouseholdJewelryMoto]
4. click  [A] [Household]
5. type  [INPUT] [] [pillows]
6. click  [BUTTON] [Apply]
7. scroll [down]
8. click [25] [A] 
9. scroll [down]
10. click  [A] [2]
11. click  [A] [Household]
12. scroll [down]
13. click  [A] [2]

Output:

Plan: I need to find the exact item of pillows with the title "Commentary" to add a comment. I've already searched for pillows, filtered by the Household category, and navigated through the pages. The item I'm looking for is not visible on the current screen, so I must continue to scroll down to find it. Once I locate the item, I can proceed to add the comment with the specified text "How funky...".

Summary: Scrolling down will allow me to find the exact pillows listing where I need to add the comment, progressing towards the task completion.

Predicted Next State: The next state will display additional listings of household items, including more pillow options, as the page is scrolled down to reveal further content that was previously not visible.

Action: In summary, the next action I will perform is ```scroll [down]```

Abstraction comments:
1. When the end of a page is reached and the task requires viewing more items, navigate to the next page instead of attempting to scroll further.
2. When posting a comment on a classifieds website, ensure that all required fields, such as the email address, are filled in before submitting the comment.
3. When searching for items in a specific location on a classifieds website, it is important to use the location filter to narrow down the search results to the desired area before applying the filter.
4. When a comment has been typed into the appropriate field, the next step is to submit the comment using the provided submission button, rather than retyping or editing the comment.
5. When searching for items within a specific price range, it is essential to set the minimum and maximum price filters before applying the search to narrow down the results.
6. When tasked with selecting a specific item in a sequence, ensure the correct order is followed based on the given instructions. In this case, the item must be selected (clicked on) before completing the task.
7. When the objective is to leave a comment with both a title and text, ensure that both fields are completed before submitting the comment.
\end{lstlisting}

\subsection{Ego4D}
Ego4D is a daily life activity video dataset of hundreds of scenarios. We focus on the long-term action anticipation task to predict the future user actions given an RGB egocentric video. Models must choose from 115 verbs
and a set of 478 nouns for action predictions. For evaluation, we take 100 seen validation videos that come from the same videos used for \model{} example learning but at a different, unseen location, and a separate 200 completely unseen validation videos for evaluation. We follow previous work and use edit distance as a metric, which is computed as the Damerau-Levenshtein distance over sequences of predictions of verbs, nouns and actions. The goal of this measure is to assess performance in a way which is robust to some error in the predicted order of future actions. All GPT4V evaluations give image inputs annotated with DEVA tracking masks~\citep{cheng2023tracking} with Set-of-Marks labels~\citep{yang2023setofmark}. For in-context examples to GPT4V, we concatenate 3 uniformly spaced video frames and give it as a single image input. For the input video to GPT4V, we take 12 video frames uniformly spaced and provide 4 images each with 3 concatenated frames. The supervised baseline uses a SlowFast backbone with a Transformer aggregator and trains on Ego4D V2 (243 video hrs)~\citep{grauman2022ego4d}.

\subsubsection{Noisy Trajectories} 100 demonstrations from validation set were abstracted using \model{}. Due to the passive nature of this task, we perform \model{} without the abstraction verification with a human-in-the-loop phase, and only perform the VLM-driven Abstraction Generation (Section~\ref{sec:introspection}). 92 demonstrations (8 failed due to GPT4V filters) were successfully abstracted by \model{} for an example set size of 92 for evaluation.

\lstset{escapeinside={<@}{@>}, language=}
\onecolumn\begin{lstlisting}[caption={Prompt template for VLM abstraction generation phase},captionpos=t,label={prompt_abstraction_phase}] 
**Objective:** As a helpful assistant with expertise in {DOMAIN}, your task is to produce useful abstractions and language comments to help someone else perform the task.

**Information Provided:**
You will receive:
{INPUT INFORMATION}

**Output Format:**
1. Summary: Provide a summary of the task the user is performing. Start this with 'Summary:" and limit it to a single line, no more than 6 sentences.
2. Abstracted State: List the elements that are relevant for the task that the user is performing and are important for the task. Refer to the elements by their object ID, and for each element, a description of the object and and relevant attributes. Start the list with 'Abstracted State:', and put each element that you choose on a new line.
3. Step-by-step Reasoning: Explain each step of the demonstration and the reasoning for each step. Mention specific object numerical IDs when referencing objects. Start this section with "Step-by-step Reasoning:" and limit it to a single line, no more than 6 sentences.
4. Predicted State Change: Provide in natural language any relevant state changes of objects and visual elements that will take place due to future actions. Remember to focus on state changes that will help someone else perform the task.
5. Abstraction Comments: Provide a numbered list of useful language abstraction comments, such as causal abstractions, task abstractions, and other abstractions that will help someone learn the task. Put each abstraction on a new line. Mention specific object IDs when referencing objects.
6. Optimized Demonstration Script: Present any optimized actions for completing the task more efficiently in the current environment. It is possible that the provided demonstration script is already optimally efficient and no revisions are needed.

**Action Space**
{ACTION API}

**In-Context Examples:**
{EXAMPLES}

**Guidelines:**
Follow these strict guidelines:
1. Adhere to the previously defined output format without deviating. Refer to the examples provided for proper format.
2. Reason through each step methodically, as shown in examples.
3. Reference object/part IDs in your reasoning when it's relevant.
4. Your primary focus should be on generating useful comments that will help someone else accurately perform the task.
\end{lstlisting}

\lstset{escapeinside={<@}{@>}, language=}
\onecolumn\begin{lstlisting}[caption={Prompt template for human-in-the-loop revisions based on human feedback},captionpos=t,label={prompt_hitl}] 
**Objective:** You are an autonomous intelligent agent tasked with {DOMAIN}. Your primary goal is to revise an action taken on a website based on natural language corrective feedback so that the action successfully makes progress towards completing the task. 

**Information Provided:**
Here's the information you'll have:
{INPUT INFORMATION}

**Output Format:**
1. Explain: Why does the action not complete the task? What does the human feedback imply? What revisions should be made to fix the error? This should be a single line, and at most six sentences.
2. Summary: Single-line summary of what the proposed new action will carry out and how it will make progress towards the objective.
3. Abstracted State: List the elements that are relevant for the task that the user is performing and are important for the task. Refer to the elements by their object ID, and for each element, a description of the object and and relevant attributes.
4. Step-by-step Reasoning: Explain each step of the demonstration, the reasoning for each step, and why the revised action would make the most sense.
5. Predicted State Change: Predict what the next state will look like after taking the proposed revised action.
6. Correction Abstraction: 
Provide a numbered list of useful language abstraction comments, such as causal abstractions, task abstractions, and other abstractions that will help someone learn the task. Put each abstraction on a new line. Mention specific object IDs when referencing objects. Also, incorporate the correction into some generalizable knowledge about the error, why it is a mistake, and how to fix it.
7. Revised Action: Output the revised action to take from the actions provided below. 

**Action Space**
{ACTION API}

**In-Context Examples:**
{EXAMPLES}

**Guidelines:**
Follow these strict guidelines:
1. Adhere to the previously defined output format without deviating. Refer to the examples provided for proper format.
2. Reason through each step methodically, as shown in examples.
3. Reference object/part IDs in your reasoning when it's relevant.
4. Your primary focus should be on generating useful comments that will help someone else accurately perform the task.
\end{lstlisting}

\lstset{escapeinside={<@}{@>}, language=}
\onecolumn\begin{lstlisting}[caption={Prompt template for VLM planning after examples are learned},captionpos=t,label={prompt_planner}] 
**Objective:** As a helpful assistant with expertise in {DOMAIN}, your task is to {DOMAIN TASK}

**Information Provided:**
You will receive:
{INPUT INFORMATION}

**Output Format:**
1. Summary: Provide a summary of the task you are performing. Start this with 'Summary:" and limit it to a single line, no more than 6 sentences.
2. Abstracted State: List relevant objects in the scene by their numerical IDs, providing a description and any pertinent attributes for each. Start the list with 'Abstracted State:', and put each element that you choose on a new line.
3. Step-by-step Reasoning: Explain each step of the task and the reasoning for each step. Mention specific object numerical IDs when referencing objects. Start this section with "Step-by-step Reasoning:" and limit it to a single line.
4. Predicted State Change: Provide in natural language any relevant state changes that will occur throughout the task.
5. Abstraction Comments: Provide a numbered list of useful language abstraction comments, such as causal abstractions, task abstractions, and other abstractions that will help someone learn to predict the future actions from the egocentric video. Put each abstraction on a new line. Mention specific object numerical IDs when referencing objects.
6. Predicted Actions: Present the actions the agent should take to carry out the task.

**Action Space:**
{ACTION API}

**In-Context Examples:**
{RETRIEVED EXAMPLES}

**Guidelines:**
Follow these strict guidelines:
1. Adhere to the previously defined output format without deviating. Refer to the examples provided for proper format.
2. Reason through each step methodically, as shown in examples.
3. Reference object/part IDs in your reasoning when it's relevant.
\end{lstlisting}

\clearpage
\section*{NeurIPS Paper Checklist}

\begin{enumerate}

\item {\bf Claims}
    \item[] Question: Do the main claims made in the abstract and introduction accurately reflect the paper's contributions and scope?
    \item[] Answer: \answerYes{} % Replace by \answerYes{}, \answerNo{}, or \answerNA{}.
    \item[] Justification: The abstract and introduction outline the proposed method (ICAL), its objectives, the types of abstractions it handles, and the benchmarks used for evaluation, all of which are detailed further in the paper. 
    \item[] Guidelines:
    \begin{itemize}
        \item The answer NA means that the abstract and introduction do not include the claims made in the paper.
        \item The abstract and/or introduction should clearly state the claims made, including the contributions made in the paper and important assumptions and limitations. A No or NA answer to this question will not be perceived well by the reviewers. 
        \item The claims made should match theoretical and experimental results, and reflect how much the results can be expected to generalize to other settings. 
        \item It is fine to include aspirational goals as motivation as long as it is clear that these goals are not attained by the paper. 
    \end{itemize}

\item {\bf Limitations}
    \item[] Question: Does the paper discuss the limitations of the work performed by the authors?
    \item[] Answer: \answerYes{} % Replace by \answerYes{}, \answerNo{}, or \answerNA{}.
    \item[] Justification: We discuss the limitations in the conclusion and Appendix.
    \item[] Guidelines:
    \begin{itemize}
        \item The answer NA means that the paper has no limitation while the answer No means that the paper has limitations, but those are not discussed in the paper. 
        \item The authors are encouraged to create a separate "Limitations" section in their paper.
        \item The paper should point out any strong assumptions and how robust the results are to violations of these assumptions (e.g., independence assumptions, noiseless settings, model well-specification, asymptotic approximations only holding locally). The authors should reflect on how these assumptions might be violated in practice and what the implications would be.
        \item The authors should reflect on the scope of the claims made, e.g., if the approach was only tested on a few datasets or with a few runs. In general, empirical results often depend on implicit assumptions, which should be articulated.
        \item The authors should reflect on the factors that influence the performance of the approach. For example, a facial recognition algorithm may perform poorly when image resolution is low or images are taken in low lighting. Or a speech-to-text system might not be used reliably to provide closed captions for online lectures because it fails to handle technical jargon.
        \item The authors should discuss the computational efficiency of the proposed algorithms and how they scale with dataset size.
        \item If applicable, the authors should discuss possible limitations of their approach to address problems of privacy and fairness.
        \item While the authors might fear that complete honesty about limitations might be used by reviewers as grounds for rejection, a worse outcome might be that reviewers discover limitations that aren't acknowledged in the paper. The authors should use their best judgment and recognize that individual actions in favor of transparency play an important role in developing norms that preserve the integrity of the community. Reviewers will be specifically instructed to not penalize honesty concerning limitations.
    \end{itemize}

\item {\bf Theory Assumptions and Proofs}
    \item[] Question: For each theoretical result, does the paper provide the full set of assumptions and a complete (and correct) proof?
    \item[] Answer: \answerNA{} % Replace by \answerYes{}, \answerNo{}, or \answerNA{}.
    \item[] Justification: The paper does not include theoretical results.
    \item[] Guidelines:
    \begin{itemize}
        \item The answer NA means that the paper does not include theoretical results. 
        \item All the theorems, formulas, and proofs in the paper should be numbered and cross-referenced.
        \item All assumptions should be clearly stated or referenced in the statement of any theorems.
        \item The proofs can either appear in the main paper or the supplemental material, but if they appear in the supplemental material, the authors are encouraged to provide a short proof sketch to provide intuition. 
        \item Inversely, any informal proof provided in the core of the paper should be complemented by formal proofs provided in appendix or supplemental material.
        \item Theorems and Lemmas that the proof relies upon should be properly referenced. 
    \end{itemize}

    \item {\bf Experimental Result Reproducibility}
    \item[] Question: Does the paper fully disclose all the information needed to reproduce the main experimental results of the paper to the extent that it affects the main claims and/or conclusions of the paper (regardless of whether the code and data are provided or not)?
    \item[] Answer: \answerYes{} % Replace by \answerYes{}, \answerNo{}, or \answerNA{}.
    \item[] Justification: We provide anonymized code and detail all implementation in our main paper and Appendix. 
    \item[] Guidelines:
    \begin{itemize}
        \item The answer NA means that the paper does not include experiments.
        \item If the paper includes experiments, a No answer to this question will not be perceived well by the reviewers: Making the paper reproducible is important, regardless of whether the code and data are provided or not.
        \item If the contribution is a dataset and/or model, the authors should describe the steps taken to make their results reproducible or verifiable. 
        \item Depending on the contribution, reproducibility can be accomplished in various ways. For example, if the contribution is a novel architecture, describing the architecture fully might suffice, or if the contribution is a specific model and empirical evaluation, it may be necessary to either make it possible for others to replicate the model with the same dataset, or provide access to the model. In general. releasing code and data is often one good way to accomplish this, but reproducibility can also be provided via detailed instructions for how to replicate the results, access to a hosted model (e.g., in the case of a large language model), releasing of a model checkpoint, or other means that are appropriate to the research performed.
        \item While NeurIPS does not require releasing code, the conference does require all submissions to provide some reasonable avenue for reproducibility, which may depend on the nature of the contribution. For example
        \begin{enumerate}
            \item If the contribution is primarily a new algorithm, the paper should make it clear how to reproduce that algorithm.
            \item If the contribution is primarily a new model architecture, the paper should describe the architecture clearly and fully.
            \item If the contribution is a new model (e.g., a large language model), then there should either be a way to access this model for reproducing the results or a way to reproduce the model (e.g., with an open-source dataset or instructions for how to construct the dataset).
            \item We recognize that reproducibility may be tricky in some cases, in which case authors are welcome to describe the particular way they provide for reproducibility. In the case of closed-source models, it may be that access to the model is limited in some way (e.g., to registered users), but it should be possible for other researchers to have some path to reproducing or verifying the results.
        \end{enumerate}
    \end{itemize}

\item {\bf Open access to data and code}
    \item[] Question: Does the paper provide open access to the data and code, with sufficient instructions to faithfully reproduce the main experimental results, as described in supplemental material?
    \item[] Answer: \answerYes{} % Replace by \answerYes{}, \answerNo{}, or \answerNA{}.
    \item[] Justification: We provide a link to anonymized code which has README instructions for running our models and experiments.
    \item[] Guidelines:
    \begin{itemize}
        \item The answer NA means that paper does not include experiments requiring code.
        \item Please see the NeurIPS code and data submission guidelines (\url{https://nips.cc/public/guides/CodeSubmissionPolicy}) for more details.
        \item While we encourage the release of code and data, we understand that this might not be possible, so “No” is an acceptable answer. Papers cannot be rejected simply for not including code, unless this is central to the contribution (e.g., for a new open-source benchmark).
        \item The instructions should contain the exact command and environment needed to run to reproduce the results. See the NeurIPS code and data submission guidelines (\url{https://nips.cc/public/guides/CodeSubmissionPolicy}) for more details.
        \item The authors should provide instructions on data access and preparation, including how to access the raw data, preprocessed data, intermediate data, and generated data, etc.
        \item The authors should provide scripts to reproduce all experimental results for the new proposed method and baselines. If only a subset of experiments are reproducible, they should state which ones are omitted from the script and why.
        \item At submission time, to preserve anonymity, the authors should release anonymized versions (if applicable).
        \item Providing as much information as possible in supplemental material (appended to the paper) is recommended, but including URLs to data and code is permitted.
    \end{itemize}

\item {\bf Experimental Setting/Details}
    \item[] Question: Does the paper specify all the training and test details (e.g., data splits, hyperparameters, how they were chosen, type of optimizer, etc.) necessary to understand the results?
    \item[] Answer: \answerYes{} % Replace by \answerYes{}, \answerNo{}, or \answerNA{}.
    \item[] Justification: We provide all data splits and hyperparameters in the main paper and Appendix. We additional provide all data splits in our code release. 
    \item[] Guidelines:
    \begin{itemize}
        \item The answer NA means that the paper does not include experiments.
        \item The experimental setting should be presented in the core of the paper to a level of detail that is necessary to appreciate the results and make sense of them.
        \item The full details can be provided either with the code, in appendix, or as supplemental material.
    \end{itemize}

\item {\bf Experiment Statistical Significance}
    \item[] Question: Does the paper report error bars suitably and correctly defined or other appropriate information about the statistical significance of the experiments?
    \item[] Answer: \answerNo{} % Replace by \answerYes{}, \answerNo{}, or \answerNA{}.
    \item[] Justification: Due to the deterministic nature of our models and existing resource constraints, we do not to report error bars.
    \item[] Guidelines:
    \begin{itemize}
        \item The answer NA means that the paper does not include experiments.
        \item The authors should answer "Yes" if the results are accompanied by error bars, confidence intervals, or statistical significance tests, at least for the experiments that support the main claims of the paper.
        \item The factors of variability that the error bars are capturing should be clearly stated (for example, train/test split, initialization, random drawing of some parameter, or overall run with given experimental conditions).
        \item The method for calculating the error bars should be explained (closed form formula, call to a library function, bootstrap, etc.)
        \item The assumptions made should be given (e.g., Normally distributed errors).
        \item It should be clear whether the error bar is the standard deviation or the standard error of the mean.
        \item It is OK to report 1-sigma error bars, but one should state it. The authors should preferably report a 2-sigma error bar than state that they have a 96\% CI, if the hypothesis of Normality of errors is not verified.
        \item For asymmetric distributions, the authors should be careful not to show in tables or figures symmetric error bars that would yield results that are out of range (e.g. negative error rates).
        \item If error bars are reported in tables or plots, The authors should explain in the text how they were calculated and reference the corresponding figures or tables in the text.
    \end{itemize}

\item {\bf Experiments Compute Resources}
    \item[] Question: For each experiment, does the paper provide sufficient information on the computer resources (type of compute workers, memory, time of execution) needed to reproduce the experiments?
    \item[] Answer: \answerYes{} % Replace by \answerYes{}, \answerNo{}, or \answerNA{}.
    \item[] Justification: We provide compute resources used for training and evaluation in the Appendix.
    \item[] Guidelines:
    \begin{itemize}
        \item The answer NA means that the paper does not include experiments.
        \item The paper should indicate the type of compute workers CPU or GPU, internal cluster, or cloud provider, including relevant memory and storage.
        \item The paper should provide the amount of compute required for each of the individual experimental runs as well as estimate the total compute. 
        \item The paper should disclose whether the full research project required more compute than the experiments reported in the paper (e.g., preliminary or failed experiments that didn't make it into the paper). 
    \end{itemize}
    
\item {\bf Code Of Ethics}
    \item[] Question: Does the research conducted in the paper conform, in every respect, with the NeurIPS Code of Ethics \url{https://neurips.cc/public/EthicsGuidelines}?
    \item[] Answer: \answerYes{} % Replace by \answerYes{}, \answerNo{}, or \answerNA{}.
    \item[] Justification: We adhere to the NeurIPS Code of Ethics guidelines, ensuring that our research and practices meet the ethical standards set forth by NeurIPS.
    \item[] Guidelines:
    \begin{itemize}
        \item The answer NA means that the authors have not reviewed the NeurIPS Code of Ethics.
        \item If the authors answer No, they should explain the special circumstances that require a deviation from the Code of Ethics.
        \item The authors should make sure to preserve anonymity (e.g., if there is a special consideration due to laws or regulations in their jurisdiction).
    \end{itemize}

\item {\bf Broader Impacts}
    \item[] Question: Does the paper discuss both potential positive societal impacts and negative societal impacts of the work performed?
    \item[] Answer: \answerYes{} % Replace by \answerYes{}, \answerNo{}, or \answerNA{}.
    \item[] Justification: Yes, we include a section on societal impacts of the work.
    \item[] Guidelines:
    \begin{itemize}
        \item The answer NA means that there is no societal impact of the work performed.
        \item If the authors answer NA or No, they should explain why their work has no societal impact or why the paper does not address societal impact.
        \item Examples of negative societal impacts include potential malicious or unintended uses (e.g., disinformation, generating fake profiles, surveillance), fairness considerations (e.g., deployment of technologies that could make decisions that unfairly impact specific groups), privacy considerations, and security considerations.
        \item The conference expects that many papers will be foundational research and not tied to particular applications, let alone deployments. However, if there is a direct path to any negative applications, the authors should point it out. For example, it is legitimate to point out that an improvement in the quality of generative models could be used to generate deepfakes for disinformation. On the other hand, it is not needed to point out that a generic algorithm for optimizing neural networks could enable people to train models that generate Deepfakes faster.
        \item The authors should consider possible harms that could arise when the technology is being used as intended and functioning correctly, harms that could arise when the technology is being used as intended but gives incorrect results, and harms following from (intentional or unintentional) misuse of the technology.
        \item If there are negative societal impacts, the authors could also discuss possible mitigation strategies (e.g., gated release of models, providing defenses in addition to attacks, mechanisms for monitoring misuse, mechanisms to monitor how a system learns from feedback over time, improving the efficiency and accessibility of ML).
    \end{itemize}
    
\item {\bf Safeguards}
    \item[] Question: Does the paper describe safeguards that have been put in place for responsible release of data or models that have a high risk for misuse (e.g., pretrained language models, image generators, or scraped datasets)?
    \item[] Answer: \answerYes{} % Replace by \answerYes{}, \answerNo{}, or \answerNA{}.
    \item[] Justification: We will add a safeguard agreement to our github repository when the code is publicly released. 
    \item[] Guidelines:
    \begin{itemize}
        \item The answer NA means that the paper poses no such risks.
        \item Released models that have a high risk for misuse or dual-use should be released with necessary safeguards to allow for controlled use of the model, for example by requiring that users adhere to usage guidelines or restrictions to access the model or implementing safety filters. 
        \item Datasets that have been scraped from the Internet could pose safety risks. The authors should describe how they avoided releasing unsafe images.
        \item We recognize that providing effective safeguards is challenging, and many papers do not require this, but we encourage authors to take this into account and make a best faith effort.
    \end{itemize}

\item {\bf Licenses for existing assets}
    \item[] Question: Are the creators or original owners of assets (e.g., code, data, models), used in the paper, properly credited and are the license and terms of use explicitly mentioned and properly respected?
    \item[] Answer: \answerYes{} % Replace by \answerYes{}, \answerNo{}, or \answerNA{}.
    \item[] Justification: We are the original creators of the assets used and give credit to previous work when building upon others' code or using external code or data. 
    \item[] Guidelines:
    \begin{itemize}
        \item The answer NA means that the paper does not use existing assets.
        \item The authors should cite the original paper that produced the code package or dataset.
        \item The authors should state which version of the asset is used and, if possible, include a URL.
        \item The name of the license (e.g., CC-BY 4.0) should be included for each asset.
        \item For scraped data from a particular source (e.g., website), the copyright and terms of service of that source should be provided.
        \item If assets are released, the license, copyright information, and terms of use in the package should be provided. For popular datasets, \url{paperswithcode.com/datasets} has curated licenses for some datasets. Their licensing guide can help determine the license of a dataset.
        \item For existing datasets that are re-packaged, both the original license and the license of the derived asset (if it has changed) should be provided.
        \item If this information is not available online, the authors are encouraged to reach out to the asset's creators.
    \end{itemize}

\item {\bf New Assets}
    \item[] Question: Are new assets introduced in the paper well documented and is the documentation provided alongside the assets?
    \item[] Answer: \answerYes{} % Replace by \answerYes{}, \answerNo{}, or \answerNA{}.
    \item[] Justification: Our accessible website (anonymized) offers easy access to our results and code. Additionally, we provide anonymized code that is well-documented with README files for setting up the models and environments.
    \item[] Guidelines:
    \begin{itemize}
        \item The answer NA means that the paper does not release new assets.
        \item Researchers should communicate the details of the dataset/code/model as part of their submissions via structured templates. This includes details about training, license, limitations, etc. 
        \item The paper should discuss whether and how consent was obtained from people whose asset is used.
        \item At submission time, remember to anonymize your assets (if applicable). You can either create an anonymized URL or include an anonymized zip file.
    \end{itemize}

\item {\bf Crowdsourcing and Research with Human Subjects}
    \item[] Question: For crowdsourcing experiments and research with human subjects, does the paper include the full text of instructions given to participants and screenshots, if applicable, as well as details about compensation (if any)? 
    \item[] Answer: \answerNA{} % Replace by \answerYes{}, \answerNo{}, or \answerNA{}.
    \item[] Justification: We do not use crowdsourcing or human subjects.
    \item[] Guidelines:
    \begin{itemize}
        \item The answer NA means that the paper does not involve crowdsourcing nor research with human subjects.
        \item Including this information in the supplemental material is fine, but if the main contribution of the paper involves human subjects, then as much detail as possible should be included in the main paper. 
        \item According to the NeurIPS Code of Ethics, workers involved in data collection, curation, or other labor should be paid at least the minimum wage in the country of the data collector. 
    \end{itemize}

\item {\bf Institutional Review Board (IRB) Approvals or Equivalent for Research with Human Subjects}
    \item[] Question: Does the paper describe potential risks incurred by study participants, whether such risks were disclosed to the subjects, and whether Institutional Review Board (IRB) approvals (or an equivalent approval/review based on the requirements of your country or institution) were obtained?
    \item[] Answer: \answerNA{} % Replace by \answerYes{}, \answerNo{}, or \answerNA{}.
    \item[] Justification: We do not use crowdsourcing or human subjects.
    \item[] Guidelines:
    \begin{itemize}
        \item The answer NA means that the paper does not involve crowdsourcing nor research with human subjects.
        \item Depending on the country in which research is conducted, IRB approval (or equivalent) may be required for any human subjects research. If you obtained IRB approval, you should clearly state this in the paper. 
        \item We recognize that the procedures for this may vary significantly between institutions and locations, and we expect authors to adhere to the NeurIPS Code of Ethics and the guidelines for their institution. 
        \item For initial submissions, do not include any information that would break anonymity (if applicable), such as the institution conducting the review.
    \end{itemize}

\end{enumerate}

\end{document}